**Ph.D. Dissertation**

# Microscopic Pedestrian Flow Characteristics: Development of an Image Processing Data Collection and Simulation Model

Department of Human Social Information Sciences
Graduate School of Information Sciences
Tohoku University
Japan

Kardi Teknomo

Reviewers:

Prof. Hajime Inamura

(Chairman)

Dr. Yasushi Takeyama

Prof. Hisa Morisugi

Prof. Kazuaki Miyamoto

Dr. Takashi Akamatsu

# ACKNOWLEDGMENTS

This dissertation has benefited from the comments and suggestions of many people. Thanks to Mr. Jin Nai and his foundation for the financial support. I am indebted to Prof. Hajime Inamura and Dr. Yasushi Takeyama who gave many useful improvements during the research and the reports. Their influence through useful discussions is immeasurable. I would like to express my gratitude to those who reviewed the manuscript and made useful comments and suggestions: Prof. Hisa Morisugi, Prof. Kazuaki Miyamoto and Dr. Takashi Akamatsu. I am very grateful to Prof. Kochiro Deguchi who also made valuable suggestions especially on a part of chapter 3. I am also grateful to Dr. Katsuya Hirano for his comments on Chapter 4. Thanks to Mr. Tetsuro Harayama who helped to collect the manual data. Finally, thanks to my wife, Gloria P. Gerilla who caught misprints and mistakes in the earlier draft of this manuscript and for her love and encouragement.

Kardi Teknomo

# Abstract

## Microscopic Pedestrian Flow Characteristics: Development of an Image Processing Data Collection and Simulation Model

Kardi Teknomo

Microscopic pedestrian studies consider detailed interaction of pedestrians to control their movement in pedestrian traffic flow. The tools to collect the microscopic data and to analyze microscopic pedestrian flow are still very much in its infancy. The microscopic pedestrian flow characteristics need to be understood. Manual, semi manual and automatic image processing data collection systems were developed. It was found that the microscopic speed resemble a normal distribution with a mean of 1.38 m/second and standard deviation of 0.37 m/second. The acceleration distribution also bear a resemblance to the normal distribution with an average of 0.68 m/ square second.

A physical based microscopic pedestrian simulation model was also developed. Both Microscopic Video Data Collection and Microscopic Pedestrian Simulation Model generate a database called $a$TXY database. The formulations of the flow performance or microscopic pedestrian characteristics are explained. Sensitivity of the simulation and relationship between the flow performances are described. Validation of the simulation using real world data is then explained through the comparison between average instantaneous speed distributions of the real world data with the result of the simulations.

The simulation model is then applied for some experiments on a hypothetical situation to gain more understanding of pedestrian behavior in one way and two way situations, to know the behavior of the system if the number of elderly pedestrian increases and to evaluate a policy of lane-like segregation toward pedestrian crossing and inspects the performance of the crossing. It was revealed that the microscopic pedestrian studies have been successfully applied to give more understanding to the behavior of microscopic pedestrians flow, predict the theoretical and practical situation and evaluate some design policies before its implementation.

# TABLE OF CONTENTS

# LIST OF FIGURES

# LIST OF TABLES

# CHAPTER 1 INTRODUCTION

## 1.1 MICROSCOPIC PEDESTRIAN STUDIES

Increased awareness of environmental problems and the need for physical fitness encourage the demand for provision of more and better pedestrian facilities. To provide better pedestrian facilities, the appropriate standard and control of the facilities need to be determined. To decide the appropriate standard and control of pedestrian facilities, pedestrian studies, which consist of pedestrian data collection and pedestrian analysis, need to be done. One of the objectives of the pedestrian studies is to evaluate the effects of a proposed policy on the pedestrian facilities before its implementation. The implementation of a policy without pedestrian studies might lead to a very costly trial and error due to the implementation cost (i.e. user cost, construction, marking etc.). On the other hand, using good analysis tools, the trial and error of policy could be done in the analysis level. Once the analysis could prove a good performance, the implementation of the policy is straightforward. The problem is how to evaluate the impact of the policy quantitatively toward the behavior of pedestrians before its implementation.

As suggested by [[1]], the traffic flow characteristics could be divided into two categories, microscopic level and macroscopic level. Microscopic level involves individual units with traffic characteristics such as individual speed and individual interaction. Most of the pedestrian studies that have been carried out are on a macroscopic level. Macroscopic pedestrian analysis was first suggested by [[2],[3]] followed by many researchers and has been adopted by [[4]]. While the macroscopic pedestrian data-collection is recommended by [[5]] wherein all pedestrian movements in pedestrian facilities are aggregated into flow, average speed and area module. The main concern of macroscopic pedestrian studies is space allocation for pedestrians in the pedestrian facilities. It does not consider the direct interaction between pedestrians and it is not well suited for prediction of pedestrian flow performance in pedestrian areas or buildings with some street furniture (kiosk, benches, telephone booths, fountain, etc.). Microscopic pedestrian studies, on the other hand, treat every pedestrian as an individual and the behavior of pedestrian interaction is measured. Though the microscopic pedestrian study does not replace the macroscopic one, it considers a more detailed analysis for design and pedestrian interaction.

In contrast to a pedestrian who walks alone, the increase in the number of pedestrians in the facilities creates problems of interaction. The pedestrians influence each other in their walking behavior either with mutual or reciprocal action. They need to avoid or overtake each other to be able to maintain their speed, they need to change their individual speed and direction and sometimes they need to stop and wait to give others the chance to move first. In a very dense situation, they need to maintain their distance / headway toward other pedestrians and surroundings to reduce their physical contact to each other. Thus, a pedestrian tends to minimize the interaction between pedestrians. Because of the interaction, the pedestrians feel uncomfortable, and experience delay (inefficiency). Interaction between pedestrians, as the important point in the microscopic level, can be modeled as a repulsive and attractive effect between pedestrians and between pedestrian with their environment.

The importance of detailed design and pedestrian interaction is best exemplified using the case studies that have been done by [[6]]. The case studies used microscopic pedestrian simulation to determine the flow performance of pedestrians in the intersection of pedestrian malls and doors as illustrated in Figure 1-1. The left figure shows the intersection of pedestrian malls with roundabout. Each pedestrian is denoted by an arrow. The study compares the flow performance (comfortability and delay) of the pedestrians in the intersection with and without the roundabout. It was revealed that pedestrian flow performance of the intersection with roundabout is better than without the roundabout.

Pedestrian flow that is more efficient can even be reached with less space. Those simulations have rejected the linearity assumption of space and flow in the macroscopic level. The right figure represents two rooms connected with two doors and the pedestrians are coming from both sides of the rooms. Two simple scenarios were experimented. The first scenario was letting the pedestrians pass through any door (two way door), while in the second scenario each door can be passed by only one direction. The result of the experiments showed that one-way door is better than a two-way door. The movement of pedestrian needed to be controlled so that the interaction problem is reduced.

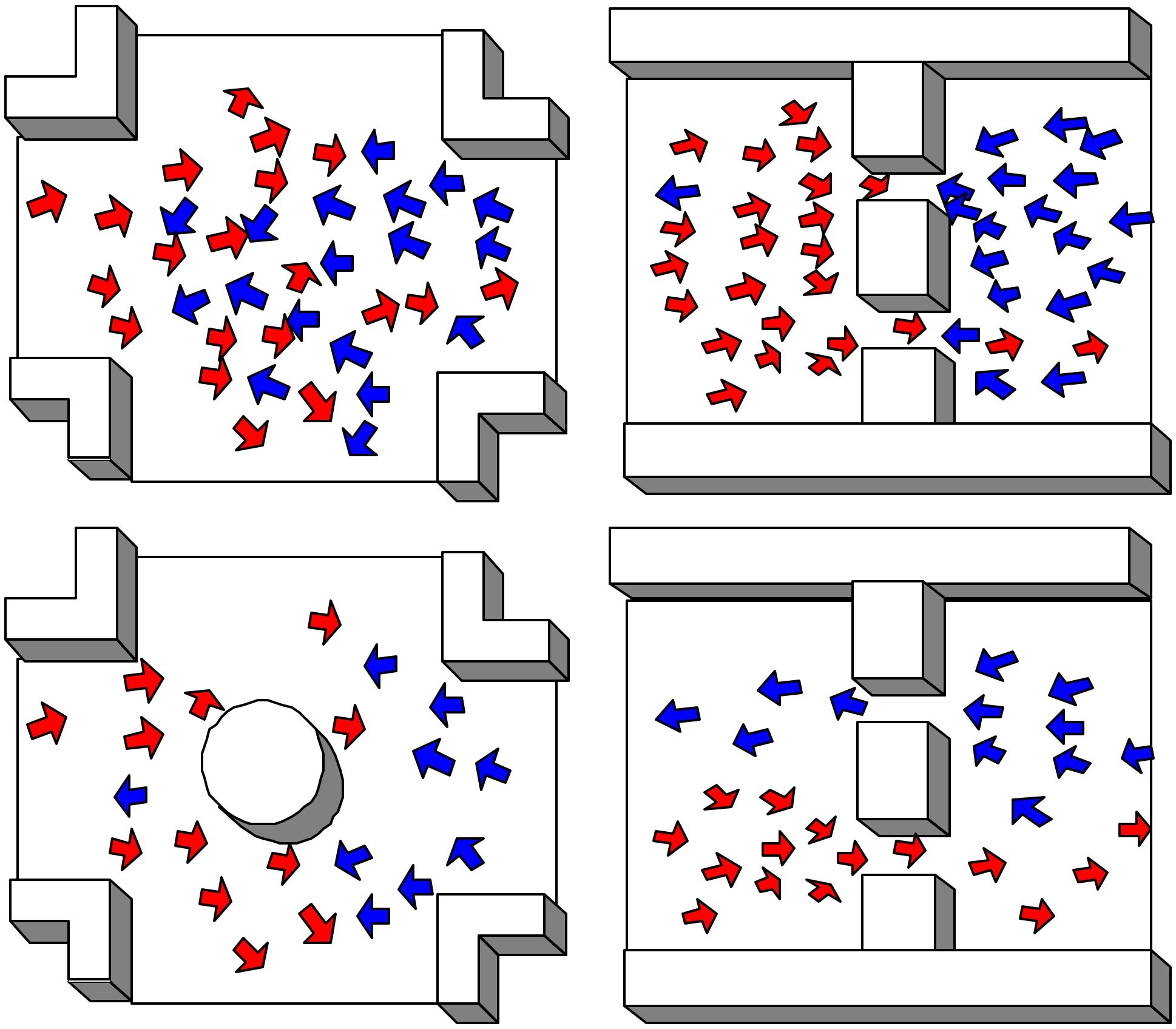

**Figure 1-1. Controlling Pedestrian Movement to Reduce the Interaction Problem**

The generalization of those case studies leads to a new paradigm. Instead of merely allocating a space for pedestrians, the movement quality of pedestrians is considered as the new goal. In the old paradigm, using the macroscopic pedestrian studies, given a number of pedestrians and a level of service, the model may give the space allocation (i.e. width of the facilities). In the microscopic level, however, given the same number of pedestrian and the same space, with better set of rules and detailed design, a better flow performance may be produced. In the macroscopic level of analysis, space of the pedestrian facilities is only a way to control the pedestrian flow. Using the microscopic pedestrian studies, a wider way to control the pedestrian facilities can be utilized. Pedestrian interaction can be measured and controlled. Pedestrian flow performance is defined as the indicators to measure the interaction between pedestrians. The pedestrian interaction can be controlled by time, space and direction. Pedestrians may be allowed to wait for some time, or walk at a particular space (e.g. door) or right of way (e.g. walkway), or at certain directions. This more comprehensive pedestrian-flow control happens because microscopic pedestrian studies consider pedestrian interaction. Since the movement quality of pedestrians can be improved by controlling the interaction between pedestrians, better pedestrian interaction is the objective of this approach. Figure 1-2 shows the system approach to improve the movement quality of pedestrians.

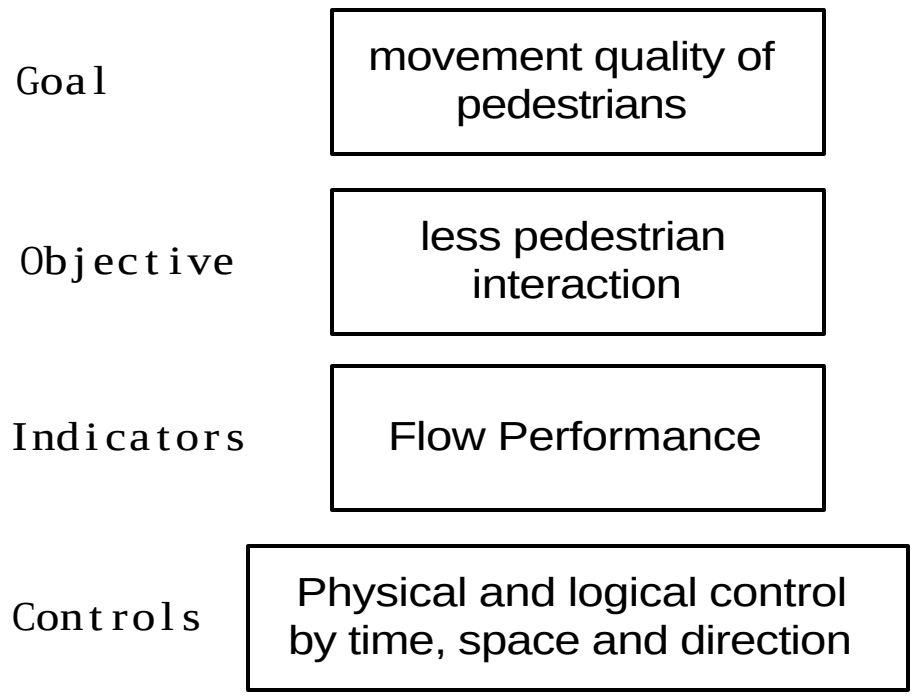

**Figure 1-2. Paradigm to Improve the Quality of Pedestrian Movement**

Compared to the macroscopic pedestrian studies, the microscopic pedestrian studies are still very much in its infancy. Despite the greater benefit of the microscopic pedestrian studies, the number of researches and papers on this subject has been remarkably few. Among those researches, some microscopic pedestrian analyses have been developed (see Chapter 2 for more details). The analytical model for microscopic pedestrian model has been developed by [[7] and [8]], but the numerical solution of the model is very difficult and simulation is more practical and favorable. Though microscopic pedestrian analysis exists through simulations, several problems have not been addressed by the previous researchers.

1. Most of previous microscopic pedestrian analyses were not concerned with the traffic characteristic or flow performances of pedestrians because the main concern was in the modeling of the simulation. Using the simulation model, what are the microscopic characteristics of pedestrian flow?
2. Microscopic pedestrian data collection has not been developed. Recently, several studies to perform pedestrian surveillance have been actively developed in the computer vision and image processing fields. Those studies, however, do not specify the purpose toward traffic engineering field, especially the microscopic level of pedestrian data. How should we use the pedestrian traffic surveillance system to collect microscopic pedestrian data?
3. Once such microscopic pedestrian data is collected, another problem on how to measure the flow performance from the microscopic data collection arises. The results of microscopic data collection are the locations of each pedestrian at each time slice. How to reduce these huge data into information that can be readily understood and interpreted?

Thus, this study as reported in this dissertation is done to solve those aforementioned problems.

## 1.2 AIMS OF STUDY

The purpose of this study is to improve the quality of pedestrian movement behavior through microscopic pedestrian studies. The specific objectives in this study are:

1. To identify the existing stage of microscopic pedestrian studies;
2. To develop a data collection system for microscopic pedestrian studies;
3. To improve the existing microscopic simulation models;
4. To examine the microscopic pedestrian flow characteristics; and
5. To discuss the application of the microscopic pedestrian simulation models.

## 1.3 SCOPE OF STUDY

This study is mainly concerned with the microscopic pedestrian traffic characteristics from both the simulation and the real world data. The systems that were developed in this study consider only pedestrians in two-dimensional areas. Pedestrians in stairs or elevators are not investigated. Mixed traffic between pedestrian and vehicular traffic is not examined either.

## 1.4 DISSERTATION OUTLINE

The intention has been to make this report self-contained. The structure of this report is illustrated in Figure 1-3 and described as follows. This chapter of introduction has presented the background and motivation of the microscopic pedestrian studies, the purpose, and the scope of the dissertation. The state of the art, in the following chapter, considers the result of the previous studies about microscopic pedestrian studies, from data collection to analysis.

Chapter 3 introduces the microscopic pedestrian data collection. Three image-processing systems were developed to gather pedestrian database that is called $a$TXY database, namely manual, semi manual and automatic system. The chapter is intended to explain the detailed step of the development of those systems and its results. Chapter 4 describes the development of Microscopic Pedestrian Simulation Model. Comparison with the existing models is also explained. Chapter 5 is the core of the dissertation. The result of the data collection and the simulation model are combined through the $a$TXY database. The formulations of the Microscopic Pedestrian Characteristics from the $a$TXY database are examined. Sensitivity analysis of the simulation and its calibration and validation are described. Chapter 6 demonstrates the application of the analysis. The simulation model is applied for some experiments on a hypothetical situation to gain more understanding of pedestrian behavior in one way and two way situation, to know the behavior of the system if the number of elderly pedestrian increases and to evaluate a policy of lane-like segregation toward pedestrian crossing and inspects the performance of the crossing. Chapter 7 concludes the dissertation with several new paradigms and summaries of the results. Further research recommendations are also included.

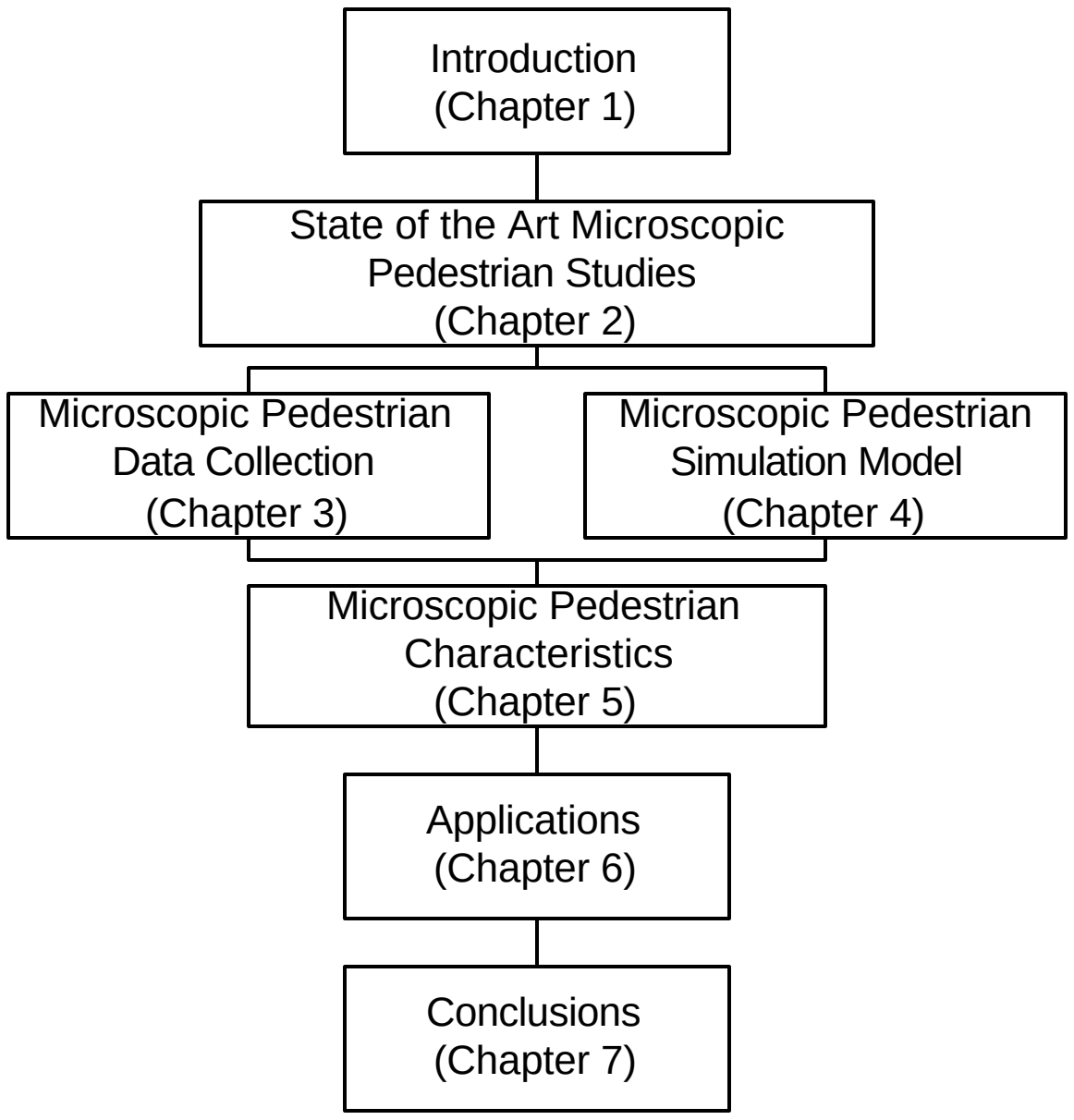

**Figure 1-3 Structure of Dissertation**

# CHAPTER 2 STATE OF THE ART: MICROSCOPIC PEDESTRIAN STUDIES

This chapter introduces some previous studies up to the current state in the Microscopic Pedestrian Studies and gives reviews that are used in the later chapters. The chapter is divided into two main parts, which are pedestrian studies and pedestrian characteristics. Emphasis is given towards the microscopic pedestrian studies and characteristics.

## 2.1 PEDESTRIAN STUDIES

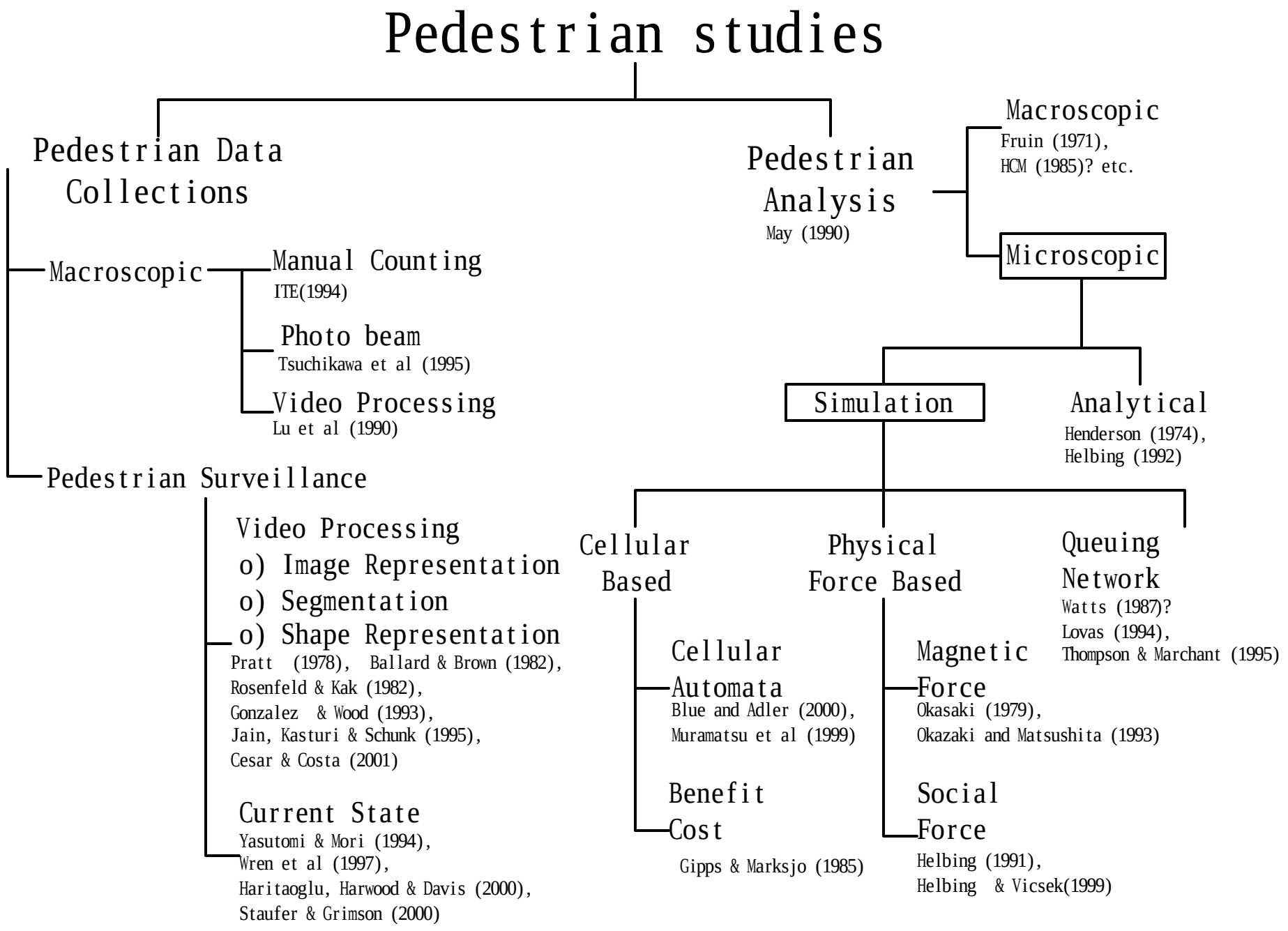

**Figure 2-1 Schematic of Pedestrian Studies with emphasis on the microscopic level**

The knowledge of the pedestrian traffic system mainly comes from observations and empirical studies. Pedestrian studies can be divided into pedestrian data collection and pedestrian analysis. The data collection consists of the task associated with the observation and recording of pedestrian movement data while pedestrian analysis is focused on the interpretation of the data in order to understand the observed situation and to plan and design improvements. Figure 2-1 shows the overall picture of pedestrian studies that will be explained in this chapter.

Similar to vehicular traffic as suggested by [1], pedestrian traffic studies can also be divided into two categories, microscopic level and macroscopic level. Microscopic level involves individual units with traffic characteristics such as individual speed and individual interaction while the macroscopic pedestrian studies aggregates the pedestrian movements into flow, average speed and area module.

The macroscopic pedestrian studies have been developed since 1971 by [2],[3] and many other researchers. The analysis has even been adopted by the HCM standard [4]. The microscopic pedestrian analysis, however, begin with Henderson [5] that compares the pedestrians crowds data with a gas kinetic and fluid dynamic model. Helbing [6] revised the Henderson model and took into account the intention, desire velocities and pair interactions of individual pedestrians. The numerical solution of the mathematic model, however, is very difficult and simu lation approach is more practical [7].

### 2.1.1 Pedestrian Analysis by Simulations

The Microscopic Pedestrian Simulation Model (MPSM) is a computer simulation model of pedestrian movement where every pedestrian in the model is treated as an individual. Based on the internal model of the simulation, the MPSM can be categorized into three types, cellular based, physical force based and queuing network model (see Figure 2.1 for more detail category). Among the cellular based, two types of models were established. Gipps and Marksjo [8] developed microscopic simulation using cost and benefit cell, while Blue and Adler [9] developed the cellular automata model for pedestrian. Among the physical based model, two models were recognized which is magnetic force model and social force model. The magnetic force model was started by Okazaki [10] and followed by [11]. Social force model was developed by Helbing [12] and improved by several researchers ([13],[14]). The use of microscopic pedestrian simulation for evacuation purposes was developed by several researchers ([15],[16],[17],[18]) that use queuing network model.

It is interesting to note two things. Firstly, there are many types of MPSM and each of them do not relate to each other. The data from one type of MPSM cannot be used interchangeably with another type of model. In chapter 5, a unifying language is proposed to relate the data from all types of MPSM. Secondly, most of the microscopic pedestrian simulations were not calibrated statistically and none of them has been calibrated using microscopic level data. It has no statistical guarantee that the parameters will work for general cases or even for a specific region. Such calibration was not possible without the ability to measure individual pedestrian-movement data.

There are several similarities and differences between the models. In this sub section, general comparison toward the existing models and the proposed model is explained. In general, Microscopic Pedestrian Simulation Model consist of two terms:

1. Term that makes the pedestrian move towards the destination
2. Term that makes a repulsive effect toward other pedestrian or obstacles.

The first term, represented by Gain score in the Benefit Cost Cellular model, is similar to attractive force between goals, pedestrian in the Magnetic Force model, and equivalent to Intended velocity in the Social Force model. The second term, indicated by cost score in the Benefit Cost Cellular model, is comparable to repulsive force plus force to avoid collision with other pedestrian or obstacles in Magnetic Force model. This term also resembles the Interaction forces in the Social force model. However, Cellular Automata does not show the two terms explicitly, but it can be derived from the movement updating rules. Queuing network model use weighted random choice to make the pedestrian move toward destination and priority rule (i.e. first in first serve) to govern the interaction between pedestrians.

The following are a brief description about the Microscopic Pedestrian Simulation Models that have been mentioned above.

#### 2.1.1.1 Benefit Cost Cellular Model

Gipps and Marksjo [8] propose this model. It simulates the pedestrian as a particle in a cell. The walkway is divided into a square grid (i.e. 0.5 by 0.5 $m^2$ per cell). Each cell can be occupied by at most one pedestrian and a score assigned to each cell based on proximity to pedestrians. This score represents the repulsive effect of the nearby pedestrians and has to be balanced against the gain made by the pedestrian in moving toward his destination. Where the field of two pedestrians overlap, the score in each cell is the sum of the score generated by each pedestrian individually.

Initially, a cell occupied by a pedestrian is given a score of 1000, the score of a cell with a side in common is 40 and cell with corner is scored 13. The scoring is arbitrary. The score of the surrounding cell of a pedestrian is approximately inversely proportional to the square of the separation of pedestrian in two cells as shown:

$$S = \frac{1}{(\Delta - a)^2 + b} \qquad \textbf{(2. 1)}$$

Where

S = Cost Score of a cell $k$ of moving closer to other pedestrians or objects (repulsive effect)

$\Delta$ = Distance between cell $i$ and the pedestrian.

$a$ = 0.4, a constant slightly less than diameter of pedestrian (=0.5 m)

$b$ = 0.015, arbitrary constant number to moderate fluctuation in score close to the pedestrian.

The gain score is given by

$$P(s_i) = K.\cos(s_i).\left|\cos(s_i)\right| = \frac{K(S_i - X_i)(D_i - X_i).\left|(S_i - X_i)(D_i - X_i)\right|}{\left|S_i - X_i\right|^2.\left|D_i - X_i\right|^2} \qquad \textbf{(2. 2)}$$

Where

$P(s_i)$ = Gain score for moving closer to his destination. It is defined to be zero if the pedestrian remain stationary,

$K$ = constant of proportionality to enable the gain of moving in a straight line to be balanced against the cost of approaching other pedestrian closely,

$s_i$ = the angle by which the pedestrian deviates from a straight line to his immediate destination when moving to cell $i$,

$S_i$ = vector location of target cell,

$X_i$ = vector location of the subject,

$D_i$ = vector location of destination.

The net benefit,

$$B = S - P(s_i) \qquad \textbf{(2. 3)}$$

is calculated in the nine cell neighbors of the pedestrian (including the location of the pedestrian). The pedestrian will move to the next cell that has maximum net benefit.

The main benefit of this model is its simplicity but the model suffers much problem due to the arbitrary scoring of the cells and the pedestrians. The scoring system makes the model difficult to be calibrated with the real world phenomena.

#### 2.1.1.2 Cellular Automata Model

Cellular Automata models have been applied for simulating car traffic and validate adequately with the real traffic data. Recently, cellular automation model has been used for pedestrians ([19],[20],[21]).

The model simulates pedestrians as entities (automata) in cells. The walkway is modeled as grid cells and a pedestrian is represented as a circle that occupies a cell. The occupancy of a cell depends on localized neighborhood rules that are updated every time. Each pedestrian movement includes both lane changing and cell hopping. In each time step, each cell can take on one of two states: occupied and unoccupied.

Two parallel stages to update the rules ares applied in each time step of the simulation. The first stage is the rule of lane changing: if either or both adjacent lanes, immediately to the left or right of a pedestrian are free (unoccupied and within the defined walkway), then the pedestrian is assigned to the lane, current or adjacent, which has the maximu m gap. If there is more than one lane available, lane assignment is determined randomly with some probability distribution. The second stage is assigning speed, based on the available gap and advanced forward by this speed. A gap is the number of empty cells ahead. The range of allowable movement is equal to minimum of one of gap or maximum walking speed.

Though the cellular automata model is also simple to develop and fast to update the data, the heuristic approach of the updating rules is undesirable since it does not reflect the real behavior of the pedestrian. The inherent grid cells of the cellular based model make the behavior of pedestrians seems rough visually. The pedestrian gives the impression of jumping from one cell to another. Nevertheless, Blue and Adler (2000), give an excellent idea on validation of the microscopic model using the existing model fundamental diagram.

#### 2.1.1.3 Magnetic Force Model

Okazaki ([22],[23]) developed this model with Matsushita ([24],[25]) and Yamamoto [26]. The application of magnetic models and equations of motion in the magnetic field cause pedestrian movement. Each pedestrian has a positive pole. Obstacles, like walls, columns, handrails also have positive pole and negative poles are assumed located at the goal of pedestrians. Pedestrians move to their goals and avoid collisions. Each pedestrian is attracted by 'an attraction', with a negative magnetic charge, as his destination of movement, walks avoiding other pedestrians or 'obstructions' such as walls with positive magnetic charges. If a force from another pole influences a pedestrian, the pedestrian moves with accelerated velocity. The velo city of the pedestrian increases as the force continues to act on it until the upper limit of velocity. At the same time a pedestrian and another pedestrian and an obstruction repulse each other. Coulomb's law calculates Magnetic Force, which acts on a pedestrian from a magnetic pole:

$$\mathbf{F} = \frac{k.q_1.q_2.\mathbf{r}}{r^3} \qquad \textbf{(2. 4)}$$

where:

$\mathbf{F}$ = magnetic force (vector),

$k$ = constants,

$q_1$ = intensity of magnetic load of a pedestrian,

$q_2$ = intensity of a magnetic pole,

$\mathbf{r}$ = vector from a pedestrian to a magnetic pole, and

$r$ = length of r.

Another force acts on a pedestrian to avoid the collision with another pedestrian or obstacle exerts acceleration **a** and is calculated as:

$$\mathbf{a} = \mathbf{V}.\cos(alpha).\tan(beta) \qquad \textbf{(2. 5)}$$

Where:

$\mathbf{a}$ = acceleration acts on pedestrian A to modify the direction of **RV** to the direction of line AC,

**V** = velocity of pedestrian A,

alpha = angle between **RV** and **V,**

beta = angle between **RV** and AC,

**RV** = relative velocity of pedestrian A to pedestrian B.

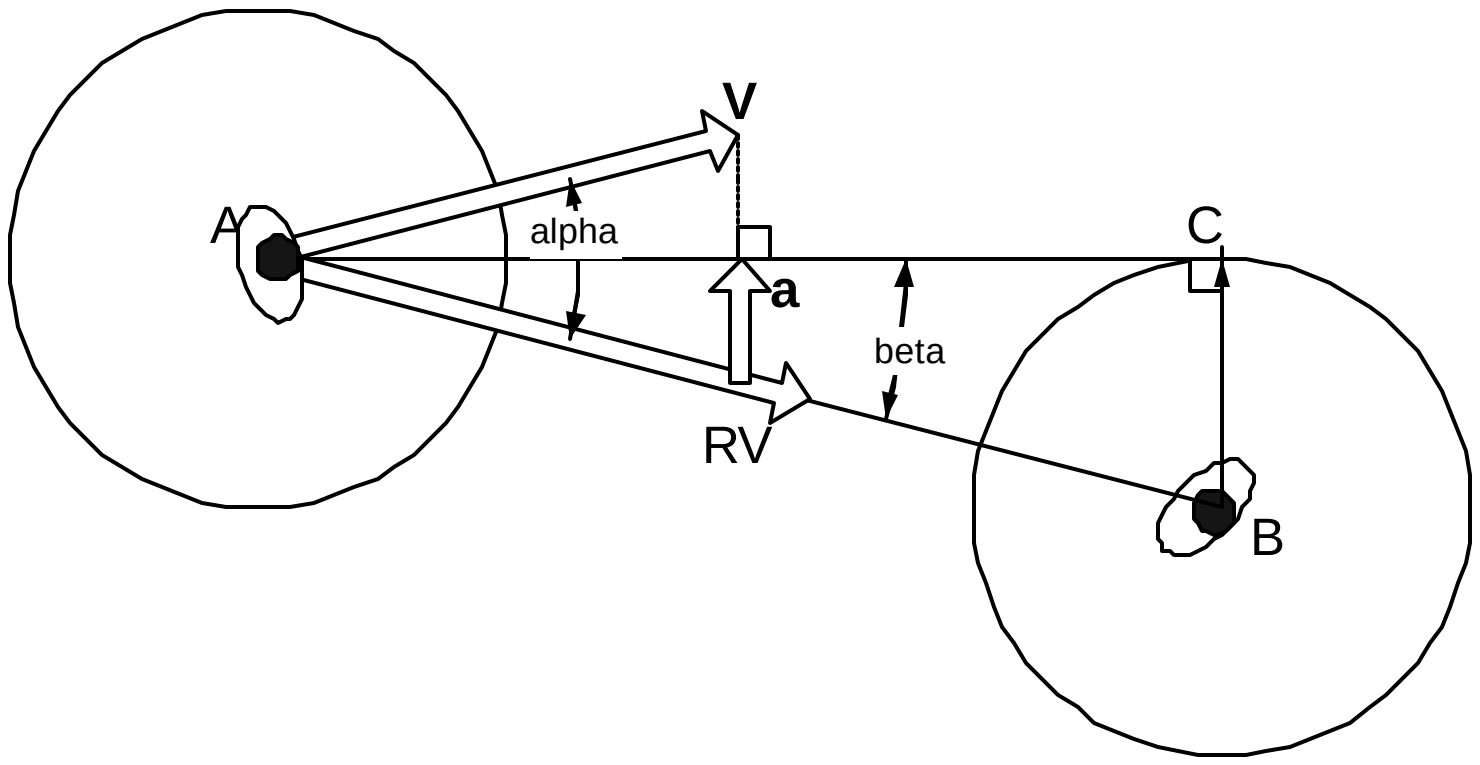

**Figure 2-2 Additional force to avoid collision in Magnetic Force model**

The sum of forces from goals, walls and other pedestrians act on each pedestrian, and it decides the velocity of each pedestrian each time. The intensity of magnetic load is an arbitrary number, however, if a large value of magnetic load is assigned to a pedestrian, the repulsive force is larger and the distances to wall and other pedestrians are longer.

Walls and other obstacles are given as sequences of points. Lines, which connect the sequences of points, are displayed for show. In a complicated plan where pedestrians cannot directly move to their goal, special points on the wall (called Corner), is assumed as temporary goals which lead them to their final destination.

The idea of using additional force to avoid collision is excellent and will be used in the proposed model. This model, however, undergo a similar problem as the benefit cost cellular model where the value of the magnetic intensity are set as arbitrary numbers. Due to those arbitrary setting of the magnetic load, the validation of the model can only be done merely by visual inspection. No real world phenomena can be validated using this model.

#### 2.1.1.4 Social Force Model

Helbing [12] has developed the Social Force Model with Molnar [13], Schweitzer and Vicsek [14], which has similar principles to both Benefit Cost cellular Model and Magnetic Force Model. A pedestrian is assumed subjected to social forces that motivate the pedestrian. The summation of these forces that act upon a pedestrian create acceleration $d\mathbf{v}/dt$ as:

$$m\frac{d\mathbf{v}_i(t)}{dt} = m\frac{v_o\mathbf{e}_i - \mathbf{v}_i(t) + \mathbf{?}_i(t)}{t} + \sum_{j(\neq i)}\mathbf{f}_{ij}(\mathbf{x}_i(t), \mathbf{x}_j(t)) + \mathbf{f}_b(\mathbf{x}_i(t)) \qquad \textbf{(2. 6)}$$

where

$\mathbf{x}_i(t)$ =Location of pedestrian $i$ at time $t$,

$\mathbf{v}_i(t)$ = velocity of pedestrian $i$ at time $t$ = $d\mathbf{x}_i(t)/dt$,

$m$ = mass of pedestrian; $m/t$ may be interpreted as a friction coefficient,

$v_0$ = intended velocity with which it tend to move in the absence of interaction,

$\mathbf{e}_i$ = direction into which pedestrian $i$ is driven $\in \{(1,0),(0,1)\}$,

$\mathbf{?}_i(t)$ = the fluctuation of individual velocities,

$\mathbf{f}_{ij}$ = the repulsive interaction between pedestrian $i$ and $j$,

$\mathbf{f}_b$ = the interaction with the boundaries.

The motivation to reach the goal produces the intended velocity of motion. The model is based on the assumption that every pedestrian has the intention to reach a certain destination at a certain target time. Every movement that he makes will be directed toward that destination point. The direction is a unit vector from a particular location and the destination point. The direction is given by

$$\mathbf{e}_i = \frac{\mathbf{x}_i^0 - \mathbf{x}_i(t)}{\left\|\mathbf{x}_i^0 - \mathbf{x}_i(t)\right\|} \qquad \textbf{(2. 7)}$$

The ideal speed is equal to the remaining distance per remaining time. Remaining distance is the length of the difference between destination point and the location at that time, while the remaining time is the difference between target time and the simulation time. The ideal speed is obtained by

$$u = \frac{\left\| \mathbf{x}_i^0 - \mathbf{x}_i(t) \right\|}{T_i - t} \qquad \textbf{(2. 8)}$$

Intended Velocity is the ideal speed times the unit vector of direction. We can put a speed limitation (maximum and minimum) to make the speed more realistic.

Two types of interaction is noted:

1. Interaction between pedestrian;
2. Interaction between pedestrian and obstacles.

Interaction between pedestrians and ped estrian to obstacles (i.e. column) is calculated as:

$$\mathbf{f}_{ij}(\mathbf{x}_i(t), \mathbf{x}_j(t)) = -\nabla A(d_{ij} - D)^{-B} \qquad \textbf{(2. 9)}$$

Where

$B$ = constants;

$d_{ij}$ = distance between pedestrian $i$ and $j$;

$D$ = diameter that represents space occupied by particle $j$;

$A$ = a monotonic decreasing function.

Interaction of pedestrian with the boundaries is given by:

$$\mathbf{f}_b(\mathbf{x}_i) = -\nabla A(d_{i\perp} - D/2)^{-B}) \qquad \textbf{(2. 10)}$$

Where $d_{i\perp}$ = shortest distance to the closest wall.

The social force model is the best among all microscopic models that has been developed so far. The variables are not arbitrary because they have physical meaning that can be measured. The results of the model also show self-organizing phenomena. Nevertheless, there are two critiques for this model. First, the interaction model does not guarantee that the pedestrian will not collide (overlapping) with each other. It is unrealistic if the pedestrian can enter another pedestrian visually, especially when the pedestrian density is very high. Another force is needed to avoid collision, similar to the magnetic force model. Second, the model has never been validated with the real world data or phenomena. It seems that the researchers of social force model are more focused on the physical interactions to explain biological and physical behaviors rather than the real pedestrian traffic flow.

#### 2.1.1.5 Queuing Network Model

The use of microscopic pedestrian simulation for evacuation purposes was developed by several authors ([15],[16],[17],[18]). They used a queuing network model as evacuation tools from fire in the building. The approach is a discrete event Monte Carlo simulation, where each room is denoted as a node and the door between rooms as links. Each person departs from one node, queue in a link, and arrive at another node. A number of pedestrians move from one node to another in search for the exit door. Each pedestrian has a location goal. Each person has to move from its present position to an exit as quickly and safely as possible. Route, which each person use and the evacuation time is recorded in each node. When a pedestrian arrives in a node, he makes a weighted-random choice to choose a link among all possible links. The weight is a function of actual population density in the room. If the link cannot be used, a pedestrian will wait or find another route to follow. In the source node (initial condition at time 0), a person needs a certain time to react before movement begins, while in the final destination node he will stop the movement process.

Pedestrian crossing has a similar goal to the evacuation where the pedestrians have to move from their original position to the other side of the road as quickly and safely as possible. The evacuation time (dissipation time), as one of the performance measurement will be used in the proposed model. The queuing network model has implicit visual interaction. The behavior of the pedestrians is not clearly shown and the collisions among pedestrians are not clearly guaranteed. The FIFO priority rule that is inherent in the model is not very realistic especially in a crowded situation.

### 2.1.2 Pedestrian Data Collection

Aside from pedestrian analysis, pedestrian studies also include pedestrian data collection. Technological advance of computer and video processing over a decade has changed pedestrian studies significantly. Progression of analysis has demanded better data collection and the progress in data collection method improves the analysis further toward a more detailed design. To decide the appropriate standard and control of pedestrian facilities, pedestrian data collection and analysis need to be done. Planning and design of pedestrian facilities should obviously reflect their anticipated usage. Surveys to provide information about current usage are often carried out at intersections, at mid block crossing, along pathways or at public transport terminals (modes).

Typically, manual counting was performed by tally sheet or mechanical or electronic count board to collect volume and speed data for pedestrian. Pedestrian behavior studies are recommended by manual observation or video. Though vehicular automatic counting has been improved through pneumatic tube or inductance loops, the similar technology cannot be used to detect pedestrian.

By mid 90's, video and CCTV were increasingly popular as an "automatic" source of vehicular and pedestrian traffic data. Their advantage is to store the data in videotape, which can be revisited to provide information on other aspect of the scene. Volume and speed data can be gathered separately at different times in the laboratory. Taping and filming provides an accurate and reliable means of recording volumes, as well as other data, but requires time-consuming data reduction in the office [27]. The expense of reducing video data was very high because it must be done manually in the laboratory.

The macroscopic pedestrian counting device was developed as a research work of Lu et al [28] using video camera. Macroscopic flow characteristic can be gathered automatically. The researcher however, limited themselves for special background and special treatment of the camera location. Tsuchikawa et al [29] use one-line detection as a development of photo -beam technology to count the number of pedestrian p assing that line with the camera on top.

#### 2.1.2.1 Current State Pedestrian Surveillance

The microscopic pedestrian data collection methods have not been developed by any researchers; on the other hand, the pedestrian traffic surveillance however, has been developed in the computer vision fields. Traffic surveillance is used for security and monitoring and not for data collection. The following are the brief summary of some methodologies in Pedestrian Surveillance to discuss their different methods and to show the current state of pedestrian surveillance.

[30] and [31] detect pedestrians based on rhythm with the camera from the side. The advantage of using the rhythm does not depend on clothes, distance, weather, and simple to perform in real time. The model is based on the motion model of pedestrian. Motion object was detected and segmented by image difference. Then the Sobel edge detector and thresholding performed in parallel to emphasize the moving object region. Image was binarized and projected horizontally and vertically. The bottom position and the width of a moving object were determined by slicing two projections. If the width and height of a projected image was within a certain threshold, then a window was put in the image. Feet were assumed located 1/5 of the bottom of the window, divided into left and right window. To detect the occlusion, if the normalized variance of object width is smaller than a threshold, it is assumed that there is no occlusion. If there is occlusion, a pedestrian was tracked with its prediction.

Tracking utilizes the estimation of the object position and velocity based on a kinematics model, a measurement model and tracking filter.

The ordinate represent distance of the person from the camera, and its time series was predicted by least square method. Object recognition based on rhythm that is spatial frequency and temporal frequency, are used to discriminate pedestrian from other moving objects. The recognition based on the finding when both feet of a pedestrian are in the ground, the motion has only small change in intensity. When one foot is moving forward, its motion has large change in intensity. Fast Fourier Transform (FFT) was applied on the time series of two binarized areas. When the first component of the Fourier spectrum of two time series are matched and located between two standard deviation of the mean rhythm of walking, the two windows are judge as the feet of a person. A pedestrian moving alone can be tracked with very good accuracy.

The main weakness of this method is its dependency on the result of experiment to determine many thresholds as parameters that they used. Another weakness is the usage of motion detection by projection. If more than one object move near each other, the algorithm will detect that as an occlusion though the object does not occlude (but the projection is occluded).

Different methods of pedestrian tracking and recognition were used by different researchers. The tracking and recognition is related to the shape representation of the pedestrian as object. [32] use active contour model so called the active deformation model to represent the pedestrian. Image was subtracted from the static background scene and made into binary image using a certain threshold to get the moving object. If the object was large enough to be detected as a possible pedestrian, a bounding box is placed and an active deformable model is placed around the possible pedestrian with the control point on the bounding box. Then movements are controlled by minimization of energy equations. The result was reported as able to overcome occlusion but was not possible to tell if it tracked the same individual.

Another way to segment and estimate the motion was reported by [33] that use the block-matching algorithm for motion estimation of a face for coding video sequence. Reliability measurement increases if the motion vector belongs to a moving object with *constant speed*, and if the reliability is smaller than certain threshold, temporal prediction is abandoned, and spatial prediction is used.

Tsuchikawa [29] uses PedCount, a pedestrian counter system using CCTV. It extracts the object using the one line path in the image by background subtraction to make a space-time (X-T) binary image. The passing direction of each pedestrian is determined by the slant of pedestrian region in the X-T image. They reported the need of background image reconstruction due to image illumination change. The background image was captured later when there was no object or pedestrian and then averaging the background images to reduce illumination variation. An algorithm to distinguish moving object from illumination change is explained based on the variance of the pixel value and frame difference. Another method that implemented space-time image was reported by [34] which used the area of pedestrian region in the X-T plane as descriptor. They used the algorithm to count the number of pedestrians. They also used a white background, linear camera (from the top) and illumination level adjuster materials. The detection used a line detector, similar to X-T plane of Tsuchikawa, but in here, two parallel lines were used to be able to detect the speed of pedestrian. To detect a pedestrian into a unity region, morphology erosion was used. Different sizes and shapes of structuring element were used for each category of speed level. Another researchers [35] exploited spatio-temporal XT slices to obtain a trajectory pattern of a human walking. The method, however, has not been successful to detect many pedestrians with occlusion cases.

In case of descriptors, several attempts have been made to represent the pedestrians. Oren et al [36] found out that wavelet transformation from pedestrian image might have a clear pattern that can be used to classify pedestrian objects from the scene. Wavelet is invariant to the change of color & texture. However, it is not clear if the wavelet coefficients of an image are unique for each pedestrian. Grove et al [37] attempt to use color (hue and saturation) as descriptor of segmentation and tracking by assumption that objects have a distinct color from the background or other objects. The color of an object is relatively constant under a viewpoint change and therefore insensitive to modest rotation or camera movement. They used the Gaussian Mixture Model (GMM) as distribution of color within the image and the probability of color of a pixel. The parameters are calculated based on Expectation Maximization (EM). Based on the color distribution, an object is separated from the background by a threshold. After some morphological processes were done to remove noises and merge large region to find the connected component, simple features (bounding box, area, centroid coordinates, and eccentricity) are calculated. Threshold over areas are done to get the most likely object, and it is taken as the Region of Interest (ROI) at time t. Position of ROI at time t + 1 is predicted by the centroid and the probability of non-background color within the object. These probabilities model a color histogram. The width and height of the ROI (box) is determined by the variance of color probability. Matching of the color histogram is performed by a normalized cross correlation between recent histogram and histogram to date. If it has a strong correlation, the background model is updated; otherwise, the object is tracked with a non-background to obtain more samples to refine the histogram. To overcome the occlusion, an occlusion buffer is made. If objects occlude, the system reads from the buffer rather than from the scene. Occlusion status is determined by sorting objects according to their lowest point. It is assumed that objects with the lowest point are nearer to the camera. The algorithm has relatively low computational cost compared to 3D geometric model, but a little bit poorer than background subtraction. The assumption of distinct color of object is doubtful.

Heisele and Woehler [38] tried to segment pedestrians from the moving background. The scene image is clustered based on the color and position (R, G, B, X, Y) of pixel. A box containing of the clustered leg of pedestrian is used for pedestrian representation during the training period and the first two coefficient of the Fast Fourier Transform was used as descriptor for matching. Matching is done by a time delay neural network for object recognition and motion analysis.

Onoguchi [39] wanted to estimate the pedestrian based on size, shape and location but got difficulties due to the shadow of a moving object. To overcome that problem, two cameras were used to remove the shadow image. Image from camera 1 was inversely projected into a road plane view and then transformed back to the image of camera 2 coordinates using the Kanatani transformation. Several corresponding points of the two images were determined manually and calibrated by regression. The transformed image was threshold based on a cross correlation to get a mask image. The thresholding was done empirically (manual). The object was detected by image subtraction of consecutive frame and it took only the moving part from the merging mask.

Significant advancement of pedestrian motion analysis was recently developed with side view camera. Staufer and Grimson [40] employ event detection and activities classification on the video camera for monitoring people activities (direction, coming and going). Haritaoglu et al [41] detect single and multiple people and monitor their activities in an outdoor environment. It detects the people through their silhouettes and recognizes their activities with reasonable accuracy.

## 2.2 PEDESTRIAN CHARACTERISTICS

In this section, pedestrian characteristics based on the previous studies are discussed. The pedestrian characteristics can be divided into macroscopic and microscopic characteristics. The formulation of the fundamental diagram is also described.

### 2.2.1 Macroscopic Characteristics

Fundamental characteristics of traffic flow are flow, speed and density. These characteristics can be observed and studied at the microscopic and macroscopic levels. The macroscopic characteristics concern with the groups of pedestrians rather than the individual unit of pedestrian. Macroscopic analysis may be selected for high density, large-scale systems in which the behavior of groups of unit is sufficient. There are many macroscopic pedestrian characteristics but for this report, the main concerns are those characteristics that relate to the simulation and data collection in a very short distance walkway (i.e. pedestrian trap). Other characteristics such as journey distance, trip purposes, socio economic characteristics etc. are not discussed.

The US Highway Capacity Manual standard [4] produced Table 2-1 that shows the relation of space, average speed and flow rate at different levels of service.

Pedestrian flow rate denoted by $q$ is a result of a movement of many individuals. Pedestrian flow rate or volume is defined as the number of pedestrian that pass a perpendicular line of sight across a unit width of a walkway during a specified period of time ([28]) and normally has a unit of ped/min/m (number of pedestrian per minutes per meter width). Pedestrian volume is useful for examining the trend and planning facilities, evaluating safety and level of service. If $w$ and $L$ denote the width and length of the pedestrian trap respectively, and $N$ indicates the number of pedestrians observed during the observation time $T$, then the flow rate can be calculated as

$$q = \frac{N}{T.w} \qquad \textbf{(2. 11)}$$

**Table 2-1 Pedestrian Level of Service on Walkway**

| Level of service | Space m$^2$/ped (ft$^2$/ped) | Expected Flow and speed* Average speed m/min (m/sec) | Flow rate ped/min/m (ped/min/ft) | V/C ratio |
|---|---|---|---|---|
| A | ≥12.077 (130) | ≥79.248 (1.321) | ≤6. 562 (2) | ≤0.08 |
| B | ≥3.716 (40) | ≥76.200 (1.270) | ≤22.966 (7) | ≤0.28 |
| C | ≥2.230 (24) | ≥73.152 (1.219) | ≤32.808 (10) | ≤0.40 |
| D | ≥1.394 (15) | ≥68.580 (1.143) | ≤49.213 (15) | ≤0.60 |
| E | ≥0.557 (6) | ≥45.720 (0.762) | ≤82.021 (25) | ≤1.00 |
| F | < 0.557 (6) | < 45.720 (0.762) | variable | variable |

* Average condition for 15 minutes

Source: [4] with unit converted

Walking speed is an important element of design, particularly at at-grade road crossing. It provides sufficient crossing time to enable the entire pedestrian to complete the road-crossing maneuver before vehicular traffic begins to move. For uncongested corridor, [15] assumed that walking speed depends only on personal factors and it follows a log normal distribution. However, [42] and [43] found that the desired speeds within pedestrian crowds are Gaussian distributed.

There are two common ways to compute the average or mean speed, which is called time mean speed and space mean speed. The *time mean speed* is the average speed of all pedestrian passing a line on the pedestrian trap over a specified period of time and it is calculated as an arithmetic average of the spot speed or instantaneous speed, that is

$$\tilde{v}(t) = \frac{\sum_{i=1}^{N} v_i(t)}{N} \qquad \textbf{(2.12)}$$

where $N$ is the number of observed pedestrian and $v_i$ is the instantaneous speed of the $i^{th}$ pedestrian. The time mean speed, $\tilde{v}$ , is taken as an average value over specified duration of time corresponding to the observation of flow, density, space mean speed and other characteristics (e.g. every 5 minutes of observation). If the walking distance of all individual pedestrians, $w_i$, during fixed observation periods $T$ can be gathered, the time mean speed can be also be calculated using

$$\tilde{v} = \frac{\sum_{i=1}^{N} w_i}{N.T} \qquad \textbf{(2.13)}$$

The s*pace mean speed* is the average speed of all pedestrian occupying the pedestrian trap over a specified time period and calculated based on the average travel time for the pedestrian to traverse a fixed length of a pedestrian trap, $L$ . If $t_i^{out}$ and $t_i^{in}$ represent time of pedestrian $i^{th}$ to go out and go in the pedestrian trap, the space mean speed, $u$, is calculated as

$$u = \frac{L}{\tilde{t}} \qquad \textbf{(2.14)}$$

Where the denominator is the average travel time

$$\tilde{t} = \frac{\sum_{i=1}^{N}(t_i^{out} - t_i^{in})}{N} \qquad \textbf{(2.15)}$$

Fruin (1971) suggests that people are able to walk at their characteristic speed if density is below 0.5 ped/m$^2$. O'Flaherty [44] summarized that road crossing speed has indicated an average value in the range of 1.2 m/s to 1.35 m/s at busy crossings with mix of pedestrian age groups. However, if the crossings are less busy, the average walking speed approximating to the free-flow walking speed of 1.6 m/s. For disabled persons, 0.5 m/s is the more appropriate value.

The relationship between speed, flow and density or area module, which is called the fundamental traffic flow formula, is given by

$$q = u.k \qquad \textbf{(2.16)}$$

The pedestrian traffic density denoted by $k$. Pedestrians keep a certain distance to other pedestrians and borders (of streets, walls and obstacles). This distance becomes smaller, the more pedestrian hurries and it decreases with growing pedestrian density. Papacostas and Prevedouros [45] define pedestrian density or concentration as the number of pedestrians within a unit area (ped/m$^2$). The reciprocal of pedestrian density is called Space module or Area Module, denoted by $M$, which is a unit of surface area per pedestrian (m$^2$/ped). The area module is calculated as area of the pedestrian trap per number of pedestrian observed during the period $T$. Based on Equation (2. 11),(2. 14) and (2. 16), the definition of the

area module is

$$M = \frac{u}{q} = \frac{L/\tilde{t}}{N/T.w}$$

or

$$M = \frac{w.L.T}{N.\tilde{t}} \qquad \textbf{(2. 17)}$$

### 2.2.2 Microscopic Characteristics

Unlike the macroscopic pedestrian characteristics that are quite well defined, the microscopic pedestrian characteristics are not clearly defined. Fruin [3], and Navin and Wheeler [46] discussed about headway measurement of pedestrian. The headway is defined as distance of one pedestrian from another according to the direction of movement. The definition and the characteristics, however, remain unclear since the direction of pedestrian is changing over time.

Before proceeding further with other microscopic pedestrian characteristics, definition of *average* or *mean* must be cleared up. In case of microscopic pedestrian movement data, there are two kinds of average value:

1. Average of pedestrian flow performance;
2. Time average of pedestrian flow performance.

The first average is done by the summation of the specified flow performance divided by the number of pedestrian. This mean or average is denoted by a curly bar ( $\tilde{}$ ). The instantaneous speed in (2. 12) or average travel time in (2. 15) are the examples of the average flow performance. The second average is a time average of a specified flow performance at certain time t. It can be done by averaging the flow performance over time. The time average is symbolized by straight bar ( $\bar{}$ ). Brown and Hwang [47] have introduced a simple recursive formula to estimate the mean or the average value of a sequence of measurement, which can be used to ease the calculation of time average. The sequence measurement can be any value of the performance that measured every time, such as speed, delay, or uncomfortability index, etc. Let $z_i$ be the $i^{th}$ measurement of the sequence (e.g. instantaneous speed) and $\overline{m}_i$ is the estimate mean (e.g. the average of the instantaneous speed) where the subscript denotes the time at which the measurement is taken. Using the common procedure, the mean will be calculated as $\overline{m}_1 = z_1$; $\overline{m}_2 = \frac{z_1 + z_2}{2}$; $\overline{m}_3 = \frac{z_1 + z_2 + z_3}{3}$ and so on. The procedure is storing all the measurements sequence and the number of arithmetic operation is increasing according to the number of data in the sequence. It creates memory and computational speed problems. A better way to calculate the average value is using a recursive formula

$$\overline{m}_t = \left(\frac{t-1}{t}\right)\overline{m}_{t-1} + \left(\frac{1}{t}\right)z_t \qquad \textbf{(2. 18)}$$

yields identical results as the common procedure but without the need to store all the previous measurements. The recursive algorithm utilizes the result of the previous step to obtain the average value at the current step. The recursive formula can be used to calculate both average and time average. Thus, *average* is always summation of items divided by the number of measurements, while *time average* is the summation of items divided by the duration of time measurement.

Helbing and Molnar (1997) proposed a flow performance of *efficiency measure* and uncomfortableness measure as evaluation measures to optimize the pedestrian facilities. The efficiency measure, $\tilde{E}$, calculates the mean value of the velocity component into the desired direction of motion in relation to the desired walking speed, and given by

$$\tilde{E} = \frac{1}{N}\sum_i \frac{x_i}{v_i^o} \qquad \textbf{(2. 19)}$$

The uncomfortableness measure, $\tilde{U}$, reflects the frequency and degree of sudden velocity changes, i.e. the level of discontinuity of walking due to necessary avoidance maneuvers.

$$\tilde{U} = \frac{1}{N}\sum_i \frac{\bar{y}_i}{\bar{h}_i} \qquad \textbf{(2.20)}$$

Where

$$\bar{x}_i = \frac{\sum_{t=t_1}^{t_2} \mathbf{v}_i(t).\mathbf{e}_i(t)}{t_2 - t_1},\ \bar{\mathbf{g}}_i = \frac{\sum_{t=t_1}^{t_2} \mathbf{v}_i(t)}{t_2 - t_1},\ \bar{h}_i = \frac{\sum_{t=t_1}^{t_2} \mathbf{v}_i^2(t)}{t_2 - t_1} \text{ and } \bar{y}_i = \frac{\sum_{t=t_1}^{t_2} \left(\mathbf{v}_i(t) - \bar{\mathbf{g}}_i\right)^2}{t_2 - t_1}$$

$\mathbf{v}_i(t)$ = velocity of pedestrian $i$ at time $t$,

$N$ = the number of pedestrian $i$,

$v_i^o$ = intended velocity of pedestrian $i$,

$\mathbf{e}_i(t)$ = direction into which pedestrian $i$ is driven at time $t$,

The $\bar{x}_i$, $\bar{\mathbf{g}}_i$, $\bar{h}_i$, and $\bar{y}_i$ are time average from $t_1$ to $t_2$.

### 2.2.3 Fundamental Diagram

The traffic fundamental diagram is a graph that relates flow rate and space mean speed of the traffic. To make the graph, it should be kept in mind that the speed and density are dependent on each other. If the density is bigger, the distance between pedestrian tends to be smaller and the speed is reduced. If the free flow speed denoted by $m_f$ and the traffic jam density is denoted by $k_j$ and the speed-density relationship is assumed to be linear, the relationship of speed, $u$, and density, $k$, is given by

$$u = m_f - y.k \qquad \textbf{(2.21)}$$

Defining the linear gradient between speed and density relationship as $y = m_f / k_j$, the fundamental diagram or the relationship between flow rate, $q$, and speed can be derived as

$$q = \frac{u(m_f - u)}{y} \qquad \textbf{(2.22)}$$

The relationship of flow rate with the density is

$$q = m_f k - y\,.k^2 \qquad \textbf{(2.23)}$$

**Table 2-2. Comparison pedestrian characteristics of several previous studies**

| Source | Country | Type | Free flow speed $m_f$ (m/min) | Traffic jam density $k_j$ (ped/m$^2$) | Ratio $y$ (ped.min/m$^3$) | Capacity (ped/min/m) |
|---|---|---|---|---|---|---|
| Oeding [48] | Germany | Mixed traffic | 89.9 | 3.98 | 22.6 | 89.40 |
| Older [49] | Britain | Shoppers | 78.64 | 3.89 | 20.2 | 76.54 |
| Navin & Wheeler [46] | USA | Students | 97.6 | 2.70 | 36.2 | 65.79 |
| Fruin [2] | USA | Commuter | 81.4 | 3.99 | 20.4 | 81.20 |
| Tanaboriboon et al [51] | Singapore | Mixed traffic | 73.9 | 4.83 | 15.3 | 89.24 |
| Tanaboriboon & Guyano [50] | Thailand | Mixed traffic | 72.85 | 5.55 | 13.13 | 101.05 |
| Yu [52] | China | Mixed traffic | 75.45 | 5.10 | 14.83 | 95.97 |
| Gerilla [53] | Philippines | Mixed traffic | 83.23 | 3.60 | 23.11 | 74.94 |

Note: converted from the formulation with unit conversion i f necessary

The link between flow rate and the Area module, $M$ , is

$$q = \frac{m_f}{M} - \frac{y}{M^2} \qquad \textbf{(2.24)}$$

It is easy to derive that the maximum flow or the capacity, $Q$, for the linear relationship of speed and density given by

$$Q = \frac{m_f . k_j}{4} \qquad \textbf{(2.25)}$$

Based on the above formula, the graph of the relationship of flow rate, speed, and density and area module are given in Figure 2-3.

Table 2-2 shows the free flow speed, jam density and capacity from several studies in many countries. All those researchers use linear models as described above. Other models that are not linear are also used as comparison purposes only and normally failed to obtain higher correlation ([53]). Some examples of non-linear models are

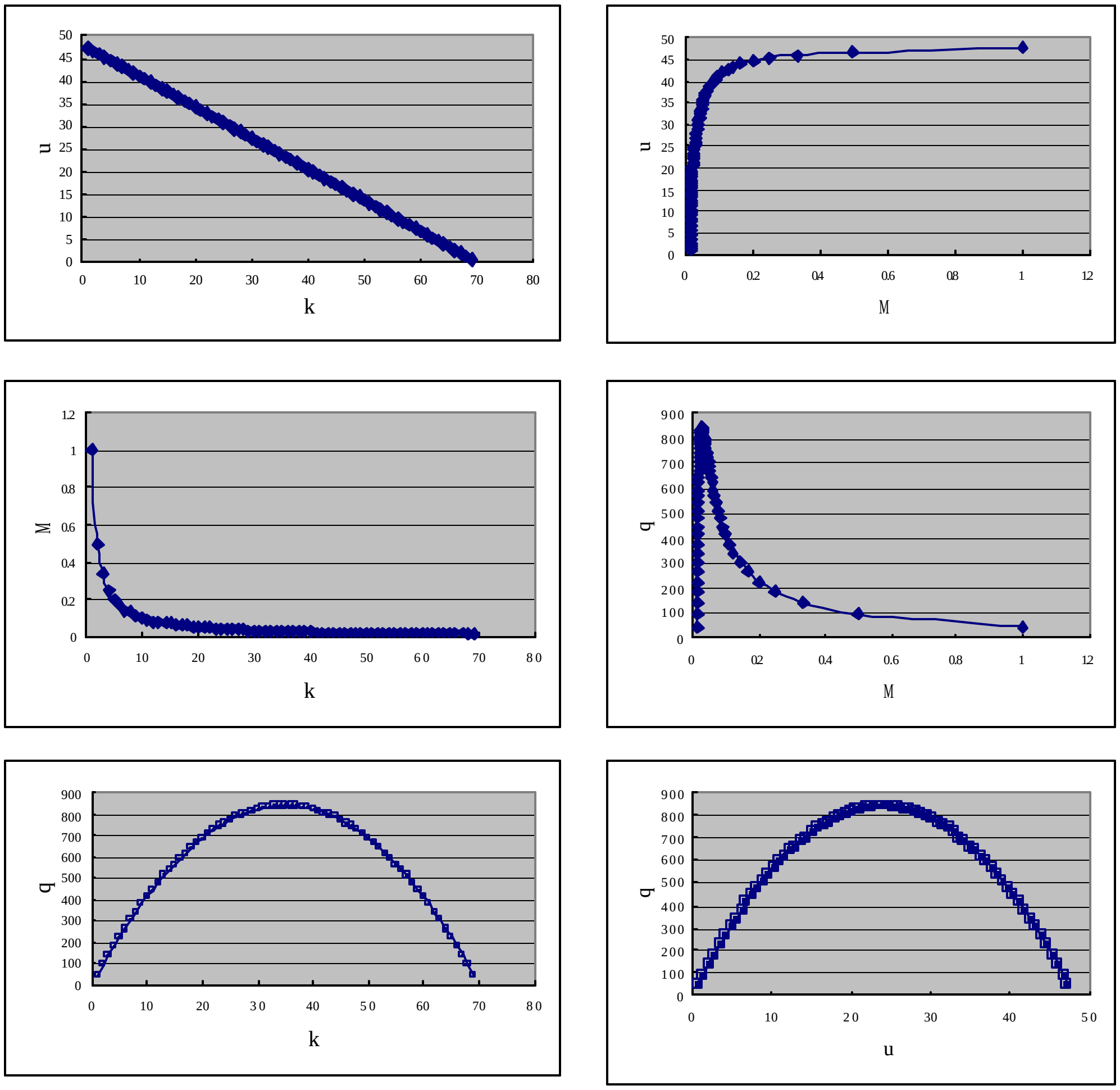

**Figure 2-3 Relationships of Speed, Density, Area Module and Flow Rate Based on Linear Relationship of Speed and Density**

$$u = u_c . \ln(\frac{k_j}{k}) \qquad \textbf{(2.26)}$$

$$u = m_f . \exp(-\frac{k}{k_c}) \quad \textbf{(2.27)}$$

$$u = m_f . \exp(-\tfrac{1}{2}(\frac{k}{k_c})^2) \quad \textbf{(2.28)}$$

# CHAPTER 3 MICROSCOPIC VIDEO DATA COLLECTION

Though one intention of this research is to develop a system to gather microscopic pedestrian flow data from the existing pedestrian traffic surveillance, the real pedestrian traffic surveillance itself still needs to be developed. The existing pedestrian traffic surveillance were developed in the image processing and computer vision fields that have no purpose to gather the microscopic pedestrian flow data. Thus, a special method to gather microscopic pedestrian flow data from the video images was developed and discussed in this chapter. Result of microscopic data collection is demonstrated in the end of this chapter. The method consisting of four steps, as shown in Figure 3-1, is proposed to enhance of the existing pedestrian video surveillance toward microscopic pedestrian flow data collection.

The video processing is explained from a simple manual data collection, semi automatic and a further attempt to automate the system. As the existing pedestrian-surveillance system still cannot fully overcome the occlusion problem, the automatic system that developed in this chapter attempt to propose a tracing method to overcome very short occlusion.

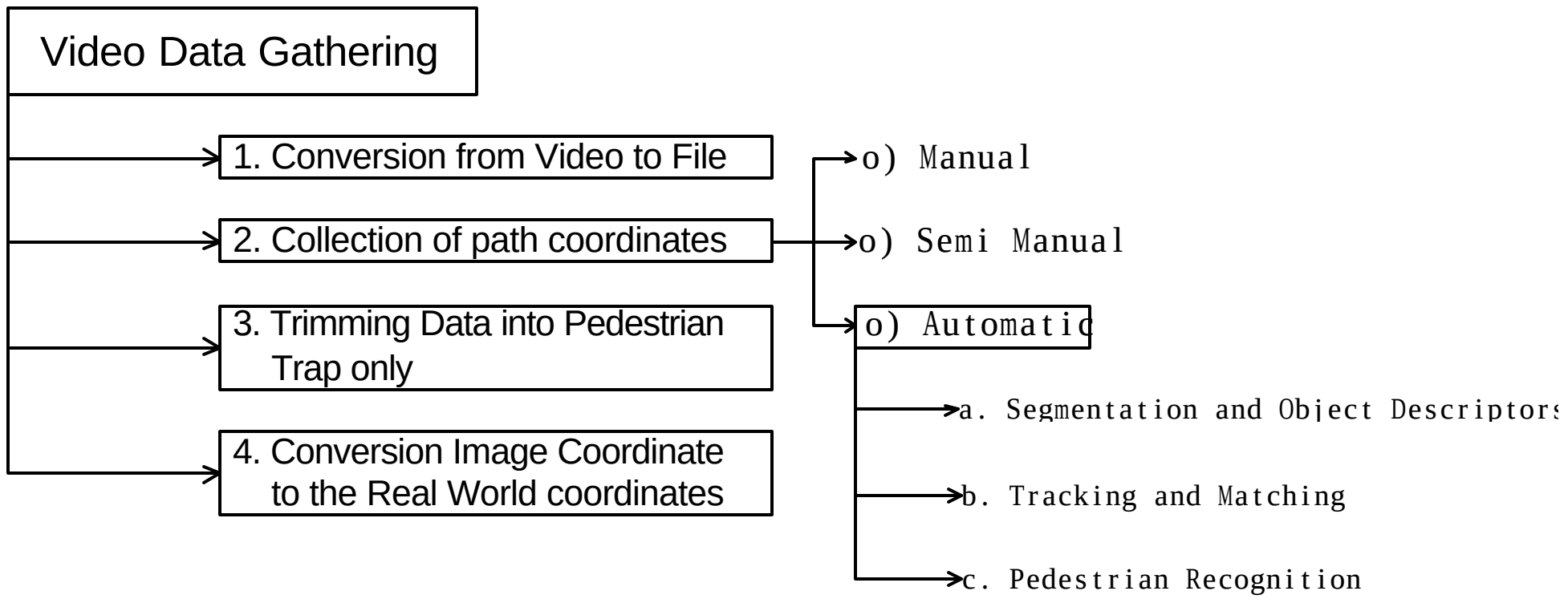

**Figure 3-1 Microscopic Video Data Gathering**

## 3.1 VIDEO DATA GATHERING

There are three reasons why video processing is a prominent tool to collect microscopic pedestrian data. First, video as vision sensor has traditionally been used to collect macroscopic pedestrian data. Second, the cost of video is relatively cheap and it is getting cheaper. The technologies of video processing is growing very rapidly and tend to automate the system. Third, since we need to track the movement of pedestrians over a relatively wide area instead of simply counting the number of pedestrians, it makes other types of sensors too costly to implement.

The existing pedestrian surveillance systems, as explained in chapter 2, have been used to monitor the people activities that are mainly for security reason. The following four steps are suggested to improve the current surveillance system into a microscopic pedestrian data collection method:

1. Conversion from Video to File;
2. Collection of path coordinates ( $a$ TXY);
3. Trimming Data into Pedestrian Trap only;
4. Conversion Image Coordinate to the Real World coordinates .

The subsequent sections are explanation of the steps.

### 3.1.1 Conversion from Video to File

Images of pedestrians are taken by a video static camera in the field, on top of pedestrian facilities (i.e. walkway). A pedestrian trap, which is an imaginary rectangle, is marked in the middle of walkway as boundary of the system. Only pedestrians who pass the trap are considered. The images from the video are then converted into files in the laboratory using any commercial software. A freeware "NIH image" [1],[2] (Macintosh system with fire wire cable) is used for this research . The video film is captured as an image sequence at the rate of 30 Hertz (30 frames per second).

### 3.1.2 Collection of Path Coordinates

The main problem is to record the position of every pedestrian in every frame in the image sequence. Based on the automation development process, the video data collection can be classified into three types:

1. Manual data collection;
2. Semi automatic data collection;
3. Automatic data collection.

All of these three types will be explained in more detail in the next section.

Based on how the pedestrian data is assembled, there are two possible methods to gather the pedestrian path coordinate:

1. Pedestrian-Based Tracking;
2. Frame-Based Tracking.

Pedestrian-based tracking is suitable for manual or semi automatic data collection where each pedestrian is followed from the time he/she enters the system until he/she goes out of the system. A pedestrian that has been tracked is marked. After one pedestrian is tracked, the next pedestrian is searched over time and tracked. The process is repeated until all pedestrians in the stack of images are tracked. Frame-based tracking records the position of all pedestrians in each frame and finds the matched pedestrian between frames. The matched pedestrian will be given the same pedestrian number. No repetition of searching the next pedestrian over time is needed. Since it is difficult for humans to memorize the pedestrian number of each pedestrian in a frame, this method is more suitable for automatic data collection. The automatic data collection, however, can utilize either pedestrian-based or frame-based method.

To ease the process and to save the memory of the computer, a pedestrian is represented by a single point in every frame. The middle part of the feet of the pedestrian is taken as the representative point. If other parts of the body are going to be used, it is difficult to be consistent for all pedestrians and it is difficult for calibration purposes. Every pedestrian that is observed in the image is numbered uniquely. This number is useful to distinguish the data of a pedestrian from another.

The result of the collection of path coordinate is called the $a$TXY database consisting of 4 fields, namely Pedestrian number $a$, slice number that represents the time $T$, and position of the pedestrian at that frame $X_i, Y_i$. Every record in the $a$TXY Database represents a single observation. One observation point is the position of a single pedestrian in one frame. Chapter 5 will discuss in more detail about the usage of $a$TXY database.

### 3.1.3 Trimming Data into Pedestrian-Trap Only

A pedestrian Trap is an imaginary rectangle marked in the middle of a walkway as boundary to count the pedestrian. Only the pedestrians who pass the trap are considered. A pedestrian trap is used to ease the measurement. The width is perpendicular to the pedestrian way while the length is parallel to the walkway. It is easier to calculate the area and to get the coordinates of a rectangle than any other shape. The images were taken by angles from the pedestrian walkway, and they covered a non-rectangular area. Due to lens distortion, the smallest error will be in the middle of the image and higher error at the edge of the image. A pedestrian trap needs to be defined to reduce the lens distortion error and to ease the measurement.

Any point in the pedestrian trap can serve as origin point of the coordinate. To make the coordinate X larger to the right, while coordinate Y larger to the north, the origin must be in the southwest of the pedestrian trap. Four points (A, B, C, and D) are the corner points of the pedestrian trap in the real world coordinate. Since the coordinate of the real world will contain errors of calibration, to reduce the error, the data trimming outside the pedestrian trap was done using image coordinates $(X_i, Y_i)$ instead of using real world coordinate $(X, Y)$. The trimming process was done as follows:

**Algorithm 3.1:**

1. The image coordinates of four points (A', B', C', D') that bound the real world rectangular are obtained from the image file.

2. The line equations of A'B', B'D', C'D', A'D' are derived from the straight-line equation.
3. Any row from the $a$TXY Database that contains $Y_i$ below line A'B' or above line C'D' or $X_i$ on the left line A'C' or on the right line B'D' is trimmed out from the $a$TXY database.

### 3.1.4 Conversion of Image Coordinate to the Real World coordinates.

Several models of camera calibrations have been considered. Since the problem to convert the image coordinate to real world coordinate is simply a transformation from 2 dimensional spaces into another two-dimensional space, the 3 dimensional sophisticated camera calibrations are not needed. A simple affine transformation to convert image coordinates $(X_i, Y_i)$ to real world coordinates $(X, Y)$ was found using linear regression:

$$\begin{aligned} X &= u + v.X_i + w.Y_i \\ Y &= x + y.X_i + z.Y_i \end{aligned} \qquad \textbf{(3. 1)}$$

As an example of our experiment, for video data that was taken from above 12-15 meters from the ground, the conversion model was obtained using 151 data:

$$X = \underset{(44.3)}{66.45} + \underset{(349.1)}{1.23X_i} - \underset{(-28.9)}{0.99Y_i} \qquad R^2 = 0.999$$

and

$$Y = \underset{(151.9)}{250.87} + \underset{(17.5)}{0.067X_i} + \underset{(344.3)}{1.29Y_i} \qquad R^2 = 0.998$$

The t-statistic is given in the parentheses. The maximum error was +17 cm for $X$ and $-21$ cm for $Y$ at the edge of the image. The errors were almost zero at the center of the image. Placing the pedestrian trap in the center of the image is one strategy to reduce the error.

## 3.2 GATHERING PATH COORDINATES

As explained in the previous section, three development types of data collection can be used to collect the path coordinate from the image sequence of each pedestrian. In this section, the automation development of the data collection is presented.

### 3.2.1 Manual Data Collection

Manual data collection is simply clicking the position of the pedestrian in each frame using pedestrian-based method. To increase the speed of the data collection, a simple macro is written for the NIH Image freeware or Scion Image freeware. The coordinate location and the frame number are saved into a text file whenever the user manually clicks the position of the pedestrian. A single point in every frame can represent every pedestrian. The head of each pedestrian is taken as the representative point. If other parts of the body are going to be used, it is difficult to be consistent for all pedestrians. The height of each pedestrian is assumed the same.

To be able to distinguish the movement of every pedestrian, each pedestrian is followed frame by frame from the time he/she shows up in the image until the head is out of the image. One observation point is the position of a single pedestrian in one frame. The frame number when a pedestrian shows up is noted to ease the searching. After a pedestrian is out of image, the next pedestrian is searched.

Manual data collection is exhaustive and very costly. A single person can collect about 40-60 pedestrian paths per eight working hours. The manual works are not only expensive and exhaustive but also reduce the accuracy of the data due to manual search. The same person using the same data may obtain different results.

### 3.2.2 Semi Automatic Data Collection

To improve the accuracy and speedof manual data collection, a program of semi automatic data collection was developed. The user searches for a pedestrian, which come into the pedestrian trap, and then click the image to start the tracking. Then the user simply follows the movement of pedestrian as the pedestrian moves (the speed can be adjusted for the user's convenience). When the pedestrian is about to go out of the screen, the user simply clicks the screen again to stop the data collection. The result of the tracking is automatically recorded as points in another picture and in the matrix of the *a* TXY database.

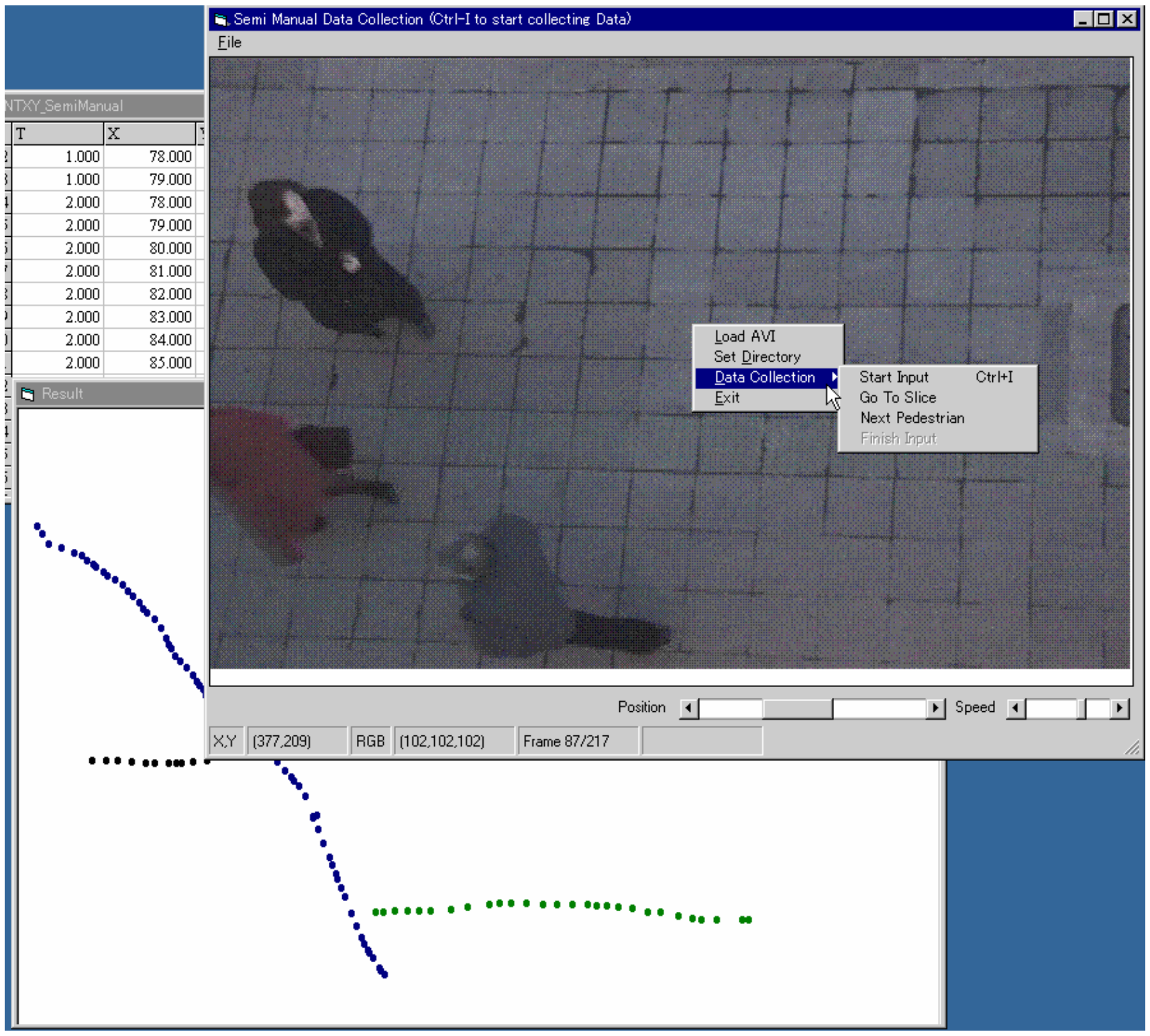

**Figure 3-2 Semi Automatic Data collection**

Whenever the user feels tired to follow the pedestrian, a click on the screen can stop the movie and the data collection. Another click will continue the movie and the tracking of the pedestrian can be done again. When the user does the right click, a popup menu will appear to support the data collection. The user can specify the frame number to go or simply using scrolls bar in the bottom. The user also needs to specify the pedestrian number of the next pedestrian.

Similar to the manual data collection, he head of each pedestrian is taken as the representative point. The choice of the head coordinate is simply to maintain the consistency of the height of the pedestrians. Again, the height of each pedestrian is assumed the same.

Both manual and semi automatic data collection needs manual tracking and the recording of the coordinate and time is automatic. Their difference lies on the way to follow the path coordinates. The semi automatic data collection will have automatic movement to the next slice without any clicking while using the manual data collection the user hardly needs to click every pedestrian to be able to move to the next slice.

When the data collection is finished, the $a$TXY database appears. The database can be edited similar to a spreadsheet program. The result is saved into a text file automatically when the $a$TXY database is closed.

Compared to the manual data collection, the semi automatic data collection is much faster, less exhaustive and more accurate. The speed of processing is about 5 to 10 times faster than the manual one. The accuracy, however, is still a problem due to manual tracking. The same person using the same data may obtain different results when it is taken in a different time.

### 3.3 AUTOMATIC DATA COLLECTION

When a large amount of data is needed, semi automatic data collection procedures become exhaustive and labor demanding as the manual one. Measurements of microscopic flow performance in the real world need to be elaborated in connection with the automation of data collection. It is desirable if the data collection can be done fully automatic without human intensive work. Such an ideal automation is still impossible at this time, but an attempt to automate the system was done. The automatic data collection that has been developed has the capability to search, track, recognize pedestrians, record the time, and coordinates automatically. The point that represents the pedestrian is mid point of the Feret Box that could represent the center of gravity of the pedestrian which is quite consistent. The height of the center of gravity of the pedestrian does not change much by their movement. This section explains about the detailed algorithm to detect the location of pedestrians for each image.

### 3.3.1 Segmentation and Object Descriptors

To detect the location of pedestrians, the pixels that picture the pedestrians must first be detected. Since in reality, pedestrians may use any color of clothes and the environment is of different colors, to ease the detection of pedestrian pixels, the background view needs to be removed. Because the image sequence comes from a static camera that is taken in a location with one focus, the background view of the image is almost at constant brightness. Only pedestrians and objects, which are moving, have changing gray levels. Based on this phenomenon, the image can be separated into background view and moving objects. Background image is a frame where the pedestrian facility (i.e. walkway) does not have any pedestrian. Background subtraction is one of the best tools for segmenting motion images from the static background. The algorithm is simple and fast compared to other methods. Manual searching traditionally initialized the background image. Automatic background image may be developed automatically from the video images when there are no objects, as explained in the Chapter 2. Simply averaging the image sequence over time will also produce the background image. Then, object images are binarized into black and white pictures. For every frame, two pictures are created. One is the original color image and the second is the binarized image. Standard binary morphological operations of closing and opening are performed to reduce the noise and increase the unity of connected components of the objects. The morphological operation has been discussed in many digital image-processing textbooks (e.g. Gonzalez and Woods (1993)). Figure 3-3 illustrates the steps of the algorithm to segment and to obtain object descriptors.

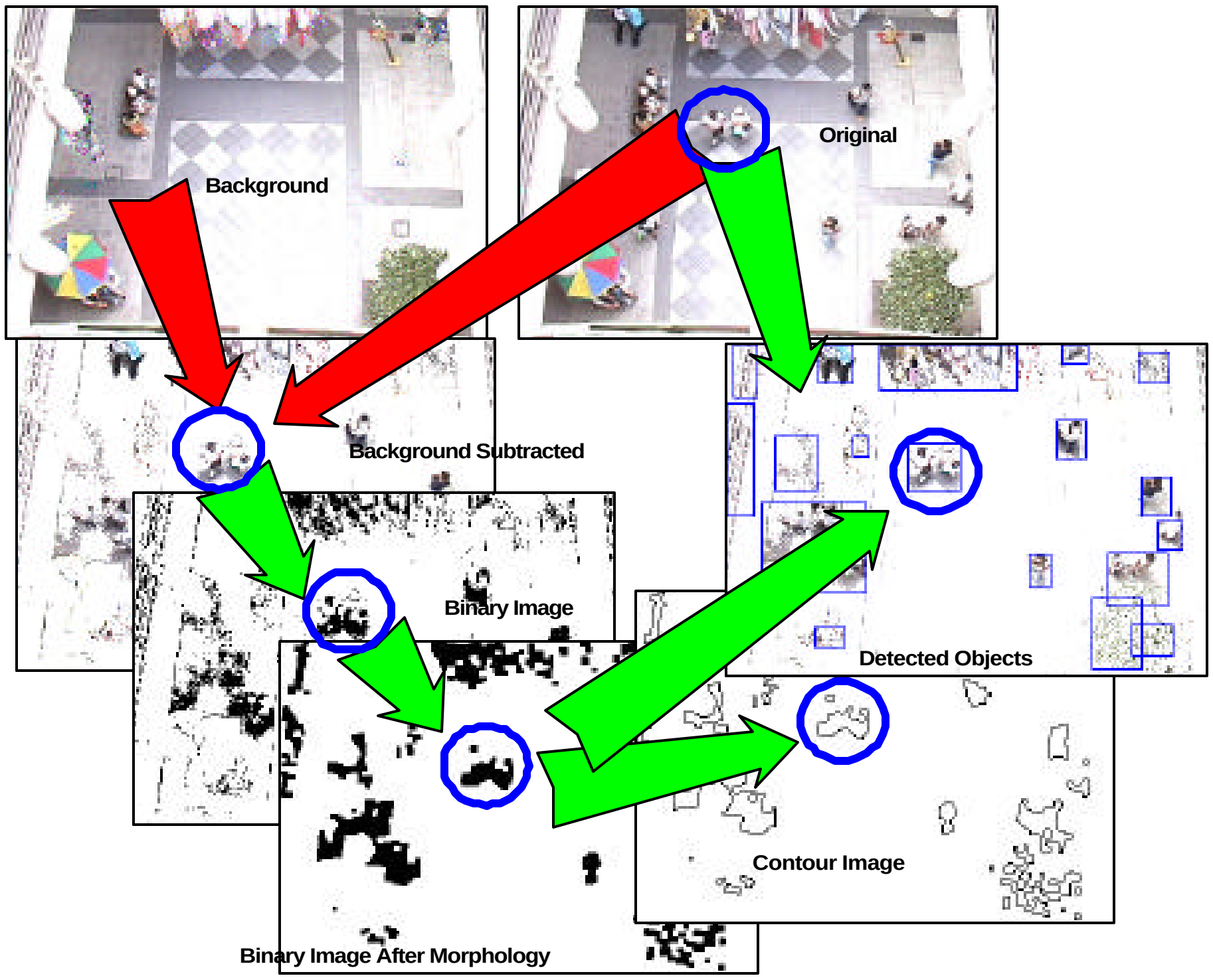

**Figure 3-3 Image processing procedures to detect the objects and calculate the features.**

After the moving objects are separated from the background, the pixels that contain the picture of pedestrians and other moving objects are detected in the binarized image. The image is scanned from the top left to bottom right. If a black pixel is found, the location of the pixel is stored and the neighboring pixels are searched. If any of the eight directions of the neighbor pixels contain a black pixel, the black pixel is called connected. All the connected black pixels may produce a region that represents a moving object. After objects in the binarized image are detected, all pixels in the corresponding location of the original color image are captured. A bounding box is drawn to mark a detected object. Based on the color and binarized pixels, object features are calculated. Statistical descriptors as well as binarized and color descriptors are calculated. The statistical descriptors are determined based on the histogram of the pixels such as mean, variance, relative smoothness, skewness, kurtosis, energy, and entropy. The binarized descriptors are calculated based on the binarized image such as area, and perimeter, ratio of box height and width. The area of an object is the number of black pixels in the region bounded by the box. Coordinate of the moving object is calculated based on the center of gravity. Each black mask of binarized image has a corresponding pixel in the color image that can be measured as color object descriptors.

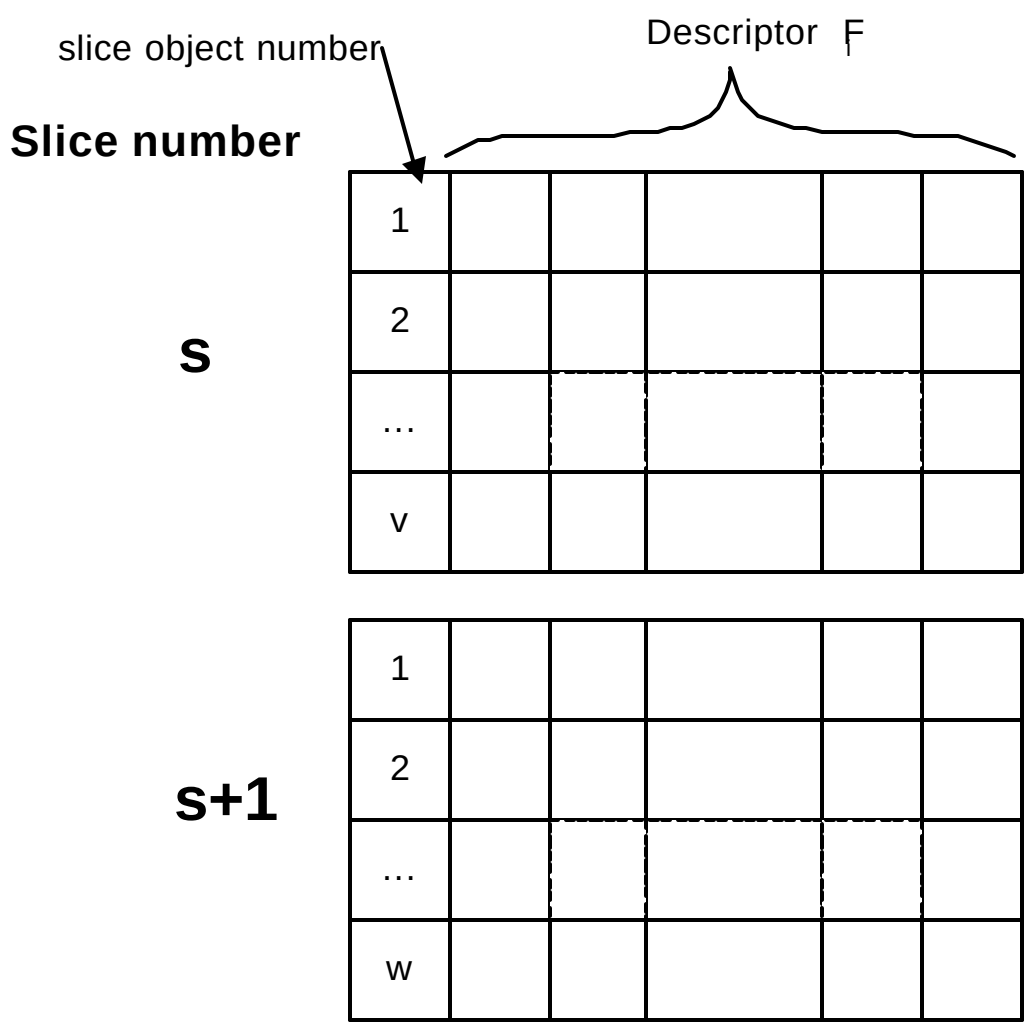

**Figure 3-4 Descriptors Database**

When the area of an object is too small for a pedestrian, it is considered as a noise and it is deleted from the data. Performing the above algorithm of background subtraction, object detection and object descriptors toward all frames of the video produce a database, called a *descriptor database*. For each slice number, the database consists of a field of slice object number, and several object descriptors. The descriptors of an object can be characterized as intensity value, speed, color, area, moment, special markers, coordinate location, gradient, distance, string code, Fourier coefficients, etc. For simplicity, however, one descriptor is a single number that describes the characteristic of the object detected. Center coordinates of an object area, for instance, is considered as two descriptors of X and Y. Slice object

**Comment:** How descriptor database is made number is an ordered number from 1 to maximum number of object detected on that slice. To each slice in a stack, a unique slice number is assigned from 1 to maximum number of frames in the stack. It is easier to explain the descriptor database if we represent it as matrix of descriptors of each slice as shown in Figure 3-4.

Each row in the descriptor database represents a moving object in a frame while the columns represent object descriptors. A unique slice-object number is assigned to each object in the descriptor database. This slice-object number does not represent the real pedestrian number since it is only an ordered number for each frame. Missing descriptors are undetected descriptors. For example, the average color for binary image is not calculated because it has no meaning. Thus, average color in this case is a missing descriptor. Any missing descriptor is coded by a minus one (-1). Each image of the pedestrian in each frame is represented by one point, which is the centroid of its area. Starting from the first frame, each point is numbered by a unique pedestrian-number. A simple tracking procedure is done to follow the movement-path coordinates of each pedestrian for every frame.

### 3.3.2 Tracking and Matching

It is assumed that there are five probable events that an object may behave in the scene, namely

1. Enter the scene
2. Go out of the scene
3. Continue in the scene
4. Merge with other objects into a single point
5. Diverge from a merging point

Tracking an object in an image sequence as feature points can be divided into two major steps: tracing and matching. The tracing step is done to search which objects and which frames in the image sequence are to be considered for matching. It is also done to mark and to follow the matched object. Matching compares the *base object* with the *matched object* based on proximity and similarity. Although the heart of tracking is the matching step, matching step alone is only able to detect the objects that are already in the scene and continue in the scene. To detect the other four probable events, a tracing algorithm should be utilized.

The principles behind the tracing algorithm to detect five probable events that an object can do in the scene are as follows:

1. If an object does not exist in frame $s$, but exists in frame $s+1$, then this object is a *newcomer* to the scene at frame $s+1$;
2. If an object exists in domain frame $s$, but does not exist in frame $s+1$ to $s+n$ consecutively, then this object goes *out* of the scene at frame $s+1$. *Maximum searching depth*, $n$, is a constant integer that is greater than one ($n>1\in Z$, Z is a set of integer). The value of the maximum searching depth is determined before hand. It represents how long occlusion is allowed;
3. There is no occlusion and an object continues in the scene if and only if an object in frame $s$ is matched with an object in frame $s+1$;
4. If an object exists in domain frame $s$ and frame $s+k$, but does not exist in frame $s+1$ to $s+k-1$, where $k$ is an integer greater than one ($1<k\leq n, k\in Z$), then we can conclude that occlusion happened in the *frames between* $s+1$ to $s+k-1$. Although merging and diverging events cannot be detected directly, we can detect the overall effect of both events in the occlusion phenomena.

Based on the above principles, the tracing algorithm is developed. In the beginning, there is no domain frame and the object pointer is assigned nothing. Although either forward or backward tracing can be used, to ease the explanation, forward tracing is used. The first frame is then utilized as the first domain frame. The first loop assigns an object number to all objects in the first frame as initialization. After the last object has an object number, the pointer then cycles back to the first object and puts this object as a base object. A searching frame is then determined among the range frames. (The determination of the searching frame is dependent on the tracing method that is used, forward tracing will operate on frame $s+1$ to $s+n$ accordingly, while backward tracing is employed on frame $s-1$ to $s-n$). Matching is then performed to search the corresponding matched object among candidate objects in the searching frame. When the matched object is found, in-between frames are calculated as the difference between the matched frame and the domain frame. If the frame in-between is one, there is no occlusion detected, else occlusion is detected.

Now if occlusion is detected, the features that represent the objects (e.g. coordinate, etc.) are interpolated based on the base object and the matched object. Those interpolated features are put in the frames between as newly created points with the same object number as the base object. If a matched object is found, the matched object is numbered as the base object regardless whether an occlusion phenomenon is detected or not. After that, the base object is marked as traced. If a matched object is not found in the searching frame, the searching frame is advanced forward (or backward) and the matching is again performed in this new searching frame. However, if the next searching frame is outside the range frames, then it is concluded that the base object has gone out of the scene and marked as traced. If all objects in the domain frame have been traced, the domain frame is advanced further. The algorithm is finished if the domain is at the end of the stack (last frame in the images sequence if using forward tracing, and first frame in the sequence if utilizing backward tracing).

Matching is based on proximity and similarity. Proximity depends on the speed of the object and distance of one object to another, while similarity relies on the object's features. Matching is done by comparing the features of a base object with all candidate objects in the searching frame. A valid candidate object can not be chosen as a matched object if it has been chosen previously unless it has the same object number as the base object. This rule avoids the representation of one feature point to more than one object. Figure 3-5 shows the flowchart of the tracing algorithm. Events detected in the algorithm are distinguished as bold letters.

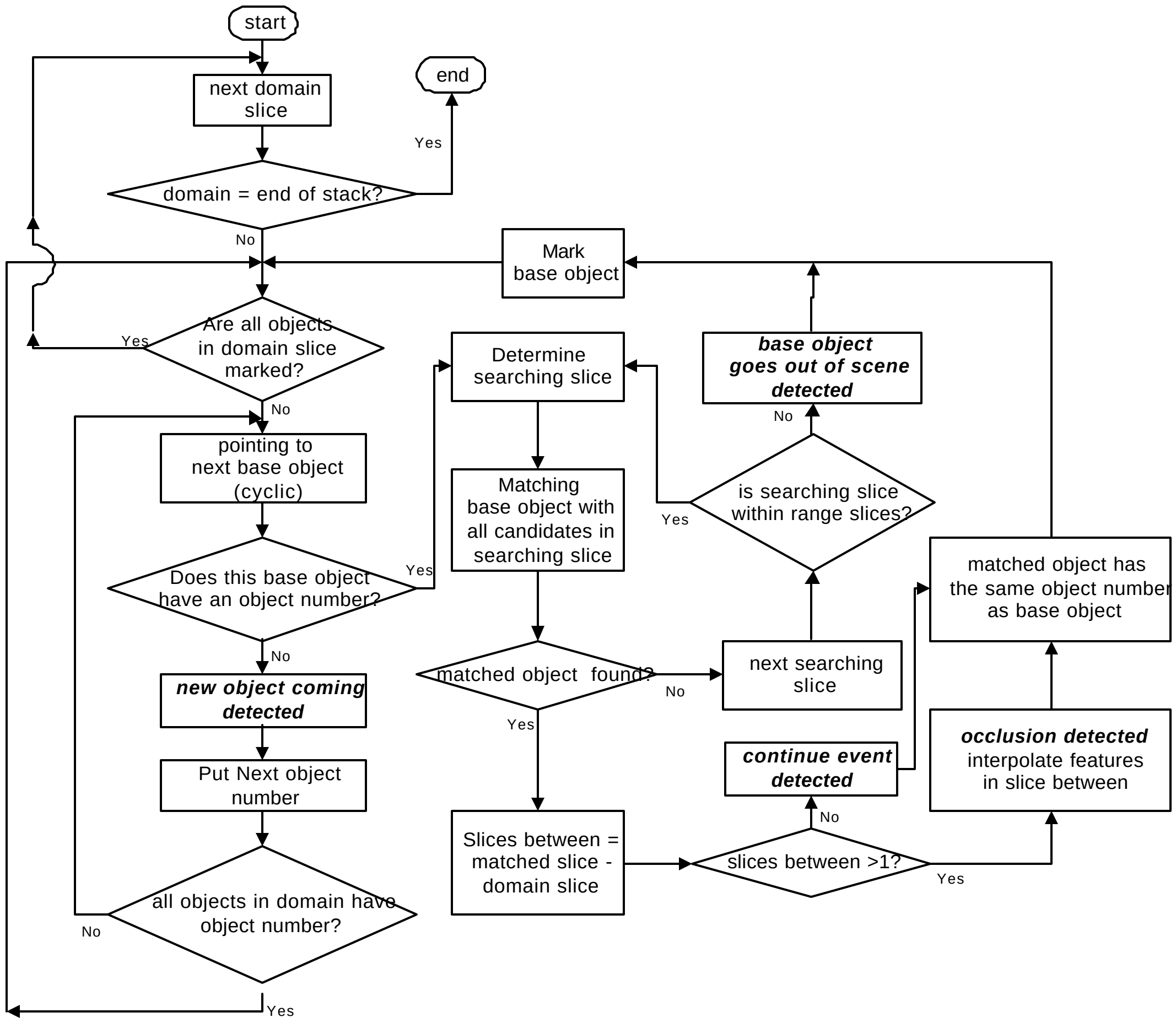

**Figure 3-5 Flowchart of the Tracing Algorithm to Detect Probable Events of Objects in the Scene.**

Since an object is assumed to be moving at a certain maximum speed limit, a speed threshold can be set to reduce the computation of similarity values. The speed threshold depends on the frame rate, camera focus, and the depth of object from the camera and the maximum speed of the object. Any candidate object that is located beyond the speed threshold should not be considered further. The number of frames in the range- frames searched increases the radius of searching for valid candidate objects linearly. The number of times a change in searching frame for a particular base object occurs is called *depth of tracing*.

The similarity measurement can be distinguished by two approaches, feature based, and pixel based. Instead of comparing pixel by pixel in the image level, this research uses feature based matching that is done after image segmentation and object detection. The tracking is performed after the image processing is done.

The absolute difference between features is taken to represent dissimilarity. The sum of squares or correlation can also be used, but the absolute difference can be computed faster and the result is close enough to the other two. The *similarity index*, $\Lambda_c$, of each candidate object is calculated as in equation (3. 2), which is the average of one minus the normalized absolute difference. The indices $i$, b and c represent the feature, base object and candidate object respectively.

$$\Lambda_c = \frac{\sum_i \left( 1 - \frac{| F^{s}_{b,i} - F^{s+k}_{c,i} |}{F^{s}_{b,i} + F^{s+k}_{c,i}} \right)}{\eta} \qquad \textbf{(3. 2)}$$

where

$h$ is the number of features;

F denotes the feature;

$s$ represents the domain frame number; and

$s+k$ denotes the searching frame.

The value of $k$ is between $1 \le k \le n, k \in Z$ for forward tracing and $-n \le k \le -1$ for backward tracing. As defined earlier, $n$ is the maximum searching depth.

The greater similarity index of a candidate object, $\Lambda_c$, the greater the chance that this candidate object is the matched object. The similarity index has a value from 0 to 1. If the denominator of the normalized absolute difference is zero, then the similarity index is undefined and that feature is not considered within the average. To separate between the matched object and the object, which goes out from the scene, a similarity threshold needs to be determined. The threshold is important since without it, there is always a candidate object that gets a maximum similarity-index. The determination of the similarity threshold can be eased by the characteristic of the similarity index. Figure 3-6 shows the nature of the similarity index against the features proportion. Suppose two feature values are A and B respectively, when the proportion of A/B (or B/A) is smaller than one, the similarity index goes up as the proportion moves from 0 to 1. The similarity index reaches the value of one when both features are an exact match. As the proportion gets higher than one, the similarity index slowly goes down and reaches zero when the features proportion reach infinity. The similarity threshold is then determined based on how much the maximum and minimum proportion is allowed for the particular feature.

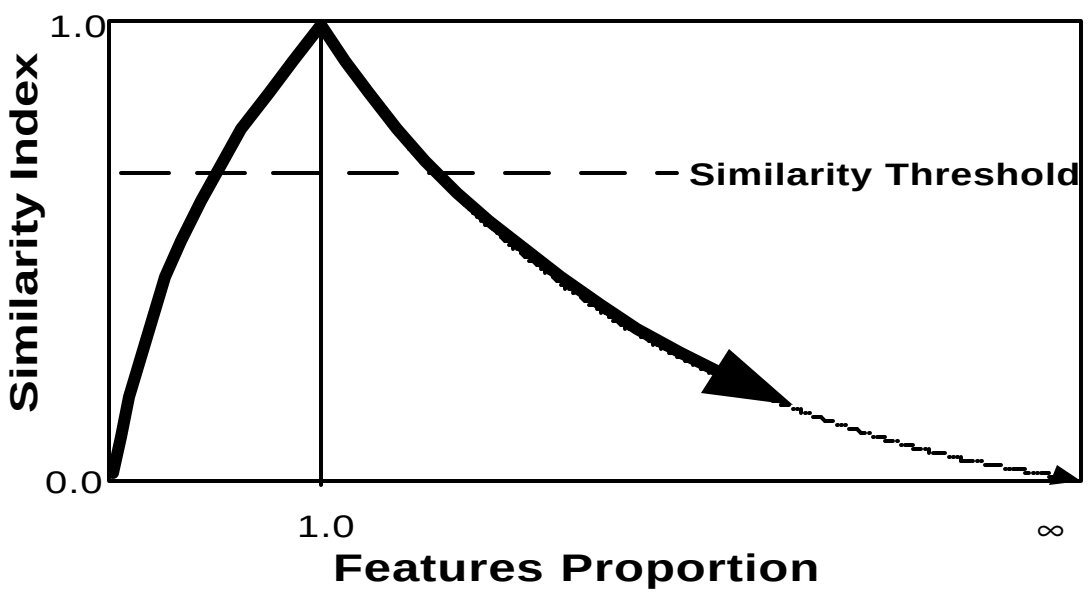

**Figure 3-6 The characteristic of similarity index which facilitates the determination of similarity threshold as features proportion.**

Thus, a matched object is found in the searching frame when:

1. There is a candidate object that has already the same object number as the base object;
2. There is a candidate object that passes all the following four criteria in order:
   a. The candidate has no object number (it has not been chosen previously);
   b. The distance between the candidate object and base object times the depth of

tracing is lower than the distance threshold;

c. The similarity index is higher than the similarity threshold;

d. The similarity index is the maximum among all candidates.

If all candidate objects in a searching frame fail to pass any of the criteria above, then no matched object is found in the searching frame.

### 3.3.3 Pedestrian Recognition

To distinguish pedestrians from other moving objects, the recognition procedure is applied. Recognition is performed based on the assumption that other moving objects are just noises and small movement objects' (e.g. paper decorations, leaves of the trees) movement are not continuous. Checking the continuity of the motion of each object distinguish pedestrians from the other objects. Deleting all non-pedestrian rows and renumbering the pedestrian numbers provide the right database. Following the recognition, calibration of image coordinates to real world coordinates is performed using linear regression. The calibration is achieved by taking several points in the field and its corresponding points in the images. An $a$TXY database, which consists of fields of pedestrian number, coordinate location, and time, is finally produced from these data collection procedures.

To recognize the pedestrian, a motion model is developed. The data resulting from tracking is assumed to contain only pedestrians, other static object and noise. We want to separate a pedestrian who is moving with other objects (such as trees) and noise.

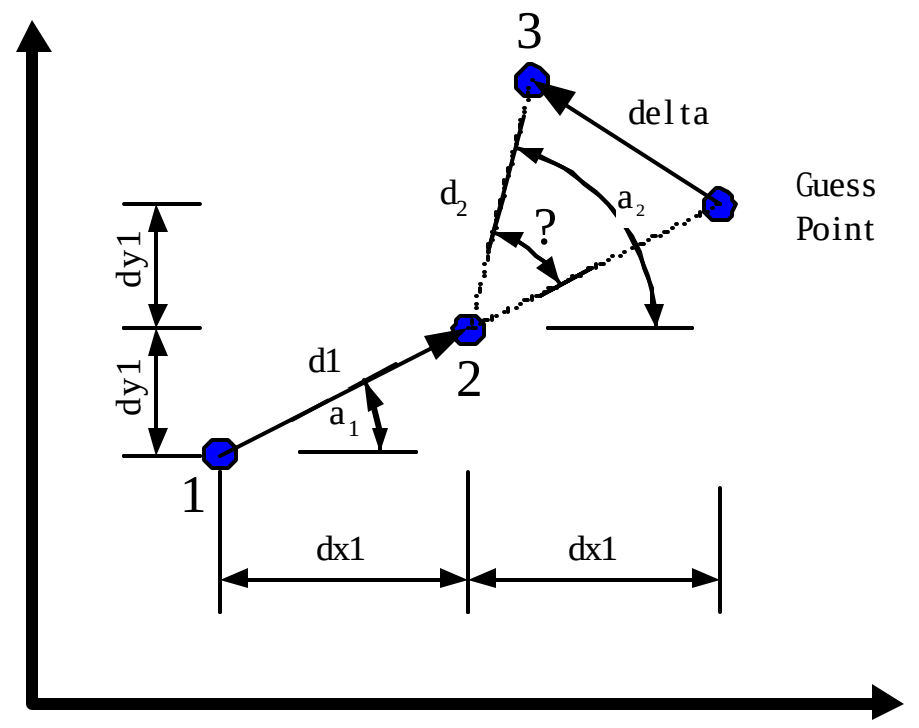

**Figure 3-7 Motion Model**

The motion index is a measurement of distance and angle to detect the location of the next point from a guess point. The number is between 0 and 1 where unity means the next point located in the exact location as the guess point. Figure 3-7 shows a trajectory of several points in time of a pedestrian. Let us name point at time t-1, t and t+1 as point number 1, 2, 3 respectively. Based on point 1 and 2, a guess point g is decided as extension of line 1-2. Let $\beta$ be defined as an absolute angle between the extension line 2-g and line 2-3, $\delta$ as distance from guess point g to point 3, $d_1$ as length of line segment 1-2, and $d_2$ as length of line segment 2-3. The distances can be calculated as $\delta = \sqrt{(x_3 - x_g)^2 + (y_3 - y_g)^2}$ and $d_2 = \sqrt{(x_3 - x_2)^2 + (y_3 - y_2)^2}$ . The angle can be calculated as $\cos a_i = \frac{dx_i}{d_i}$ and $\beta = |a_2 - a_1|, \quad 0 \le \beta < 180^\circ$. Then, motion index, P is determined as

$$P = \begin{cases} 0 & if\ d_1\ or\ d_2 = 0 \\ e^{-(C1.\frac{d}{d_1} + C2.\frac{d}{d_2}\tan(\frac{b}{2}))} & otherwise \end{cases} \qquad (3.3)$$

The motion index is formulated based on the criteria:

1. The value of the motion index is between zero to one;
2. The distance is taken as ratio with its previous distance (divided by $d_1$);

3. When angle $\beta = 0$, the motion index will vary according to the variation of the distances;
4. If the distance $\delta = 0$, any angle $\beta$ will produce the same motion index as one;
5. If the distance $d_2 = 0$, the motion index will be undefined or defined as zero because the object does not move.

The constant C1 is a scale factor. It is determined by setting an allowable error for $\delta = 2, d_2 = 1$ and $\beta = 0$. The variation of the allowable error will change the first point ($\beta = 0$) of the graph. For our purpose, C1's allowable error of 5% is reasonable to make $P = 0.95$ when the founded point 3 is in double radius from the guess point with the same angle. Thus, C1 is estimated as $C1 = -\frac{(\ln(0.95)}{2} = 0.02565$

The constant C2 is a shape factor. It is established after C1 is determined by setting an allowable error for $\delta = 2, d_2 = 1$ and $\beta = 90°$. The variation of the allowable error for C2 may change the shape of the graph. A higher allowable error for C2 is leading to a more sensitive motion index. When point 3 is found in the right angle of the guess point within double radius and the error that we allow is 15%, C2 can be calculated as $C2 = \frac{-d_2(\ln(0.85) + C1.\delta)}{\delta.\tan(\frac{\beta}{2})} = 0.055613$.

Then the recognition algorithm is as follows:

**Algorithm 3.2:**

1. Sort the result of tracking by pedestrian number by time (slice number);
2. Calculate the start and end of each pedestrian number;
3. If the number of the point (row) of one pedestrian is smaller than 5, delete that pedestrian number as we consider them as noise. Remember that we are not interested with a of pedestrian who just enter or just go out;
4. Calculate the motion index for every three consecutive points and then average them (sum and divided by number of rows minus two). If the average motion index is smaller than a threshold (say, 0.7), delete this pedestrian number, because they are not pedestrians but trees or other objects.

5. Renumbering the rest of pedestrian number and make graph of their path.

## 3.4 MICROSCOPIC DATA COLLECTION RESULTS

The results of the tracking are compared with manual tracking to examine in which events the mismatch happen. The manual tracking data is considered ground truth data. When false tracking happens, mismatched objects will always be traced as wrong object numbers. A simple experiment like the collision of two balls under full control of the light in the laboratory could be detected and traced with an error of less than 15%. While testing with a real scene for pedestrians, it could detect the new pedestrian coming and going out and full occlusion (merging and diverging) in very short frames ( $n < 3$ ).

The tracing was performed correctly to detect new objects coming even after there is a long gap of empty object in the scene. A false positive (additional number of objects) happens merely due to partial occlusion. A merging event only is falsely traced as an object going out while diverging only is falsely traced as a new coming object. A false negative happens when several pedestrians are always together throughout the scene and because of a merging event; this is always detected as one object. The errors mostly happen due to matching and unrecoverable event by changing of some thresholds. The algorithm was useful when there are many objects, which enter and go out from the scene during the video taking. Partial occlusion (diverging or merging only) events still cannot be detected properly. With this result, only individual pedestrian data can be gathered by the automatic method with quite large error (25-50%) due to matching problem.

Since the automatic path coordinate data collection is not so reliable yet, the semi automatic data collection is the resource of the data collection procedure in this dissertation. Figure 3-8 shows the results of the data collection. The movements of three pedestrians are shown including their speed profiles. Since pedestrians are moving in two dimension, the trajectory can also be projected into X and Y direction. The top left figure (Figure 3-8 (i)) shows the movement trajectory of three pedestrians. It is the path of the pedestrians in the walkway. The black rectangular is the pedestrian trap. The letters (A, B, C) are given in the figure to identify the pedestrians. The arrow represents the direction of the pedestrians. Though the path or the movement trajectory of pedestrian A is crossing B and C, they do not collide with each other because the paths cross at different time.

The Y-trajectory (Figure 3-8 (ii)) and the X-trajectory (Figure 3-8 (iii)) project the movement trajectory into Y-axis and X-axis respectively, over time. The crossing points bet ween pedestrian in these trajectory have no meaning, but the gradient of these trajectory represents the velocity in their respective direction. The gradient of pedestrian B in the Y trajectory signifies the Y velocity component of pedestrian B. The X and Y trajectory also represent the movement direction of the pedestrian. The pedestrian moves from South to North when the Y trajectory of pedestrian increases as the time increases, (the pedestrian trap is facing Y positive as North and X positive as East). On the other hand, the pedestrian move from North to South, if the Y trajectory of pedestrian decreases as the time increases (slope downward). Similarly, if the X trajectory of a pedestrian has positive slope, the pedestrian moves from West to East and if the X trajectory of a pedestrian has a negative slope, the pedestrian moves from East to West.

Figure 3-8(iv) illustrates the speed profiles of the three pedestrians. The speed is calculated as the square root of the square of the first difference of the velocity component. Suppose $x(a,t)$ and $y(a,t)$ are the coordinates of pedestrian $a$ at time $t$ from the $a$TXY database. Then the speed is calculated as

$$v(a,t) = \sqrt{\left(x(a,t+1) - x(a,t)\right)^2 + \left(y(a,t+1) - y(a,t)\right)^2} \tag{3.4}$$

Equation (3. 4) explains why the time in Figure 3-8(iv) lacks one-second compared to the trajectory.

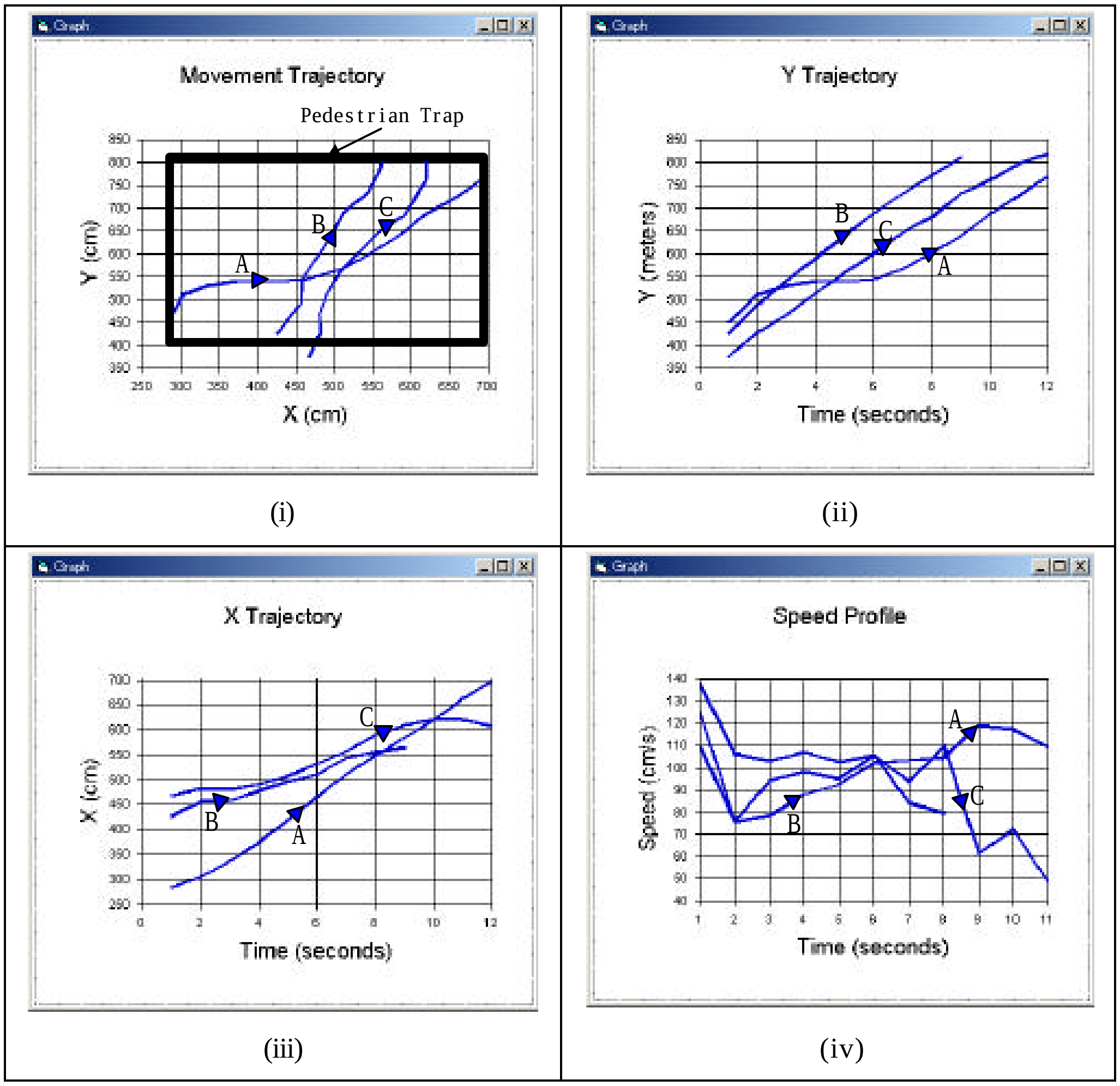

**Figure 3-8 Movement Trajectories in X and Y direction, and the speed profile**

The real world video data was taken several days on August 2000 in Higashi Nibancho Dori, Sendai, Japan, from the 5th floor of JTB parking building, as seen in Figure 3-9. It is a intersection crossing data. Each set of data is about 60 seconds (one green time) with about 150 pedestrians involved. The pedestrian trap is set as the whole crossing area (11.23 meter by 31.10 meter).

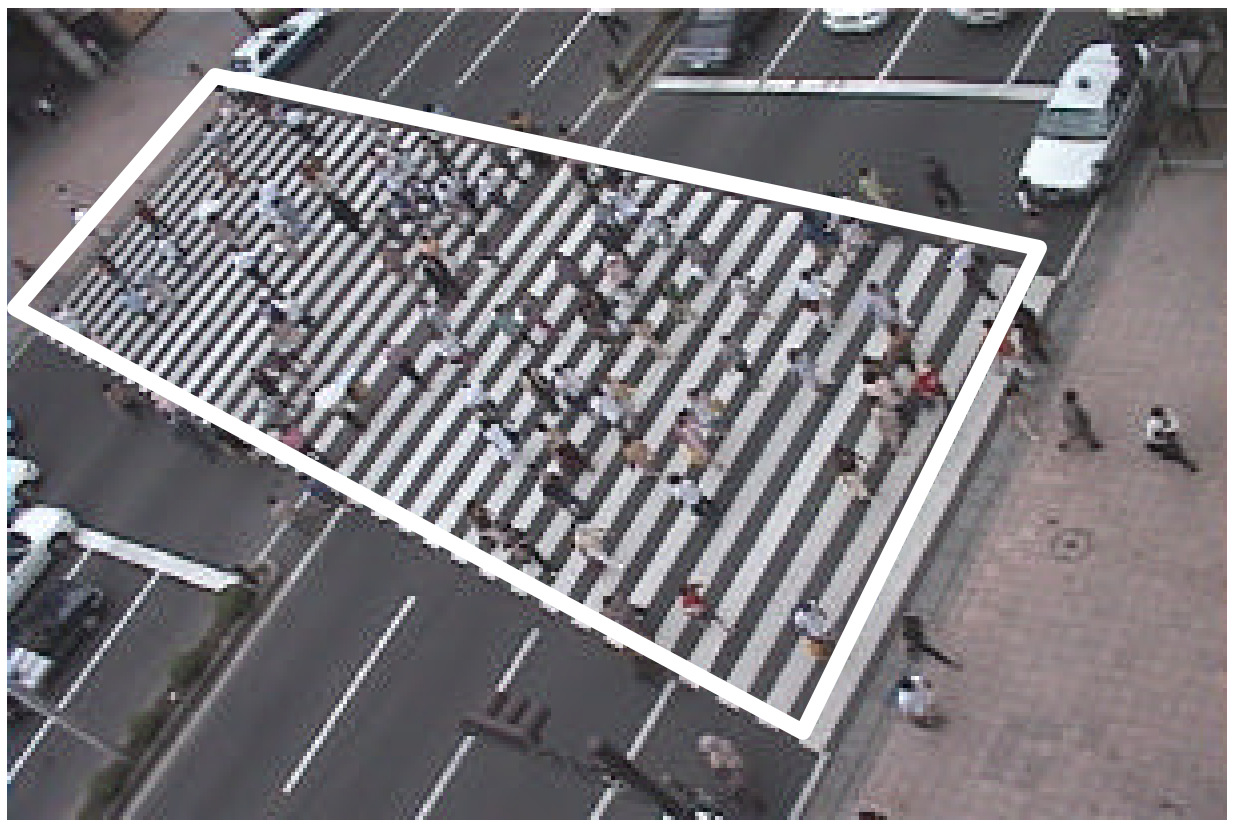

**Figure 3-9 Real World Pedestrian Trap**

After the conversion from video into file and the collection of path coordinates ($a$TXY data was gathered using the semi automatic data collection system), the data was trimmed into pedestrian trap only. The data was taken every 0.5 second (2Hz). More slices in every second (higher frequency) do not produce much movement of pedestrian due to the size of the picture. Smaller frequency will produce a rough behavior of pedestrians movement. The conversion image coordinate ($X_i, Y_i$) to the real world coordinates ($X, Y$) was found using 136 data to be

$$X = \underset{(95.14)}{36.24} - \underset{(-54.89)}{0.05X_i} - \underset{(-27.41)}{0.036Y_i} \qquad R^2 = 0.966$$

and

$$Y = \underset{(7.89)}{1.759} - \underset{(-30.15)}{0.016X_i} - \underset{(-57.41)}{0.043Y_i} \qquad R^2 = 0.969$$

The t-statistic is given in the parentheses.

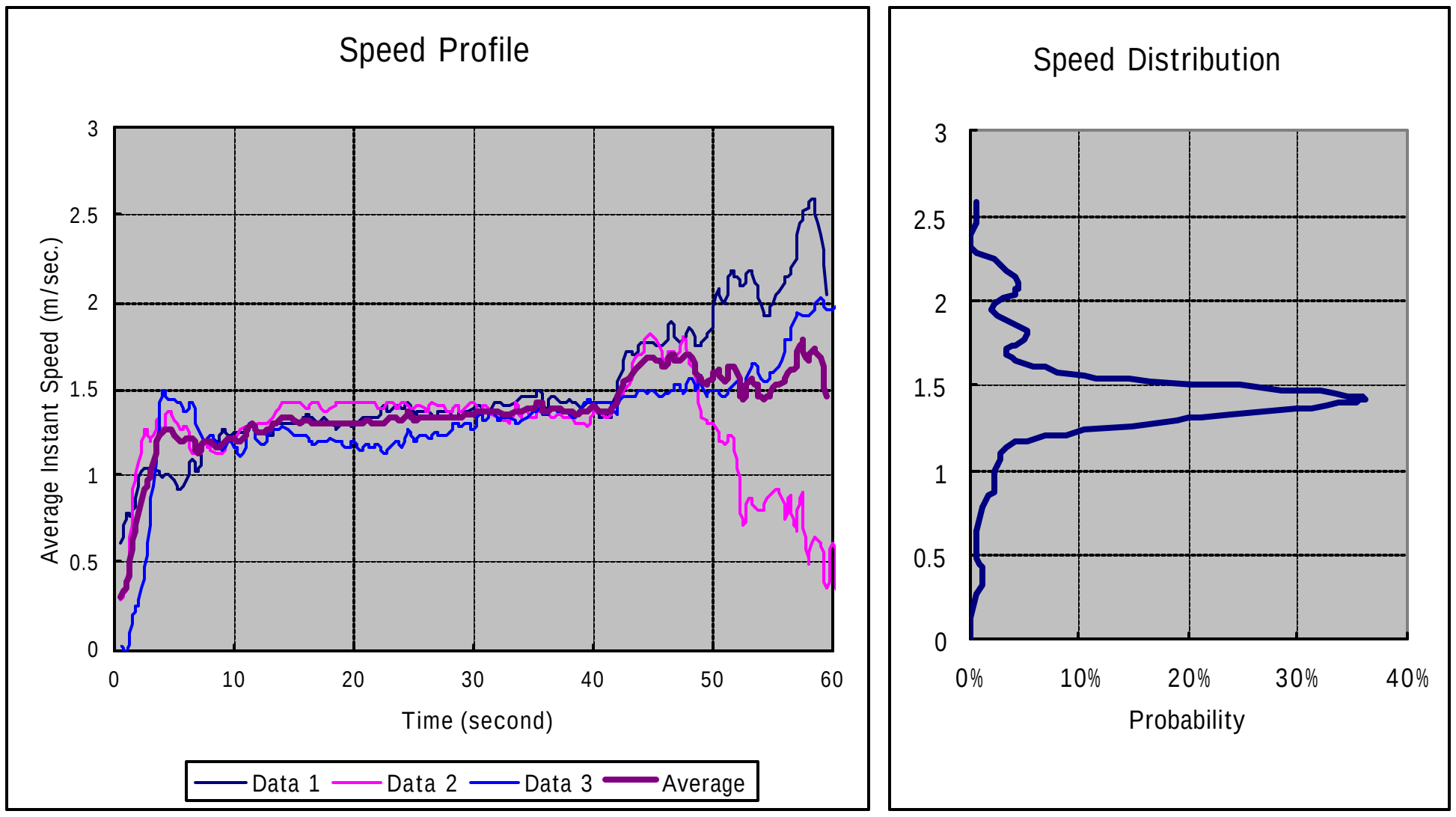

**Figure 3-10 Speed Profile and Distribution**

A total of 27187 speed data and 26668 acceleration data were analyzed. Figure 3-10 shows the speed profile and speed distribution of these data. In the beginning, the speed is slow and going up to the average speed. About the last 10 seconds before the end of the green time, the speed profile of the real world pedestrian crossing is erratic due to the intention to be fast (blinking signal) and some of the pedestrian begin to run. The speed distribution nearly resembles a normal distribution with an average of 1.38 meter/second.

The density profile, in Figure 3-11, shows that the number of pedestrians increase up to the maximum total number of pedestrians and then slows down because some of the pedestrians had already reached the other side of the road. Unfortunately, the density distribution does not resemble any theoretical distribution.

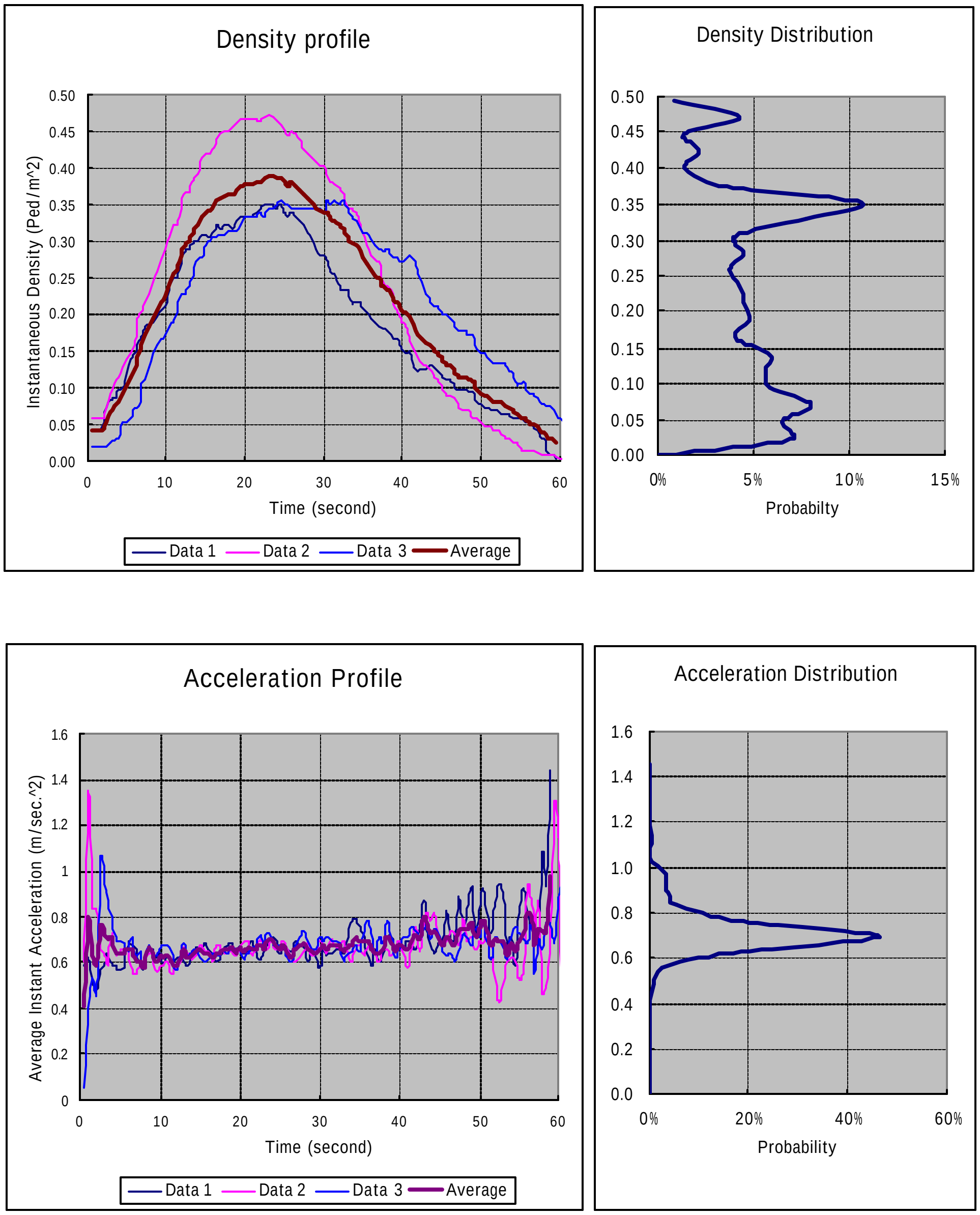

**Figure 3-11 Density and Acceleration Profile and Distribution**

The acceleration profile, in Figure 3-11, exemplifies that the acceleration is almost constant. In the beginning of the crossing time, the acceleration tends to be high because the pedestrians start from zero speed and tend to increase their speed into the average speed. In the end of the crossing time, the acceleration again becomes high because some pedestrians are running. The acceleration distribution bears a resemblance to the normal distribution with an average of 0.68 and maximum acceleration of 1.44 m per square second. Table 3-1 shows the statistics of average performances from these real world data. Some values of the tables, such as the speed and acceleration will be used for the validation of the simulation in Chapter 5.

**Table 3-1 Statistics Real World Average Performances**

| Statistics | Mean | Median | Mode | Std. Dev. | Range | Maximum | Count |
|---|---|---|---|---|---|---|---|
| Average Instantaneous Speed Data (m/sec) | 1.38 | 1.37 | 1.25 | 0.37 | 2.59 | 2.59 | 375 |
| Instantaneous Density data (ped/m^2) | 0.21 | 0.20 | 0.34 | 0.13 | 0.47 | 0.47 | 375 |
| Average Instantaneous Delay data (sec/ped) | 7.80 | 8.21 | 9.88 | 4.78 | 32.93 | 32.93 | 375 |
| Average Instantaneous Uncomfortability data | 7.39% | 7.38% | 8.77% | 4.25% | 29.24% | 29.24% | 375 |
| Average Instantaneous Acceleration data (m/sec^2) | 0.68 | 0.66 | 0.69 | 0.12 | 1.39 | 1.44 | 372 |

Total number of pedestrians = 519

# CHAPTER 4 DEVELOPMENT OF A MICROSCOPIC PEDESTRIAN SIMULATION MODEL

Microscopic Pedestrian Simulation Model (MPSM) is a computer simulation model of pedestrian movement where every pedestrian in the model is treated as an individual. This chapter describes the development of a new microscopic pedestrian simulation model (MPSM) that improves on the existing MPSM. The development of a new MPSM is needed to improve the existing simulation model and to provide the tool to determine the microscopic characteristics of pedestrian flow.

The proposed microscopic pedestrian simulation model explained in this chapter is a physical force based model similar to the social force model with forward and repulsion forces as the main force driver. The detail of the model, however, is somewhat different since it does not require target time as an input to the model. Exactly the opposite of the fact, the dissipation time is the output of the model, similar to the queuing network models. Instead of using arbitrary scores as in the magnetic force model or benefit cost cellular model, the proposed model uses the physical based variables that can be measured. The repulsive force is also added to guarantee collision avoidance. The collision avoidance algorithm is influenced by Reynolds [1],[2] which uses the steering behavior to animate birds. Thus, the proposed MPSM is developed based on the existing models to improve the deficiency of the existing models without withdrawing their main advantages.

Pedestrians in the microscopic simulation model are modeled as autonomous objects to be seen from above of the facilities. A pedestrian is modeled as a circle with a certain radius (uniform for all pedestrians). Each pedestrian is own initial location, initial time, and initial velocity, and predetermined target location (opposite to the initial location). These inputs can be determine by the user as a design experiment or specified randomly.

## 4.1 MODELING A PEDESTRIAN AND THE WALKWAY

A pedestrian is modeled as a circle as the pedestrian moves in a facility as seen from the top of the facility. The diameter of the circle represents the body of the pedestrian, which is about 50-90 cm. A circle has more benefit to model the pedestrian compared to other shapes (i.e. square, rectangular or polygon), because a circle can be represented, as a single point of its center assuming the radius of the circle is constant. The representation of a pedestrian into a single point in time is greatly simplified and it increases the computational speed. Another benefit of using a circle is revealed when two pedestrians are close to each other and collide or overlap. The collision detection as the most time consuming calculation is cut down into comparison of distance between two points and summation of the two radiuses.

The facility (i.e. walkway) is modeled as a two-dimensional space where the pedestrian is moving around. A part of pedestrian facility that the pedestrians can be seen is called a pedestrian trap.

Each pedestrian has origin and destination place. The origin and the destination of the pedestrian are always within the pedestrian trap, when a pedestrian is generated by manual operation,. When a pedestrian is generated automatically by the program, the pedestrian is always generated and attracted in a pedestrian generator. A *pedestrian generator* is a rectangular area located outside the pedestrian trap where the pedestrians appear in some point of time. The origin point of the pedestrian is inside a pedestrian generator while the destination point is in another pedestrian generator. The location of the origin point inside the pedestrian generator is determined using uniform random distribution. When a pedestrian reaches the location near the destination point, the pedestrian will disappear.

Figure 4-1 shows the pedestrian crossing facility. Pedestrians who go from right to left are modeled as black circles, generated randomly within the rectangular area of the right pedestrian generator and attracted to the left pedestrian generator. Pedestrians who move from left to right are modeled as white circles, generated from a point generator in the left and move toward the right pedestrian generator. The locations of the pedestrian generators are at an equal distance from the pedestrian trap.

**Comment:** Example of the terminology: generator, trap and coordinate system

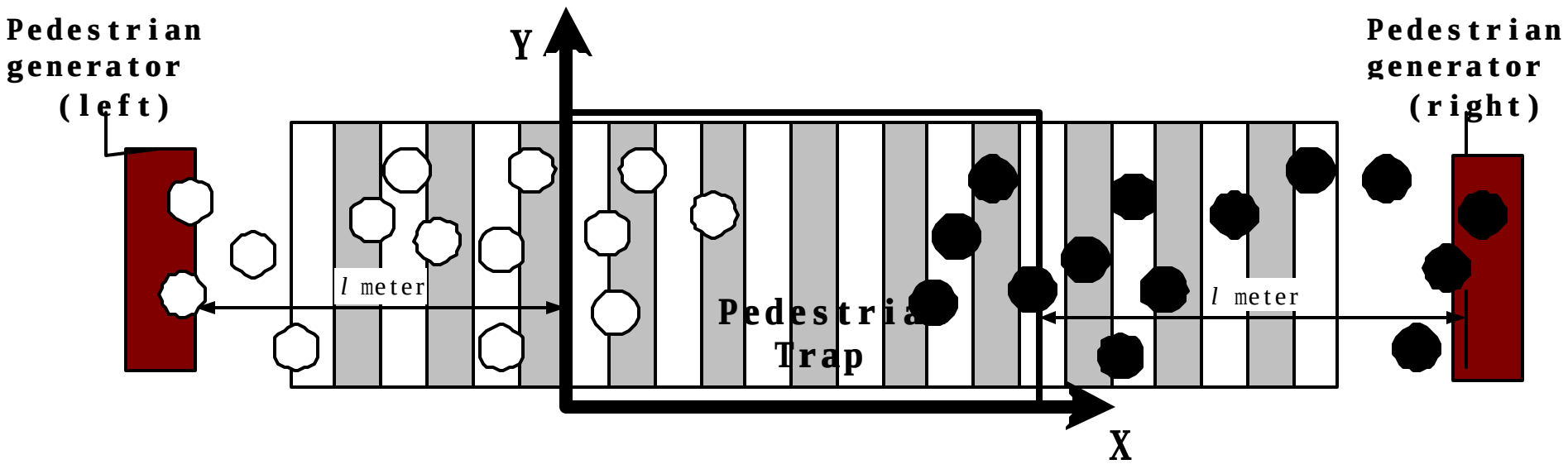

**Figure 4-1 Pedestrian Generators**

When the pedestrians are generated automatically by the program, all pedestrians are generated at the same time before the simulation. The locations of origin-points are generated randomly (uniform or normal distribution spatially) within the pedestrian generator without overlapping each other.

The coordinate system that is used for the simulation is an Euclidean coordinate system with origin in the left bottom of the pedestrian trap. The X-axis is in the bottom and Y-axis is in the left, as shown in the figure above.

## 4.2 MODELING PEDESTRIAN MOVEMENTS

Pedestrian movement shows the characteristics of a pedestrian. Compared to cars or other vehicles such as a bicycle, a pedestrian has more flexibility to move in two dimensions, a pedestrian also has more flexibility to stop and go and a pedestrian also has the smallest velocity and the highest fluctuation in acceleration and velocity compared to vehicles. However, since the attempt is not to model the step pace of pedestrians but the collective behavior of the pedestrians, such high fluctuation in acceleration and velocity are not modeled explicitly. On the other hand the pedestrian motion is modeled as quite smooth motion with high flexibility to move in the two dimensional space.

It is desirable that pedestrian motion is modeled as physical based with explicit visual interaction. The explicit visual interaction might present the validity of the model toward the collective behavior of the pedestrians.

Modeling the pedestrian movement starts with assumption that pedestrian motion is smooth (twice differentiable) if it is seen from the top view. It is also assumed that each pedestrian is subject to "forces" that represent motivation to move ahead toward the target location. The force in here is not the real physical force that has dimension of Newton (kg m/second$^2$) but only the analogy of the force that characterizes the internal driving force or motivation of the pedestrian. The force is assumed proportional with the discrepancy between the summation of intended velocities, $\hat{\mathbf{v}}(t)$ and the actual current velocity, $\mathbf{v}(t)$. Thus, the forces have dimension of meter/second$^2$.

$$\mathbf{f}(t) = m'\mathbf{a}(t) = m'\frac{d\mathbf{v}(t)}{dt}$$

$$m'\frac{d\mathbf{v}(t)}{dt} = \frac{\sum \hat{\mathbf{v}}(t) - \mathbf{v}(t)}{f}$$

$$m\frac{d\mathbf{v}(t)}{dt} = \sum \hat{\mathbf{v}}(t) - \mathbf{v}(t) \tag{4.1}$$

The intended velocity is a kind of predicted velocity on which way the pedestrian is going to move in the next time ahead. The intended velocity directs the acceleration and the forward force toward the target point[1]. The direction of the intended velocity must be the same as the force and the acceleration. Adopting the Newton law, the acceleration is proportional to the force with a constant proportion called mass, $m$. Since the acceleration has the same direction as the force, it is also the direction of the acceleration. When the intended velocity is equal to the current velocity, the force (and the acceleration) has zero value and the pedestrian may be stopped or walking with constant velocity.

Unlike the real physical dimension, the parameter mass, $m$, has a time dimension and measured in second. This unusual dimension happens due to the simplification of the model $m'\frac{d\mathbf{v}(t)}{dt} = \frac{\sum \hat{\mathbf{v}}(t) - \mathbf{v}(t)}{f}$ where the constant $f$ has the time dimension while the mass, $m'$, is dimensionless. However since both $m'$ and $f$ are constant, they can be put together as one parameter $m = m'f$. The parameter mass relates the speed and the acceleration. Since the direction of the speed difference is also the direction of the acceleration, the mass must be positive. If the value of the mass is between zero and one, the effect of the speed discrepancy toward the acceleration will be amplified. If the value of mass is bigger than one, speed difference will have lesser effect toward the acceleration.

### 4.2.1 Modeling Pedestrian Forward Force

In the absence of other pedestrians, the movement of a pedestrian should be directed from the current location $\mathbf{p}(t)$ toward the destination point, $\mathbf{e}(t)$. There should be a "forward force" that directs the pedestrian to move. The forward force makes the pedestrian path almost in a straight line. The direction of the forward force is from the current location toward the destination. The unit direction or gradient of the force is given by $\mathbf{g}(t)$,

$$\mathbf{g}(t) = \frac{\mathbf{e}(t) - \mathbf{p}(t)}{\left\|\mathbf{e}(t) - \mathbf{p}(t)\right\|} \qquad \textbf{(4.2)}$$

[1] The target point is not necessarily the destination point

The magnitude of the forward force should be determined such that, in no existence of other forces, the movement of the pedestrian is approaching a constant that represents the maximum walking speed of a pedestrian, $m_{max}$. The clause 'no existence of other forces' means that there are no other pedestrians or obstructions.

The magnitude of the forward intended velocity is desired to be similar to the maximum of the walking speed $m_{max}$ or smaller ($0 \le \hat{\mathbf{v}}(t) \le m_{max}$). If there is no other pedestrian or obstruction, the magnitude of the intended velocity must be equal to the maximum walking speed, $m_{max}$. The existence of other pedestrians or obstructions will give other forces reduce the walking speed. Thus, the intended velocity for the forward force is given by

$$\hat{\mathbf{v}}_f(t) = \frac{m_{max}}{a}\mathbf{g}(t) = \frac{m_{max}}{a}\frac{(\mathbf{e}(t) - \mathbf{p}(t))}{\|\mathbf{e}(t) - \mathbf{p}(t)\|} \qquad \textbf{(4.3)}$$

A dimensionless parameter alpha, $a$, is given to generalize the model. Since the direction of the intended velocity must be the same as $\mathbf{g}(t)$, then the value of parameter $a$ must be positive. If the parameter $0 < a < 1$, the intended velocity is bigger than the maximum speed, while if $a > 1$, the intended velocity is smaller than the maximum speed. The norm in the denominator of equation (4. 3) represents the distance between the current position and the destination.

Once the origin and destination points are set, the direction of the forward acceleration is determined by equation (4. 2). The magnitude of the acceleration is desired to make the speed to be constant as much as possible, when there are no other pedestrians or obstructions. The magnitude of the forward acceleration is able to make the speed of the pedestrian within the range of the walking speed, which is from zero to maximum of the walking speed $m_{max}$.

### 4.2.2 Modeling Pedestrian Repulsive Forces

When other pedestrians exist, the repulsive force, as the interaction with other pedestrians, is inputted to the autonomous system. The pedestrian is then optimizing the movement by taking the best path to go to the target location while avoiding other pedestrians. Two kinds of repulsive forces are working together with the forward force. One force is driving away the pedestrian actor while still quite far from other closest pedestrian, the other force strongly repulses against all other pedestrians in the surrounding.

The first repulsive force models the overtaking and meeting behavior of pedestrians. When two pedestrians meet each other, they usually move away from each other within a certain distance that is quite far from each other. They do not wait until their distance become too close and move away unless there are many pedestrians surrounding that give them no opportunity to move away. A similar behavior happens when a pedestrian is following another slower pedestrian. The faster pedestrian will move away in quite far distance if there is no other pedestrian. In the case where there are many pedestrians, the opportunity to move away from a certain distance is hindered by the lack of space and the pedestrian will either be slowing down, stop or look for another way in the emptier space.

The first repulsive force, $\mathbf{f}_a$, is working only if there is another pedestrian in front of the actor (within the sight distance). If there are many other pedestrians, it considers only the closest pedestrian to the actor because it is the closest pedestrian who will affect the decision to move away. If we sum up the forces generated from other nearest pedestrians, the first repulsive force will be unstable. It will take up/right direction in one step and down/left direction in the next step or reverse direction. In other words, the behavior of the pedestrian becomes erratic due to many considerations that must be taken at one time. Thus, the force is directing the actor away from that closest pedestrian as shown in Figure 4-2.

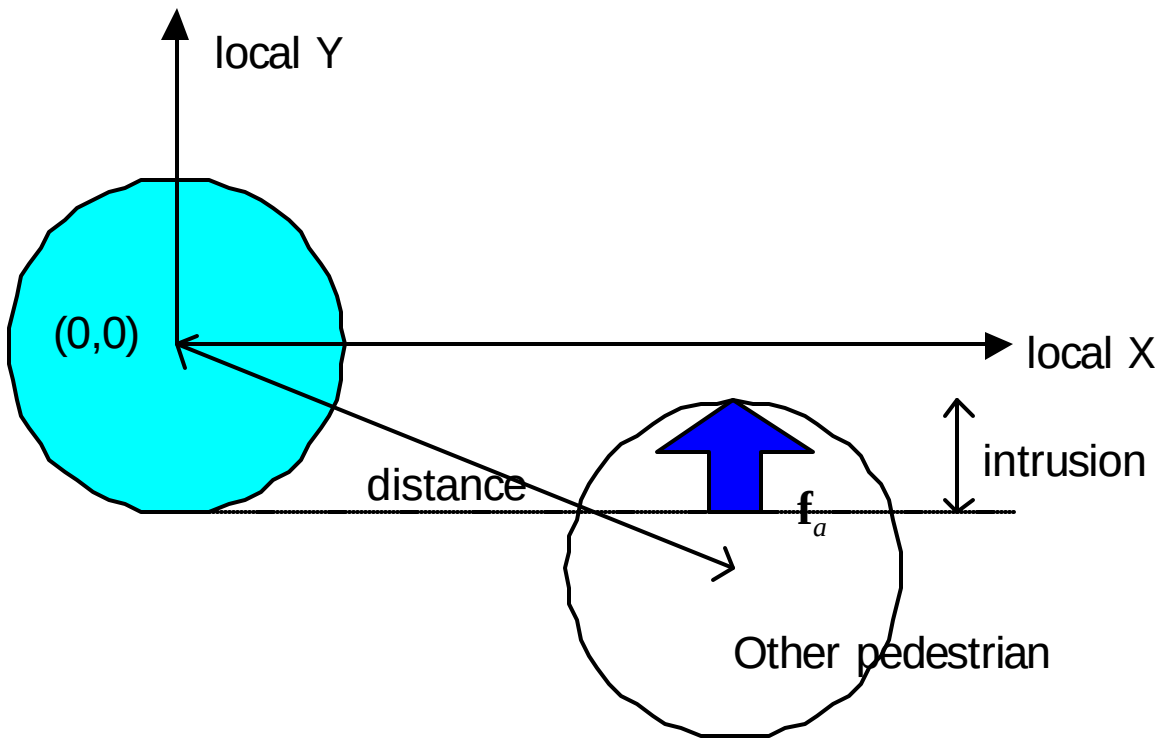

**Figure 4-2. Force to Repulse Away**

If $d(t)$, $y$ and $r$ are, respectively, representing the distance between the pedestrians, intrusion of the closest pedestrian in the area in front of the actor and the influence radius of pedestrian, the first repulsive intended velocity of pedestrian $i$, $\hat{\mathbf{v}}_a^i(t)$, in local coordinate is given by

$$\hat{\mathbf{v}}_a^i(t) = \frac{m_{\max}(2r - \mathbf{y}(t))}{c\, d(t)} = \frac{m_{\max}(2r - \mathbf{y}(t))}{c\, \|\mathbf{p}_k(t) - \mathbf{p}_i(t)\|} \quad \textbf{(4. 4)}$$

This intended velocity will drive the first repulsive force of the actor to turn away from the closest pedestrian (within the sight distance) with magnitude proportional to the intrusion of other pedestrian in the actor's way. The factor $m_{\max}/(c\, d(t))$ is the smoothing factor. The choice of using this smoothing factor is based on the assumption that the intended velocity must depend on the maximum walking speed of pedestrian and the intended velocity must increase non-linearly in proportion with the distance between the two pedestrians. A non-dimensional constant chi, $c$, is given to generalize the model and will be used as a constant calibration and validation of the model. Effect of the force to repulse away is depicted as diagrammatic in Figure 4-3 as two pedestrians avoiding each other. The smaller value of $c$ and greater the value of the sight distance will keep bigger avoidance distance between pedestrians. To determine of the sight distance will influence the value of the parameters $c$ and vice versa. If two pedestrians following each other and the pedestrian who follow has a greater maximum speed, the force to repulse away will direct the overtaking behavior. The value of parameter $c$ can be negative but not zero. In chapter 5, the effect of negative parameter $c$ will be discussed further.

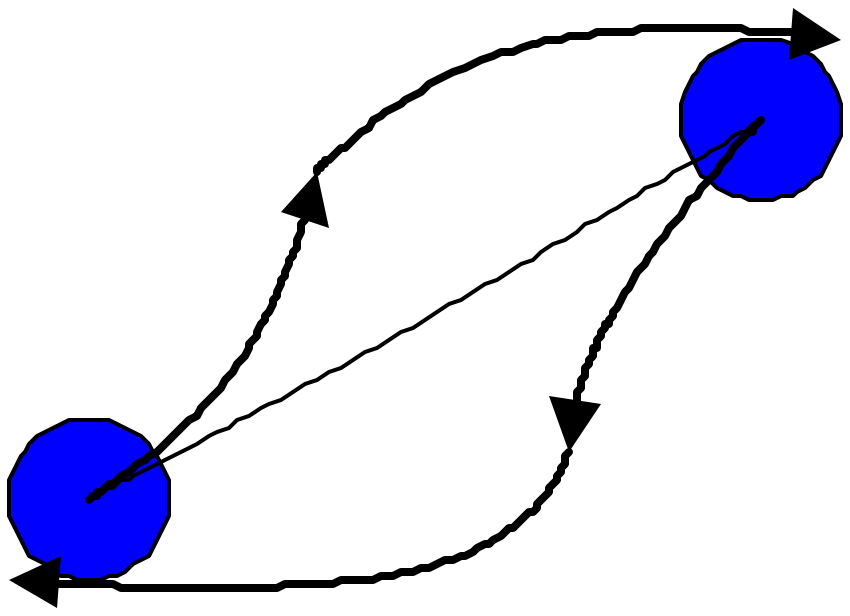

**Figure 4-3 Effect of the Force to Repulse Away**

The origin of the local coordinate system is in the center of the pedestrian actor, while the local $x$-axis is in the direction of the current velocity. The local coordinate system is a translation and a rotation from the current coordinate system into the centroid of the actor. Since the angle of rotation depends on the current velocity, then the local X direction is the current velocity direction. If $q$ is the rotation angle, then the transformation from Euclidean coordinate system to local coordinate system is given by

$$\mathbf{l} = \mathbf{R}.(\mathbf{g} - \mathbf{p}) \tag{4.5}$$

Where $\mathbf{l}$ and $\mathbf{g}$ are local and Euclidean coordinates, respectively. The rotation matrix $\mathbf{R}$ is given by

$$\mathbf{R} = \begin{bmatrix} \cos q & \sin q \\ -\sin q & \cos q \end{bmatrix} \tag{4.6}$$

The first repulsive force in equation (4. 4) should be transformed back to a current coordinate system using the inverse of the rotation matrix.

$$\mathbf{R}^{-1} = \begin{bmatrix} \cos q & -\sin q \\ \sin q & \cos q \end{bmatrix} \qquad \textbf{(4.7)}$$

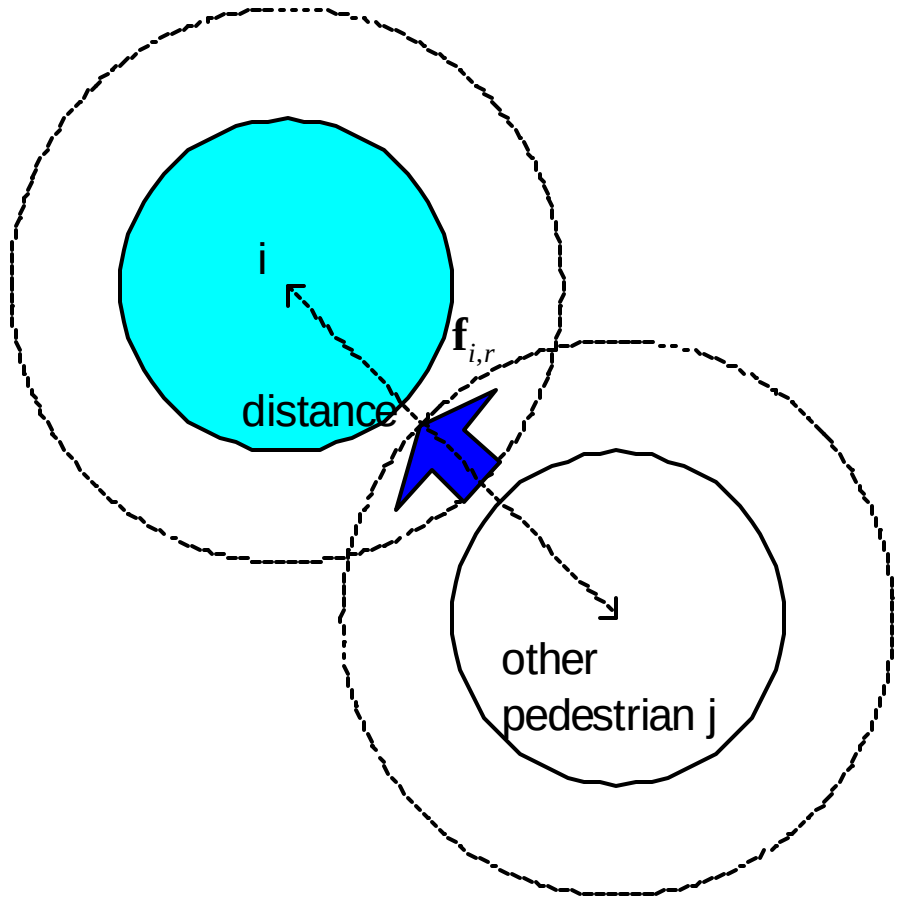

**Figure 4-4. Intended velocity to avoid collision**

By the first repulsive force, the pedestrian can move away from each other within a certain distance. However, there is no agreement that the pedestrians will not collide with each other when they are very close, especially when there are many pedestrians in the facility. Another repulsive force is needed to guarantee that no collision or minimum collision will actually happen between pedestrians. The second repulsive force, $\mathbf{f}_{i,r}$, is working to avoid the collision between pedestrians. To avoid the collision, it is assumed that each pedestrian has an influence radius that represents his or her security awareness. The force is generated when the influence radius of pedestrians overlaps each other as shown in Figure 4-4. No repulsive force is generated if the influence radius does not overlap each other. Instead of considering the closest pedestrian as the first repulsive force, the second repulsive force considers all surrounding pedestrians and the forces are summed up linearly.

Similar to the first repulsive force, the second repulsive force depends on the distance between the actor and other pedestrians surrounding it. The second repulsive force is given by its intended velocity, $\hat{\mathbf{v}}_{i,r}(t)$, as

$$\hat{\mathbf{v}}_{i,r}(t) = \frac{m_{\max}}{b}\sum_j \frac{(2r - d_{ij}(t))(\mathbf{p}_i(t) - \mathbf{p}_j(t))}{d_{ij}(t)}$$

The influence radius and distance between pedestrians are denoted by $r$ and $d_{ij}$ respectively. Rearranging this formulation and putting the distance in a vector notation will give

$$\hat{\mathbf{v}}_{i,r}(t) = \frac{m_{\max}}{b}\sum_j \left(\frac{2r}{\|\mathbf{p}_i(t) - \mathbf{p}_j(t)\|} - 1\right)\frac{(\mathbf{p}_i(t) - \mathbf{p}_j(t))}{\|\mathbf{p}_i(t) - \mathbf{p}_j(t)\|} \qquad \textbf{(4.8)}$$

The main magnitude of the intended velocity depends on the intrusion or overlap of the influence radius showed by the term $(2r - d_{ij}(t))$. The influence radius is assumed uniform for all pedestrians. The direction of the second repulsive force must be away from the other pedestrians, who have overlapped influence radius with the actor. This direction is explained using the vector term $(\mathbf{p}_i(t) - \mathbf{p}_j(t))$. As the smoothing actor, the intended velocity must depend on the maximum walking speed of the pedestrian. The intended velocity must also increase non-linearly in proportion with the distance between the two pedestrians, thus the maximum walking speed is given in the numerator and the distance is put in the denominator. The parameter beta, $b$, which has a spatial dimension (meter) is used to generalize the model and will be used as constant calibration and validation of the model. Since the direction of the intended velocity is already given by the vector terms, the value of the parameter beta must be positive. When $0 < b < 1$, the effect of the overlapping influence radius will be amplified by the parameter $b$. On the contrary, the overlapping influence radius will have a smaller effect if $b > 1$. The value parameter $b$ will influence the choice of influence radius and vice versa.

### 4.2.3 Microscopic Pedestrian Formulation

Each pedestrian in the system is influenced by the three forces. Four parameters of the model are the mass, $m$, alpha, $a$ , beta, $b$ and chi, $c$ . The mass is applied toward the three forces together (global parameter) while the other three parameters are applied only for the particular force.

The following is the proof that the mass is a global parameter. Suppose the mass is not a global parameter but also applied differently for each force. Then, there are three masses $m_f$, $m_a$ and $m_r$ for the forward-force, repulsive-force to move away and repulsive-force to avoid collision respectively. Let $\hat{\mathbf{v}}_f = m_{\max}\,\mathbf{g}(t)$ , $\hat{\mathbf{v}}_a = \dfrac{m_{\max}(2r - \mathbf{y}(t))}{d(t)}$ , and $\hat{\mathbf{v}}_r = m_{\max}\sum_j \left(\dfrac{2r}{\left\|\mathbf{p}_i(t) - \mathbf{p}_j(t)\right\|} - 1\right)\dfrac{(\mathbf{p}_i(t) - \mathbf{p}_j(t))}{\left\|\mathbf{p}_i(t) - \mathbf{p}_j(t)\right\|}$ represent the intended velocity from equation (4. 3), (4. 4), and (4. 8) respectively without the parameter alpha, $a$ , beta, $b$ and chi, $c$ . The time notation is deleted for convenient purposes. The three forces can be calculated using equation (4. 1) as $m_f\,\mathbf{a}_f = \dfrac{\hat{\mathbf{v}}_f}{a'} - \mathbf{v}$, $m_a\,\mathbf{a}_a = \dfrac{\hat{\mathbf{v}}_a}{c'} - \mathbf{v}$, and $m_r\,\mathbf{a}_r = \dfrac{\hat{\mathbf{v}}_r}{b'} - \mathbf{v}$. Since the mass for each force and the force parameter (i.e. $a'$, $b'$ and, $c'$) are dependent on each other, a common constant $m$ can be chosen such that $m = \dfrac{a'}{a} m_f = \dfrac{c'}{c} m_a = \dfrac{b'}{b} m_r$ . Thus $m_f = m a/a'$, and $m_f\,\mathbf{a}_f = \dfrac{\hat{\mathbf{v}}_f}{a'} - \mathbf{v}$ becomes $m\mathbf{a}_f = \dfrac{\hat{\mathbf{v}}_f}{a} - \dfrac{a'}{a}\mathbf{v}$ . Similar way is done for the other two forces. Summing up the three forces together produces

$$\left.\begin{aligned} m\mathbf{a}_f &= \frac{\hat{\mathbf{v}}_f}{a} - \frac{a'}{a}\mathbf{v} \\ m\mathbf{a}_a &= \frac{\hat{\mathbf{v}}_a}{c} - \frac{c'}{c}\mathbf{v} \\ m\mathbf{a}_r &= \frac{\hat{\mathbf{v}}_r}{b} - \frac{b'}{b}\mathbf{v} \end{aligned}\right\} m\left(\mathbf{a}_f + \mathbf{a}_a + \mathbf{a}_r\right) = \frac{\hat{\mathbf{v}}_f}{a} + \frac{\hat{\mathbf{v}}_a}{c} + \frac{\hat{\mathbf{v}}_r}{b} - c\mathbf{v}$$

However, $\mathbf{a} = \mathbf{a}_f + \mathbf{a}_a + \mathbf{a}_r$ by definition and $c = \dfrac{a'}{a} + \dfrac{b'}{b} + \dfrac{c'}{c}$ can be determined equal to one since $a'$, $b'$ and, $c'$ are variables. Thus, equation (4. 1) is consistent and the mass is a global parameter for the three forces.

Since $\mathbf{v}(t)=\frac{d\mathbf{p}(t)}{dt}$ and $\mathbf{a}(t)=\frac{d\mathbf{v}(t)}{dt}=\frac{d^2\mathbf{p}(t)}{dt^2}$, the formulation can be put together in terms of the current position of pedestrian $i$, $\mathbf{p}_i(t)$, as a second order differential equation

$$m\frac{d^2\mathbf{p}_i(t)}{dt^2}+\frac{d\mathbf{p}_i(t)}{dt}= m_{\max}\left\{\frac{\mathbf{e}(t)-\mathbf{p}_i(t)}{a\left\|\mathbf{e}(t)-\mathbf{p}_i(t)\right\|}+\frac{2r-\mathbf{y}(t)}{c\left\|\mathbf{p}_k(t)-\mathbf{p}_i(t)\right\|}+\sum_j\left(\frac{2r}{\left\|\mathbf{p}_j(t)-\mathbf{p}_i(t)\right\|}-1\right)\left(\frac{\mathbf{p}_j(t)-\mathbf{p}_i(t)}{b\left\|\mathbf{p}_j(t)-\mathbf{p}_i(t)\right\|}\right)\right\} \tag{4. 9}$$

Equation (4. 9) is a non-linear second order differential equation of pedestrian positions that depend on each pedestrian’s positions, speeds and accelerations. The analytical solution of the differential equation is very difficult and not practical since it is also dependent on the number of pedestrians and the sight distance. Numerical method through simulation is more favorable and it has the benefit to visualize the movement of each pedestrian in a plan as an animation.

## 4.3 BASIC PHYSICAL BASED SIMULATION

The differential equation (4. 9) is solved numerically by divide and conquer algorithm using Euler method, which provides adequate results while keeping the computational speed reasonable. Each equation is computed one by one, as each pedestrian is assumed an autonomous agent. A pedestrian has his own internal forces and influence other pedestrians only through his position. The pedestrian movement is based on the resultant forces that act upon him. Other numerical method to solve the differential equation such as Runge Kutta may produce a better approach to the differential equation but it decreases the computational speed significantly if the number of pedestrian is more than 100, thus, it is recommended for further study.

Since the computer computes things in discrete terms, the representation of the model is in the uniform time interval $\Delta t$ to approach the continuous model. Though it is a continuous approach, the real calculation using a computer is discrete, thus $\Delta t$ must be considered as a discrete time unit. As a common assumption the range value of the time interval is $0 < \Delta t \leq 1$. Smaller value of $\Delta t$ is preferable since the error of higher order terms is smaller. If the current location, velocity and acceleration are denoted by vector, $\mathbf{p}(t)$, $\mathbf{v}(t)$ and $\mathbf{a}(t)$, respectively, the basic dynamical model is given by Euler equations:

$$\mathbf{v}(t) = \mathbf{v}(t-\Delta t) + \mathbf{a}(t).\Delta t \qquad \textbf{(4.10)}$$

$$\mathbf{p}(t+\Delta t) = \mathbf{p}(t) + \mathbf{v}(t).\Delta t \qquad \textbf{(4.11)}$$

The three intended velocities are then summed up together with a weighing factor (mass) to determine the acceleration.

$$\mathbf{a}(t) = \frac{\left(\sum \hat{\mathbf{v}}(t)\right) - \mathbf{v}(t)}{\Delta t} = \frac{(\hat{\mathbf{v}}_f(t) + \hat{\mathbf{v}}_a^i(t) + \hat{\mathbf{v}}_{i,r}(t)) - \mathbf{v}(t)}{\Delta t} \qquad \textbf{(4.12)}$$

A note must be given that to maintain the distance of the current position to the destination from zero, the pedestrian must be removed from the simulation quite far from the destination. When a pedestrian reaches a certain radius of the point of destination, it is assumed that this pedestrian has arrived at the destination. In other words, the radius of destination is preserved to be quite reasonably large (i.e. depending on the radius of the pedestrian).

The transient state of the simulation happens in the first 5 second after the pedestrian is created. It corresponds to an unsteady condition of the simulations and it does not represent the correct behavior of pedestrians. Once the simulation starts, the overlapping pedestrians will repulse each other due to the repulsive force to avoid collision to behave correctly as the real pedestrians. After about 5 seconds, this unsteady condition gradually reaches the steadiness and ready to be measured. Therefore, the pedestrian generator is located outside the pedestrian trap. The location of the pedestrian generator is determined based on the time and space for the highest number of pedestrians generated to be steady.

## 4.4 COMPARISON WITH THE EXISTING MPSM

Table 4-1 compares the existing Microscopic Pedestrian Simulation Model (MPSM) with the proposed one. The italic words in the table show the weakness of the existing MPSM and the bold words represent the strong point of the proposed model. Several factors are chosen to show the similarity and the difference of each model. The factors are selected especially to reveal the main improvement of the proposed model compared to the exiting ones.

**Table 4-1 Comparison Microscopic Pedestrian Simulation Models**

| | **Benefit Cost Cellular** | **Cellular Automata** | **Magnetic Force** | **Social Force** | **Queuing Network** | **Proposed** |
|---|---|---|---|---|---|---|
| **Movement to goal** | Gain Score | Min (gap, max speed) | Positive and negative magnetic force | Intended velocity | Weighted random choice | Forward Force |
| **Repulsive Effect** | Cost Score | Gap or occupied cell | Positive and positive magnetic forces | Interaction forces | Priority rule (FIFO) | Two repulsive forces |
| **Value of the variables** | *Arbitrary score* | Binary | *Arbitrary score* | Physical meaning | Physical meaning | **Physical meaning** |
| **Higher programming orientation in** | *Scoring* | *If-then rules (heuristic)* | *Heuristic* | dynamical system (Continuous) | Queuing model | **dynamical system (Discrete)** |
| **Phenomena explained** | Queuing | Macroscopic | Queuing, way finding in maze | Queuing, Self organization | Queuing, evacuation | **Macroscopic, Queuing, Self organization, Evacuation** |
| **Evacuation or Dissipation time evaluation** | (Unknown) | (Unknown) | (Unknown) | *Not possible* | Possible | **Possible** |
| **Parameter Calibration** | *By inspection* | Compare Fundamental Diagram | *By inspection* | *By inspection* | *By inspection* | **Compare speed distribution** |

All MPSMs contain the terms that make the pedestrian move towards the destination. The detail of the term is somewhat different for each model, but this term always exists in the entire microscopic model. The benefit cost cellular model uses gain score, the cellular automata uses the minimum gap and maximum speed to drive the pedestrian toward the goal. In magnetic force model, the positive and negative magnetic load is set to the goal so that pedestrians will move to the destination. Similarly, the social force model is used the intended velocity as the main force driver. The proposed model uses a similar model to the social force model, with a slight modification. The social force model requires target time as an input to the model. Exactly the opposite of the fact, the dissipation time is the output of the proposed model, similar to the queuing network models.

Similar to the terms to move toward the goal, all MPSM also has a characteristic of the repulsive effect. The term drives the pedestrians to avoid each other. Again, the detail of the term is different by model. The benefit cost cellular model uses cost score, while the cellular automata avoid occupied cells. The magnetic force model sets the same magnetic load (positive with positive will repulse each other), and additional special force to avoid collision. The social force model uses distance between pedestrians to generate the interaction force. The proposed model is highly influenced by the social force model and the second force of the magnetic force model. It also uses distance between pedestrian to avoid each other from afar (the first repulsive force). For the second repulsive force, although it has a similar purpose as the second force of the magnetic force model, the detail is different. The magnetic force model uses angle of velocity and the relative speed, while the second repulsive force in the proposed model is generated if there is an intrusion toward the influence radius of pedestrian.

Instead of using arbitrary scores as in the magnetic force model or benefit cost cellular model, the proposed model uses the physical based variables that can be measured in the real world. Distance between pedestrians at each time, for example, can be detected using the Microscopic Video Data Collection, as explained in Chapter 3. The usage of the variables that can be measured is one of the main benefits of using physical based model. The magnetic force model, unfortunately, is still using arbitrary score for the magnetic load.

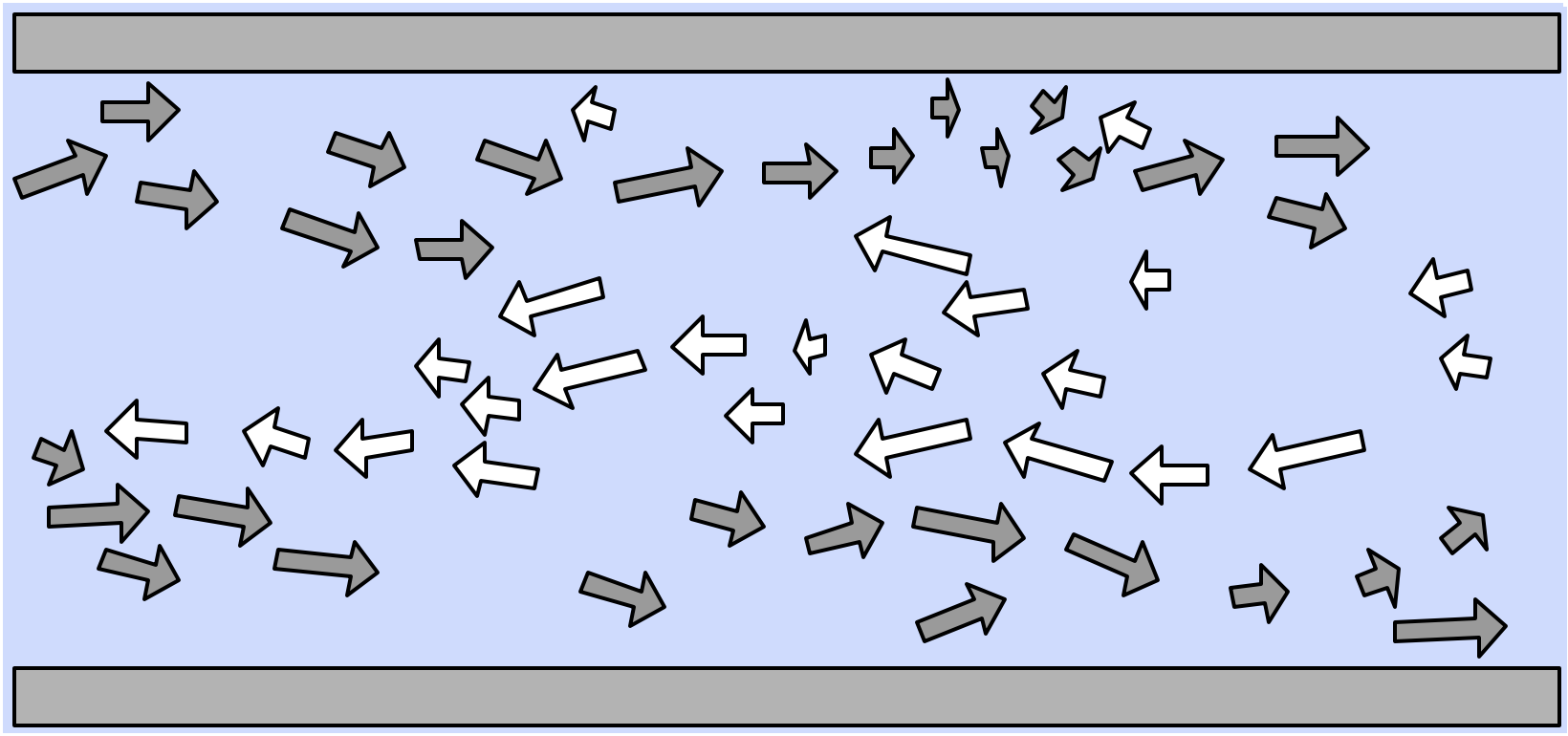

**Figure 4-5 Self Organization of Lane Formation**

Another weakness of the magnetic force model is in using the heuristic method (if then rule) as the main programming orientation. Similarly, weakness is suffered by the two cellular based models. The heuristic method has no theoretical bases and the rules, though it is working, are arguable. The proposed model uses the dynamical system, similar to the social force model but in discrete terms. The social force model assumed continuous dynamical systems.

Each model is used to explain a certain phenomena. The obvious phenomenon that exists in all the MPSMs is the queuing ability. Each autonomous pedestrian inherently is able to queue. The macroscopic behavior of pedestrian flow is only reported in the cellular automata model and the proposed model. The macroscopic behavior clarifies that the pedestrian average speed is reduced as the density of pedestrian flow increases. The magnetic force model has a unique ability to find a way from a maze due to its heuristic model [3] and evacuation [4]. The social force model reported the self-organization phenomena as shown in Figure 4-5. Though the simulation has no plan or rule to organize that behavior, the autonomous pedestrian systems create the lane formation automatically.

The proposed model is also able to form the same self-organization of lane formation behavior. This lane formation behavior can be seen from the real pedestrians, especially in the crossing when the pedestrian density is moderately high.

The performance that only existed in the queuing network model, which is the evacuation time, now also exists in the proposed model as the dissipation time. Queuing network model is developed for evacuation purposes. The proposed model has modified the intended velocity of the social force model to be able to set the dissipation time as the output.

The proposed model is calibrated and validated using real world data, as will be explained in Chapter 5. The other MPSMs are not calibrated using real world data due to their inherent model that does not allow the use of physical based variables or it is not the main interest of the researchers. Thus, the proposed model combines the main benefits of other existing MPSMs, to improve the deficiency of the existing models.

# CHAPTER 5 MICROSCOPIC PEDESTRIAN FLOW CHARACTERISTICS

The previous two chapters have given background on how the data was collected and analyzed using the simulation model. Both Microscopic Video Data Collection and Microscopic Pedestrian Simulation Model generate a database called $a$TXY database. Since the microscopic pedestrian flow characteristics can be derived from the $a$TXY database, it is useful to discuss the nature of $a$TXY database first. The formulations of the flow performance or microscopic pedestrian characteristics are followed in the next section. Sensitivity of the simulation and relationship between the flow performances are described. Calibration of the simulation using real world data is then explained through the comparison between the speed distributions.

## 5.1 NATURE OF THE $a$TXY DATABASE

$a$TXY database is a collection of microscopic pedestrians movement data where each pedestrian represented as a point to be seen from the top of the pedestrian facility and own a unique pedestrian number $a$. The movement of each pedestrian is sampled with uniform interval as coordinate X and Y on the surface of the pedestrian facility. The time of recording the movement coordinate corresponds to the observation time.

**Comment:** What is NTXY database

As explained in chapter 3, only the pedestrians who pass the pedestrian trap are considered for the $a$TXY database. The pedestrian trap is normally put in the middle of the walkway and the origin of the coordinate system is fixed into an arbitrary point. Figure 5-1 shows an example of a walkway to be seen from top view. The mosaic of the tiles along the walkway can be used to determine the coordinate system. The pedestrian trap is shown as bold lines in a rectangular box in the middle of the walkway and the origin of the coordinate system is in the bottom left outside the pedestrian trap.

The observation time starts at time 0 and recorded with a uniform time interval. The movement of each pedestrian who enters the scene is recorded over time as the $a$TXY database. The main database consists of four fields, which are pedestrian number, time and coordinate location. Pedestrian number is a unique number for each pedestrian and a new pedestrian number is given to a new person who enters the pedestrian trap. Only pedestrians in the pedestrian trap are recorded. The coordinate location of each pedestrian is the real world coordinate of the centroid area of the pedestrian. Each row in the database represents a single observation. One observation point is the position of a single pedestrian in one frame. The reference point of coordinate system is arbitrary. For the convention, in common one or two way traffic flow, the pedestrians are moving in the Y direction. Time recorded, $T$, represents the frame number or simulation clock. Because every two frames have an interval of $\Theta$ seconds, $T$ can be called as clock time.

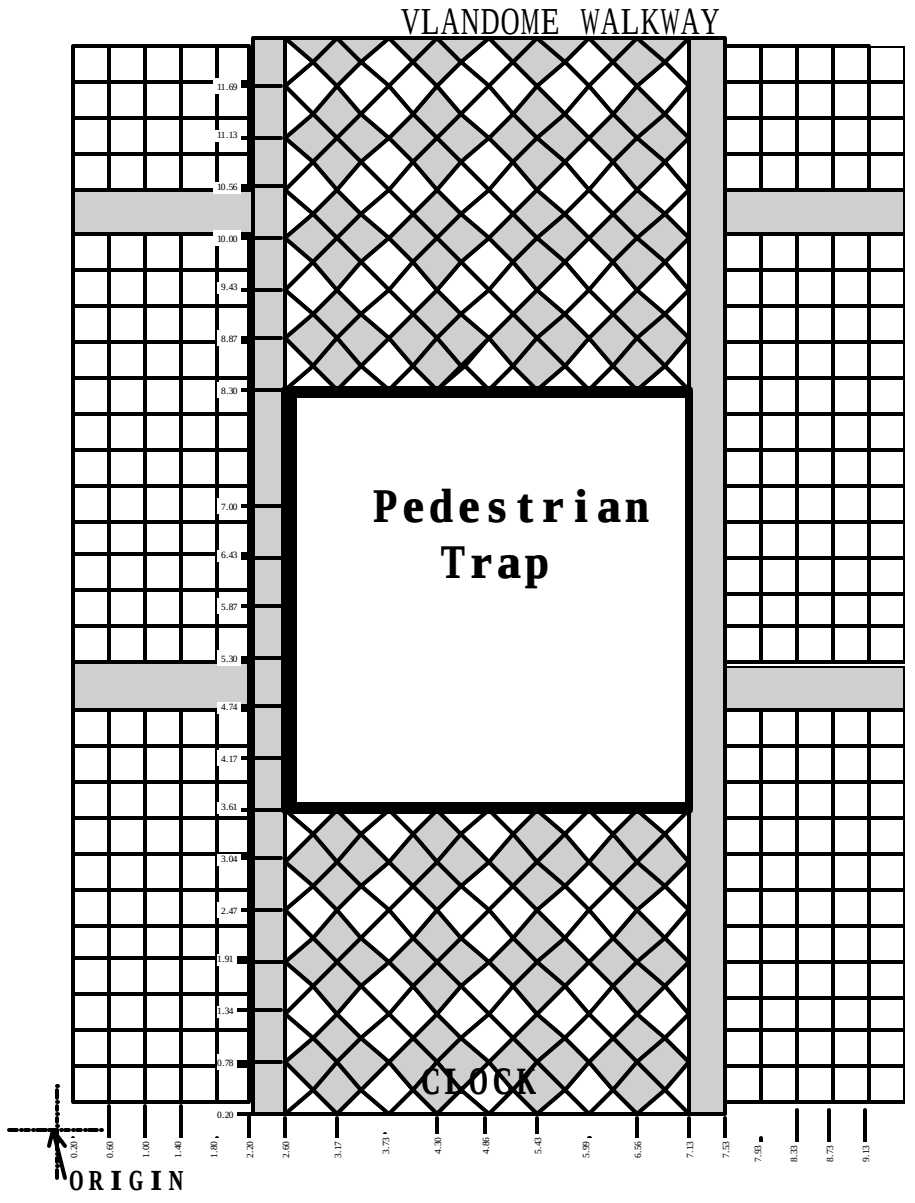

**Figure 5-1. Pedestrian Trap**

The $a$TXY database is a bridge between microscopic pedestrian simulation models, video data collection and pedestrian flow performance. Both simulator and the result of image processing may *create* the $a$TXY database. Similar $a$TXY database can be created by any Microscopic Pedestrian Simulation Model (MPSM). Both continuous and discrete based simulator can record the movement data of each pedestrian at every constant interval and write them into a text file.

Some application examples describe the usage of the formulation for design of pedestrian facilities and calibration of MPSM. The $a$TXY database may have three functions namely as a design criterion, a unifying language and a calibration criterion. As can be seen in Figure 5-2, pedestrian flow performance can be calculated based on the $a$TXY database. Whether the data comes from the real world or from simulation models, the flow performance can be calculated with the same definition of flow performance. If the design of pedestrian facilities is changed (in either real world or simulation), then a new $a$TXY database can be produced. By comparing the flow performance before and after the design change, the better design can be revealed. In this case, the flow performance functions as a design criterion for the facilities. It gives feedback for better design.

The $a$TXY database can also be used as a standard file of results from any MPSM and microscopic data collection of pedestrians by video. All MPSM can exchange data with another through the $a$TXY database. One simulator can use the pedestrian movement data produced by another type of simulator. Since there are many types of MPSM, there is a need for a unifying language so that all MPSM techniques can be used interchangeably without any confusion. It is a unifying language for all MPSM. Once the $a$TXY database is created, the movement of pedestrians can be viewed as if in the simulation using the viewer-program. Data collected from the real world also can be viewed as a simulation using the viewer-program.

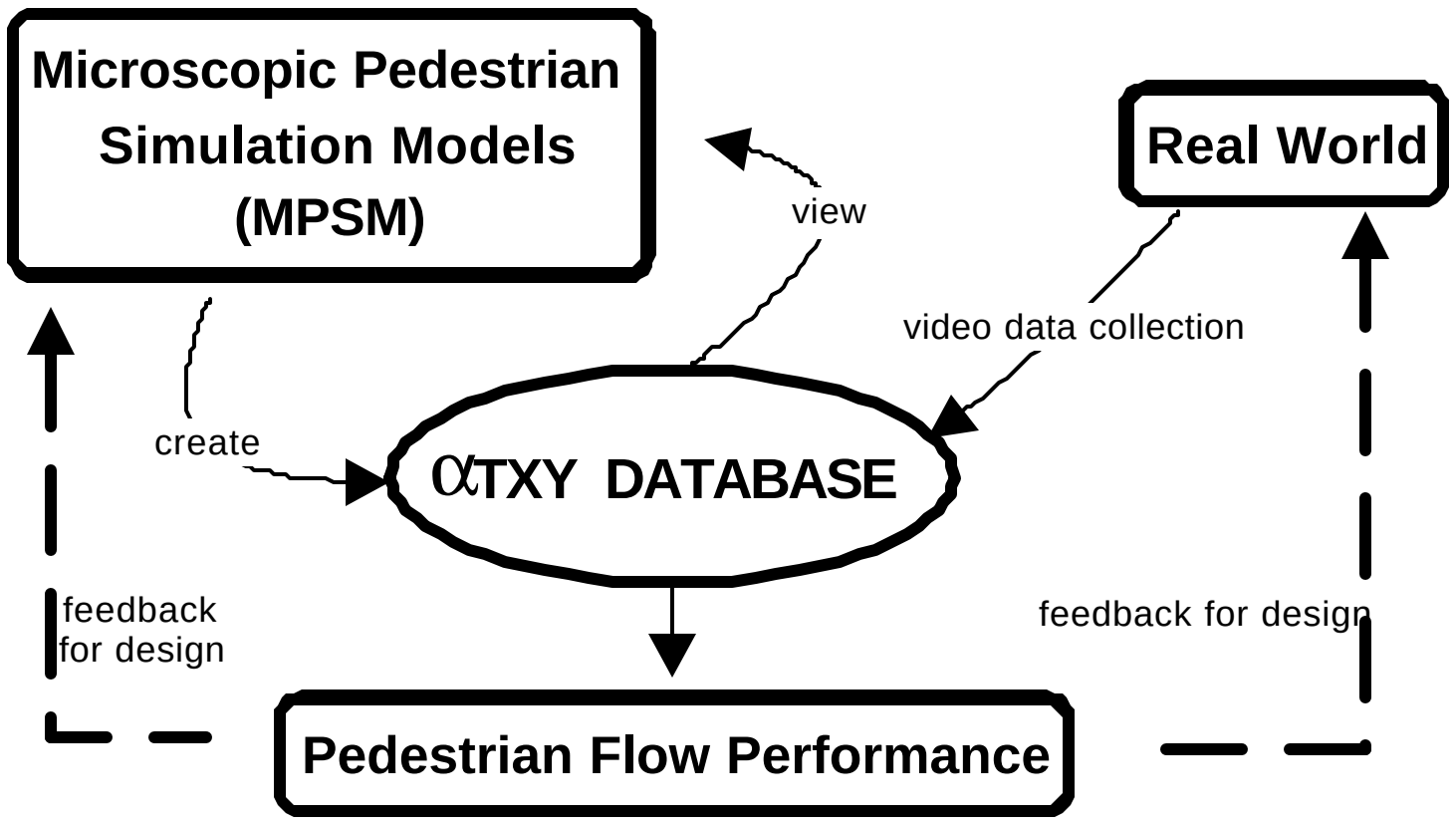

**Figure 5-2** Function of *α***TXY** Database

Different *α*TXY databases can produce almost the same flow performance. Seeking a relationship between the number of persons (as input) and the flow performance (as output) a parameter calibration of MPSM can be done as will be explained in the latter section of this chapter. Using the data from video collection, the average performance index of each pedestrian is determined. Similar calculation can be done to the data from the simulation. Least squares method is used to get the parameters of the simulation based on the assumption that the best parameters for the MPSM is the parameter that produces the most similar flow performance to the real world data.

### 5.2 FLOW PERFORMANCE DETERMINATION

Microscopic pedestrian characteristics are pedestrian flow performances, which is defined as the indicators to measure the interaction between pedestrians. Thus, the word 'flow performances' and pedestrian characteristics will be used interchangeably. Once such microscopic data is collected, another problem on how to measure the flow performance from the microscopic data collection arises. The results of microscopic data collection are the locations of each pedestrian at each time slice. These huge data need to be reduced into information that can be readily understood and interpreted. Pedestrian flow performances are numbers that measure the efficiency of pedestrian flow. They measure direct or indirect ways of the interaction between pedestrians and the interaction between pedestrians with their environment. Direct ways mean measuring the interaction itself (i.e. distance between pedestrians). Indirect ways mean measuring the result of the interaction, instead of the interaction itself. Pedestrian delay for example is caused by the interaction. Better flow performance should aim for better pedestrian interaction and better pedestrian interaction is identical with a better quality movement for pedestrians.

Individual and average delay and uncomfortability indices, speed change, jerk, change of moving direction, are typical examples of pedestrian flow performance. They measure the result of the interaction among pedestrians and interaction between pedestrians with the facilities. They measure pedestrian interaction indirectly. Pedestrian flow performance is a broader concept than merely traffic flow variables. Since traditional traffic flow variables, such as pedestrian headway, flow, speeds and area modules also measure the interaction between pedestrians indirectly, they are also part of the pedestrian flow performance.

The pedestrian flow performance can be measured through distances, and angles of moving direction. It may be valued over time as a change of distances and angles (e.g. speed) or as a rate of change (e.g. acceleration), or as a rate of change of the change (e.g. jerk). They may be quantified for an individual at each time or toward other pedestrians nearby or against the boundary of the surroundings. Variation of ratios, indices and comparison toward some references may also produce different kinds of pedestrian flow performance. Since many kinds of pedestrian flow-performances may be developed, it is recommended for further research work. For this research work, many high probable flow performances have been derived and tried through the simulation or data collection, (i.e., pace index, variation of walking displacement, to name a few) but most of them have no significant impact on the pedestrian flow. They become meaningless since their values are changing randomly and not related to pedestrian density, speed or flow. Only some prominent flow performances are discussed in this section, which are average speed, delay, and uncomfortability index and dissipation time.

Since the unifying method from simulation and the data collection required for the pedestrian characteristics are derived from the $a$TXY database, the rest of this section explains the formulation of the prominent pedestrian flow performance from the $a$TXY database. To ease the formulation, the notation uses discrete terms. The $\Delta t$ is a fixed small number representing the observation time interval, which is the time gap between two consecutive frames of image, and the reciprocal of the number of frames per second. Since the interval between times is constant, it is easier to write this step of the time as an increase of a natural number.

The following notations are used in this chapter:

$t$ = simulation time clock ,

$N(t)$ = total number of pedestrians in the trap at time $t$,

$\mathbf{x}(a,t)$ = vector location (X,Y) of the pedestrian $a$ at time $t$,

$\mathbf{x}(a,t-1)$ = vector location (X,Y) of the pedestrian $a$ at one time step before $t$, which is actually at time $t-\Delta t$,

$v(a,t)$ = instantaneous speed of the pedestrian $a$ at time $t$,

$\bar{v}(a,t)$ = time average of instantaneous speed of the pedestrian $a$ at time $t$,

$\overline{v^2}(a,t)$ = time average of the square of the instantaneous speed of the pedestrian $a$ at time $t$,

$v_{\max}(a)$ = predetermine maximum speed of pedestrian $a$ ,

$\tilde{v}(t)$ = average instantaneous speed at time $t$,

$\bar{v}(t)$ = time average of the average instantaneous speed, it is the system flow performance on speed,

$w(a,t)$ = cumulative walking distance of pedestrian $a$ from the time he/she enters the trap to time $t$ (still inside the pedestrian trap),

$d(a,t)$ = instantaneous delay of the pedestrian $a$ at time $t$,

$\tilde{d}(t)$ = average instantaneous delay of the pedestrian $a$ at time $t$,

$\bar{d}(t)$ = time average of the average instantaneous delay, it is the system flow performance on delay,

$u(a,t)$ = instantaneous uncomfortability index of the pedestrian $a$ at time $t$,

$\tilde{u}(t)$ = average instantaneous uncomfortability index at time $t$,

$\bar{u}(t)$ = time average of the average instantaneous uncomfortability index, it is the system flow performance on uncomfortability index.

In case of microscopic pedestrian movement data, there are three types of flow performance:

1. Instantaneous individual flow performance;
2. Average instantaneous flow performance;
3. System flow performance.

The first type of flow performance is calculated for each pedestrian $a$ at instant time $t$, denoted by $f(a,t)$. It is done within the movement of one pedestrian. The second type of flow performance is the average of pedestrians flow performance over all pedestrians in the system at current time t and denoted by $\tilde{f}(t)$. It is a system performance at certain time t. The average is performed by the summation of the instantaneous flow performance at that time divided by the total number of pedestrians in the system at the same time. The third type of flow performance is the overall pedestrians flow perfo rmance in the system over time. Except for dissipation time, the other three prominent flow performances are the time average of the instant system flow performance. It can be done by averaging the second average over time. The time average formulation is based on recursive formula that has been explained in chapter 2.

### 5.2.1 Speed

In the simulation, the speed is derived directly from the force and acceleration as explained in Chapter 4, while from microscopic video data collection, the speed is simply one step distance divided by constant $\Delta t$. Using discrete term $\Delta t = 1$, the instantaneous speed for individual pedestrians is calculated as

$$v(a,t) = \left\| \mathbf{x}(a,t) - \mathbf{x}(a,t-1) \right\| \tag{5.1}$$

Later in the calibration section in this chapter, the consideration of the real value of $\Delta t$ will be discussed.

The system flow performance at instant time $t$, is given by

$$\tilde{v}(t) = \frac{\sum_{a=1}^{N(t)} v(a,t)}{N(t)} \tag{5.2}$$

The overall or system average speed is formulated as

$$\bar{v}(t+1) = \frac{t-1}{t}\bar{v}(t) + \frac{\tilde{v}(t)}{t} \tag{5.3}$$

It is the time average of the instantaneous average speed from the first pedestrian who enters the trap until the last pedestrian who goes out of the trap assuming the pedestrian trap always contain at least one pedestrian. After a time gap or when the pedestrian trap is empty, a new system average speed will be determined.

### 5.2.2 Uncomfortability

The uncomfortability index follows the suggestion of Helbing and Molnar (1997) as explained in Chapter 2. In this section, the formulations are derived from the $a$TXY database. To determine the instantaneous uncomfortability index, two values must be calculated before hand. First, the time average of instantaneous speed of the pedestrian $a$ at time $t$ is determined as

$$\bar{v}(a,t+1)=\frac{t-1}{t}\bar{v}(a,t)+\frac{v(a,t)}{t} \tag{5.4}$$

Second, the time average of the square of the instantaneous speed of the pedestrian $a$ at time $t$ calculated as

$$\overline{v^2}(a,t+1)=\frac{t-1}{t}\overline{v^2}(a,t)+\frac{v^2(a,t)}{t} \tag{5.5}$$

Using those two values, the instantaneous uncomfortability index for pedestrian $a$ is established as

$$u(a,t)=1-\left(\frac{\bar{v}^2(a,t)}{\overline{v^2}(a,t)}\right) \tag{5.6}$$

The instantaneous system average of uncomfortability index is given by

$$\tilde{u}(t)=\frac{\sum_{a=1}^{N(t)}u(a,t)}{N(t)} \tag{5.7}$$

The overall system flow performance on the uncomfortability index is the time average of the instantaneous system average, or obtained by the recursive formula

$$\bar{u}(t+1)=\frac{t-1}{t}\bar{u}(t)+\frac{\tilde{u}(t)}{t} \tag{5.8}$$

### 5.2.3 Delay

The instantaneous delay is the time difference between walking with the maximum speed and walking with the average speed. To calculate the instantaneous individual delay, walking distance must first be established. The walking distance is the integral of the path of pedestrian movement, which can be derived from the $a$TXY database as

$$w(a,t) = \sum_{k=1}^{t} \left\| \mathbf{x}(a,k) - \mathbf{x}(a,k-1) \right\| \tag{5.9}$$

The individual instantaneous delay is then derived as

$$d(a,t) = \frac{w(a,t)}{\bar{v}(a,t)} - \frac{w(a,t)}{v_{\max}(a)} \tag{5.10}$$

Where the $\bar{v}(a,t)$ is calculated using equation (5. 4).

The average performance of the system is calculated at each time. Similar to the speed and uncomfortability index, the instantaneous system average of delay is given by the sum of all instantaneous delays of all pedestrians divided by the number of pedestrians in the trap.

$$\tilde{d}(t) = \frac{\sum_{a=1}^{N(t)} d(a,t)}{N(t)} \tag{5.11}$$

The overall system delay is the time-average of the instantaneous system average, or obtained by

$$\bar{d}(t+1) = \frac{t-1}{t}\bar{d}(t) + \frac{\tilde{d}(t)}{t} \tag{5.12}$$

### 5.2.4 Dissipation Time

Unlike the previous pedestrian flow performances, the dissipation time has no instantaneous flow performance. The dissipation time is calculated from the first pedestrian who enters the trap until the last pedestrian who goes out of the trap. If the time of the first pedestrian who enters the trap is recorded as well as the last pedestrian who goes out of the trap, the dissipation time is simply the time difference between the last and the first recorded time.

This pedestrian characteristic is special and very useful for pedestrian crossing. If the measurement is about pedestrian walkway where the pedestrian may come randomly, this flow performance has no meaning. Higher dissipation time mean the bulk of pedestrians need longer time to cross a road and it is considered as inefficient. Later in Chapter 6, some policy will be proposed to reduce the dissipation time of pedestrian crossing.

## 5.3 SENSITIVITY ANALYSIS

In this section, the microscopic simulation model will be further discussed. There are two control variables in the simulation, which are the maximum speed and the total number of pedestrian (or density because the area of the trap is fixed). The simulation model has four parameters, which are the mass ($m$), alpha ($a$), beta ($b$) and chi ($c$). By changing the values of the control variables and the parameters, the sensitivity of the parameters and many relationship between variables can be revealed in this and the next section. Using the speed density relationship (called *u-k graph*), the fundamental diagram of traffic flow can be determined. The density is sometimes represented as the total number of pedestrians rather than the density itself simply because it is easier to read an integer number than a decimal number.

The sensitivity of the average speed and the u-k graphs will be discussed first, then the sensitivity of the other major pedestrian characteristics such as uncomfortability, delay and dissipation time follows. Sensitivity of $\Delta t$ is discussed after that.

### 5.3.1 Average Speed and u-k Graphs

As explained in Chapter 2, the common relationship between speed and density or the u-k graph is linear. Interestingly, the microscopic simulation model that had been described in Chapter 4 also produces linear u-k graphs if the speed represents the system average speed. The density as characterized by the total number of pedestrian is the control variable of the experiments since the area of the trap is constant. Figure 5-3 shows the u-k graphs as it is influenced by the maximum speed. The speed-density relationship is linear with higher maximum speed on the top of the lower one. It is interesting to note that the gradient and the intercept of the graph are changing as the maximum speed is changing. The scattered data is added to show the variation of the data toward the model.

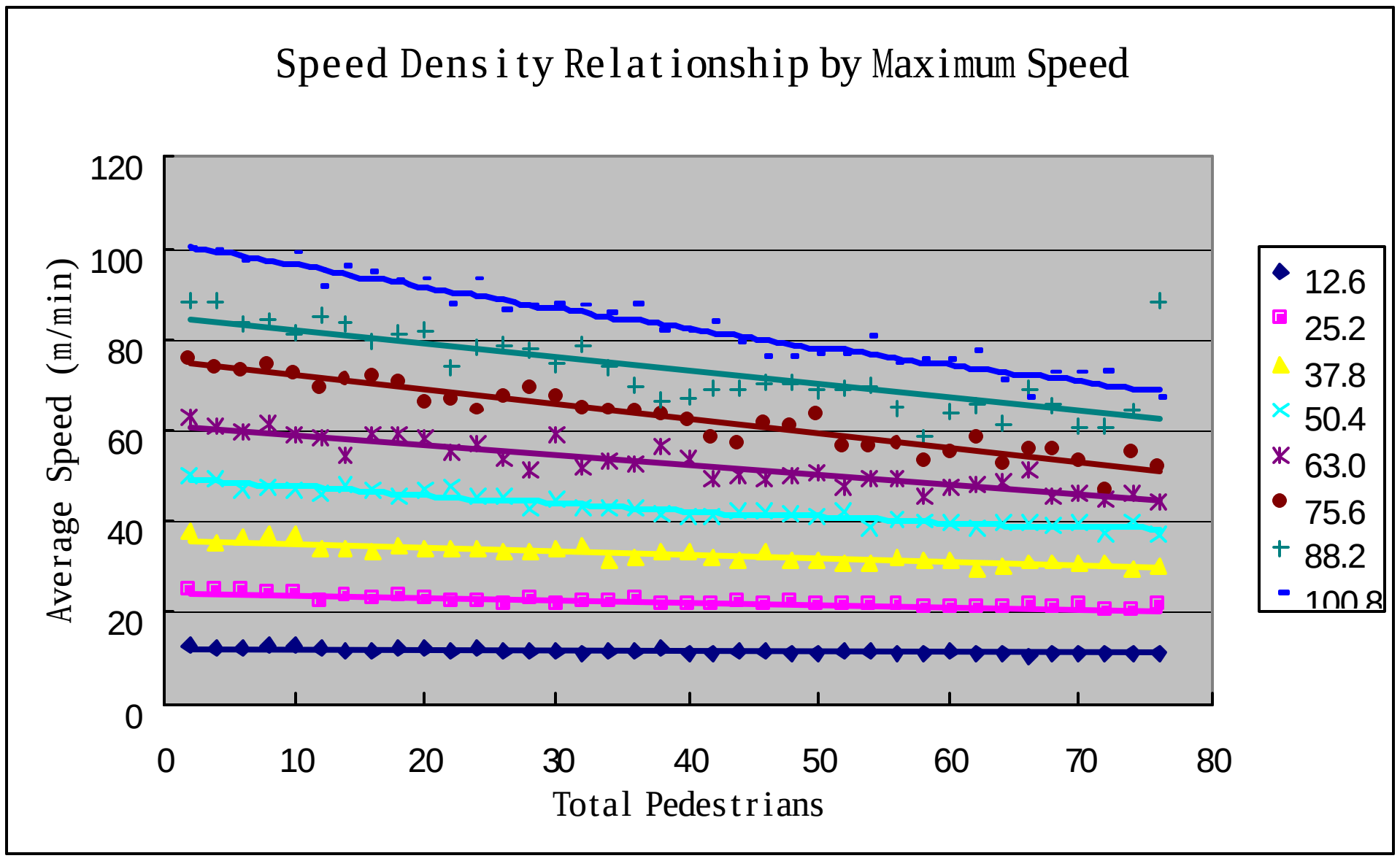

**Figure 5-3 Effect of the Maximum Speed toward Speed-Density Graphs**

Eight categories of maximum speed were utilized as the control variable; the intercept of the u-k graphs with the vertical axis represents the free flow speed or the maximum speed. Category 1 is about 12.6 m/min and category 8 is about 100.8 m/min. The r-squared value between the maximum speeds and the intercepts is 0.999 characterizing the very strong relationship between the control variable and the data.

The influence of the maximum speed can be derived further by two statistics, the mean and variance of the average speed. They show how the average speed changes as the maximum speed increases or decreases. Figure 5-4 shows the relationship between the maximum speed and the mean (left) and variance (right) of the average speed. It reveals that the average speed increases linearly as the maximum speed increases while the variance of the average speed increases exponentially. Both models have an r-squared value of 0.99 and the mean of the model is $Y = 0.789X + 2.53$ while the variance model is $Y = 4.587e - 5.X^{3.208}$.

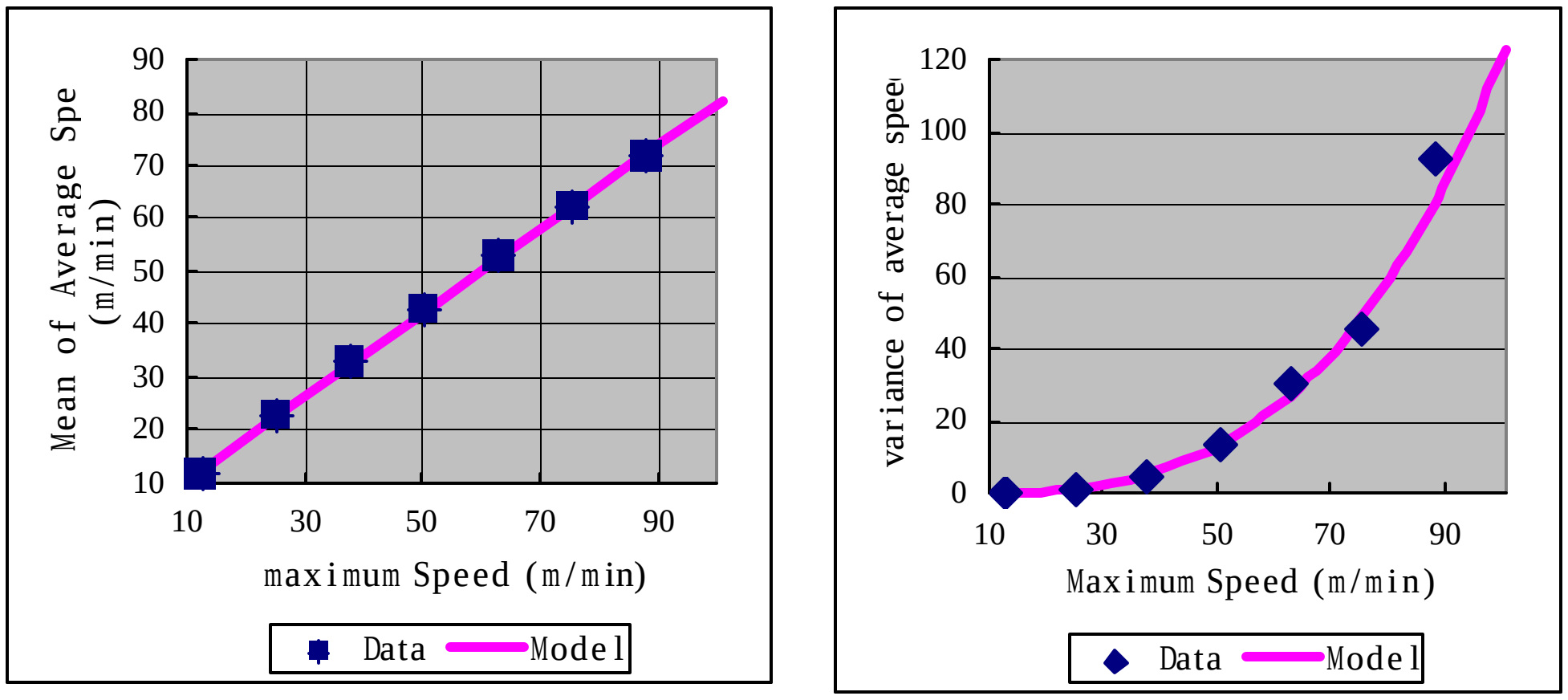

**Figure 5-4 Effect of the Maximum Speed for the Mean and Variance of the Average Speed**

The slope of the u-k graphs or the ratio between speed and density decreases linearly as the maximum speed increases. The model is $Y = 0.09485 - 0.00555X$ with an r-square value of 0.972. Figure 5-5 shows this association.

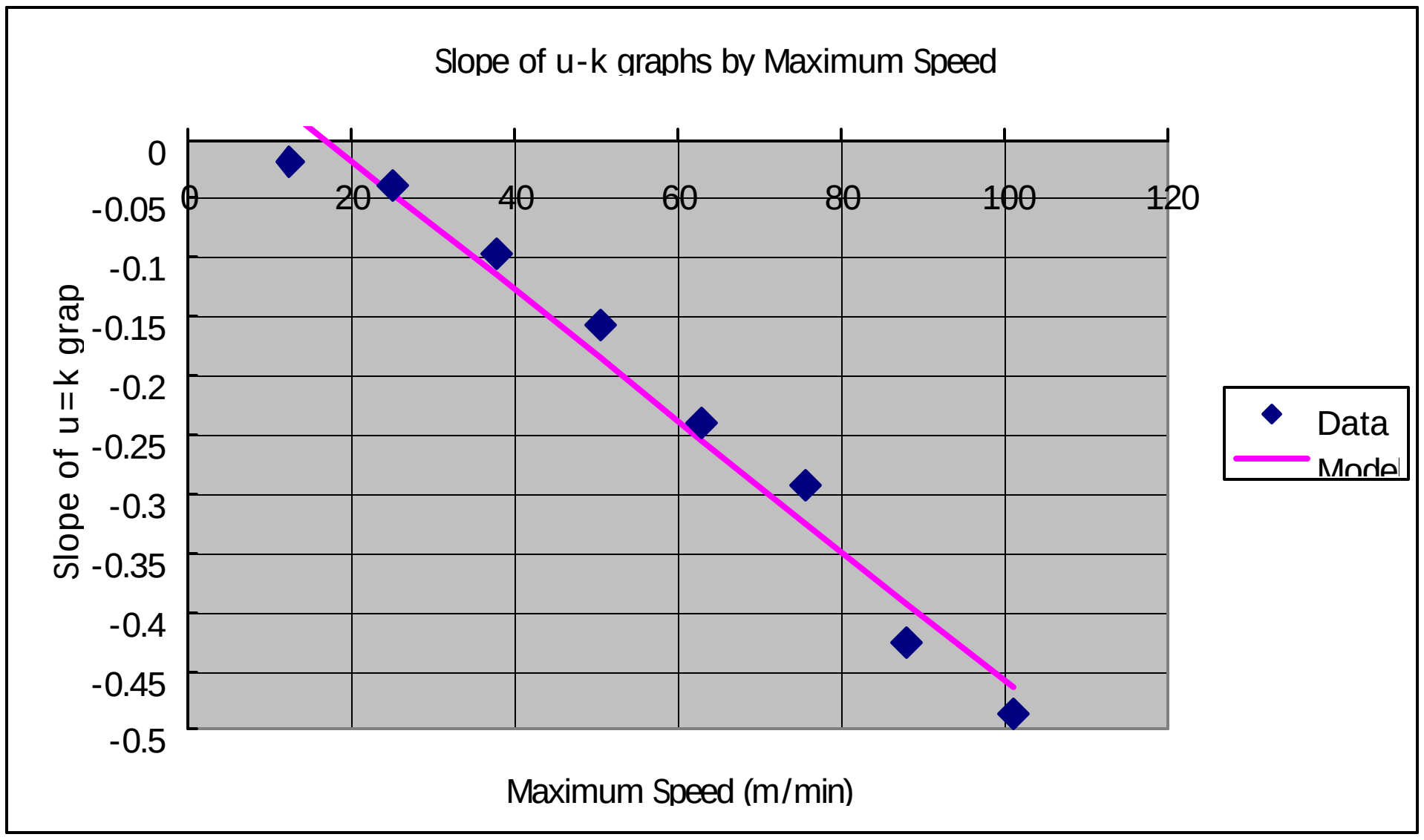

**Figure 5-5 The Influence of the Maximum Speed toward the Slope of u-k Graphs**

Intensive numerical experiments had also been done to get the sensitivity values of the parameters using one-way pedestrian traffic and maximum speed of 100.8 m/min. After a great deal of failure of the trial and errors using big numbers of parameters, it was revealed that the best sensitivity only could be achieved when the value of parameters is between zero and one. Higher than one, the parameters are not so sensitive for anything. The results of the trial and errors of the numerical experiments are summarized in Table 5-1. The way to do the trial and error was as follows. First, only single parameter was increase and the relationship between the parameter with the mean of average speed; slope of u-k graphs and free flow speed were investigated. Since there was no relationship between the parameters with those variables, further trial was done with two and three equal parameters and it was revealed that the three equal parameters of alpha, beta and chi produce better result.

The three variables mean of average speed, slope of u-k graphs and the free flow speed were set to be the standard variables of the sensitivity analysis. If the changing of the parameters has influence over the standard variables, the sensitivity is said to be successful. The choice of these three variables is based on consideration that these three variables will have great influence over the validation of the simulation, especially the slope of the u-k graphs.

**Table 5-1. Summary Result of the Sensitivity of Parameters**

| Parameters Increase | Mean of Average Speed | Slope of u-k Graphs | Free flow Speed |
|---|---|---|---|
| Alpha | No influence | No influence | No influence |
| Beta | No influence | No influence | No influence |
| Chi | Increase | No influence | Increase |
| Alpha = Beta | No influence | No influence | No influence |
| Beta = Chi | Increase | No influence | No influence |
| Alpha = Chi | Increase | Decrease | Increase |
| Alpha = Beta = Chi | Increase | Decrease | Increase |
| Mass | Decrease | Decrease | Increase |

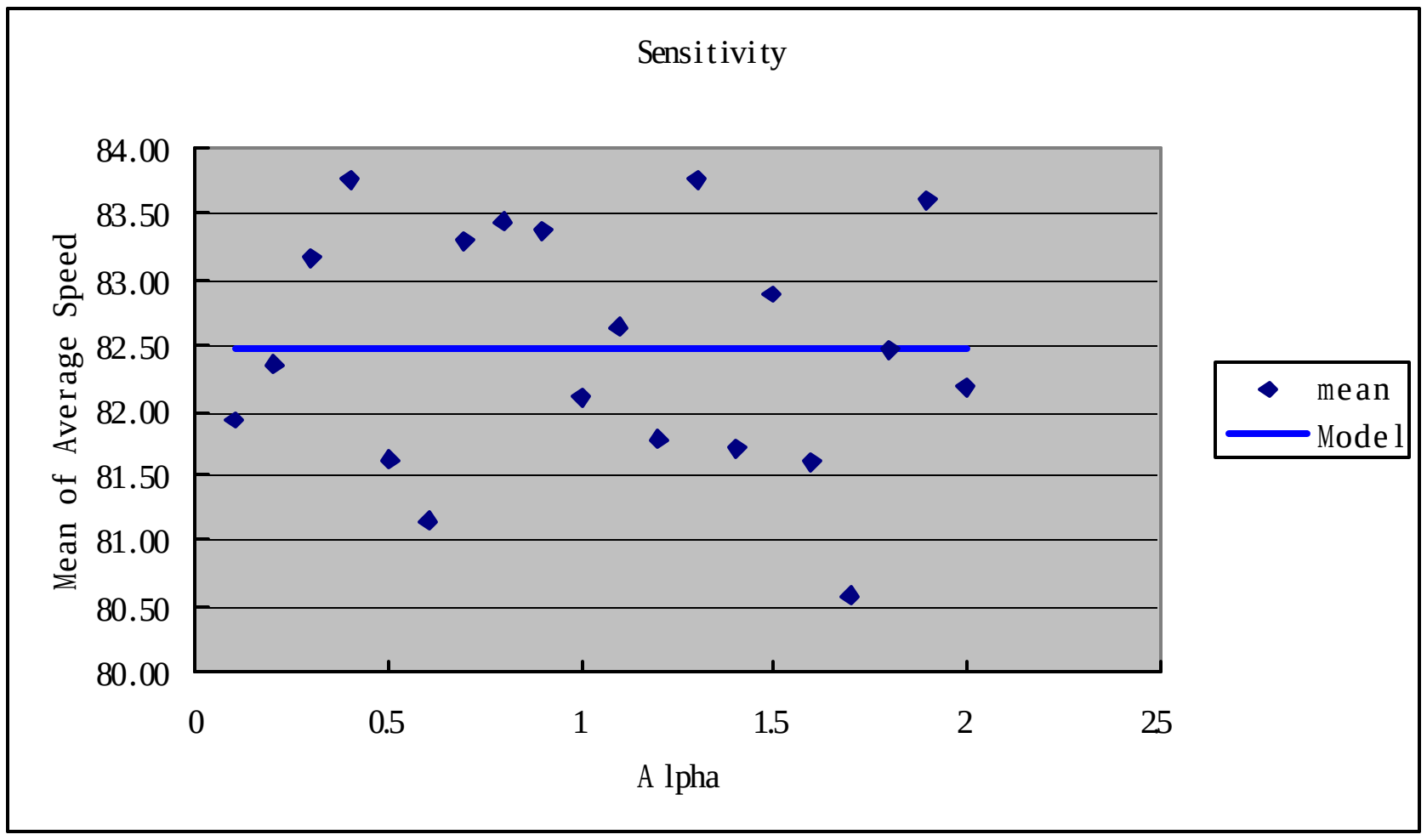

**Figure 5-6 Alpha has No Influence over the Mean of Average Speed**

The results of the numerical experiments show that the change of individual parameters alpha, beta and chi has no influence over the standard variables. As an illustration, Figure 5-6 shows that alpha has no influence over the mean of the average speed. Combination of two equal parameters has better influence over the standard variables but the best combination is that alpha should be set equal to beta and chi. Smaller sum of square error (SSE) is seen when the value of parameter alpha = beta = chi.

Figure 5-7 shows the sensitivity of the four parameters of the simulation model. In the left side, the value of parameter alpha is equal to beta and chi while in the right side; the sensitivity of the mass parameter is shown. The model and the sum of square error are shown in the bottom of each graph. It is interesting to note that the increase of alpha or beta or chi together will increase the mean of the average speed while the increase of the mass will even reduce the mean of the average speed. When the values of the parameters are higher than one, the graphs tend to reach the asymptotic value or constant.

### 5.3.2 Uncomfortability

The increase in the parameters has a tendency to reduce the uncomfortability index as shown in Figure 5-8. Similar to the average speed, the graph tends to reach the asymptote when the parameters are higher than one.

### 5.3.3 Delay

The increase of the parameter alpha, beta and chi together is likely to reduce the mean delay, while the increase of the mass is liable to increase the mean of the delay. The mean delay is asymptotically constant when the parameters are higher than one. Figure 5-9 illustrates the influence of the parameters toward the average delay.

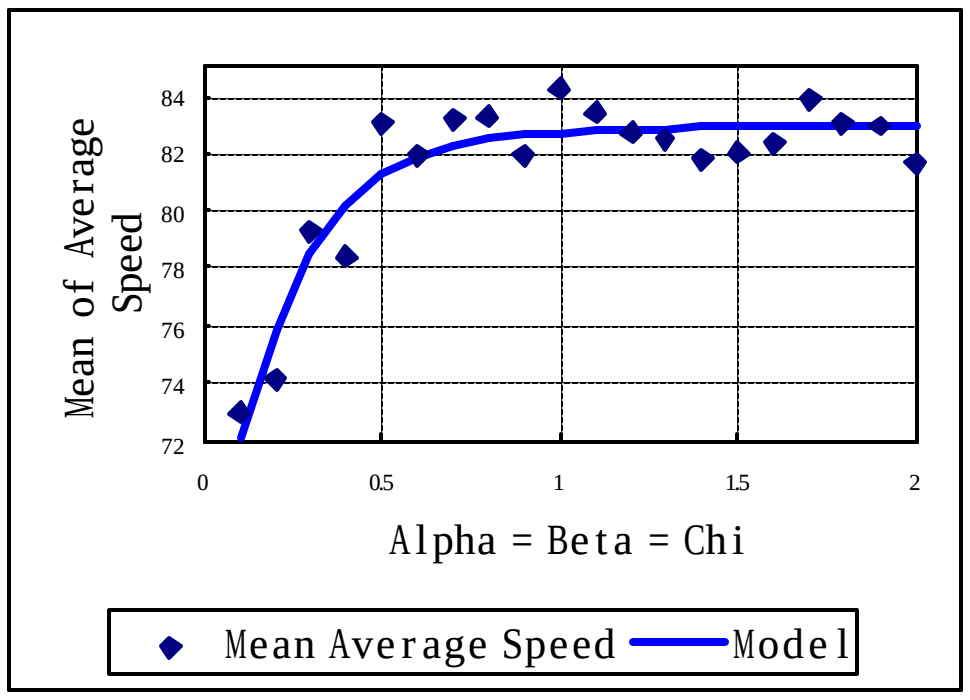

$$Y = \frac{82.91}{1+0.246.e^{-4.926X}},\ \text{SSE} = 20.22$$

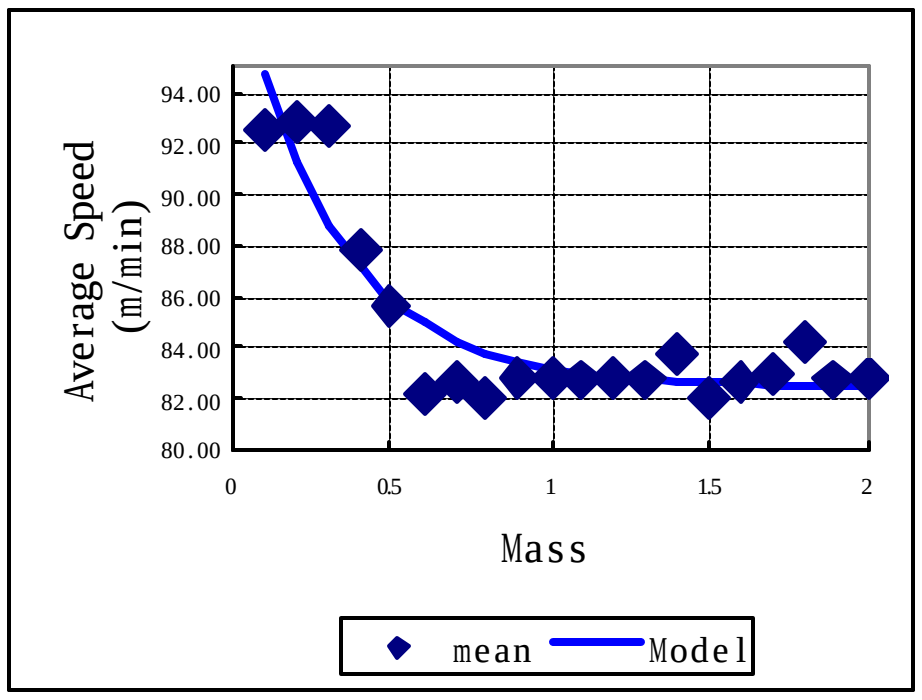

$$Y = \frac{82.39}{1-0.175.e^{-2.9435X}},\ \text{SSE} = 41.82$$

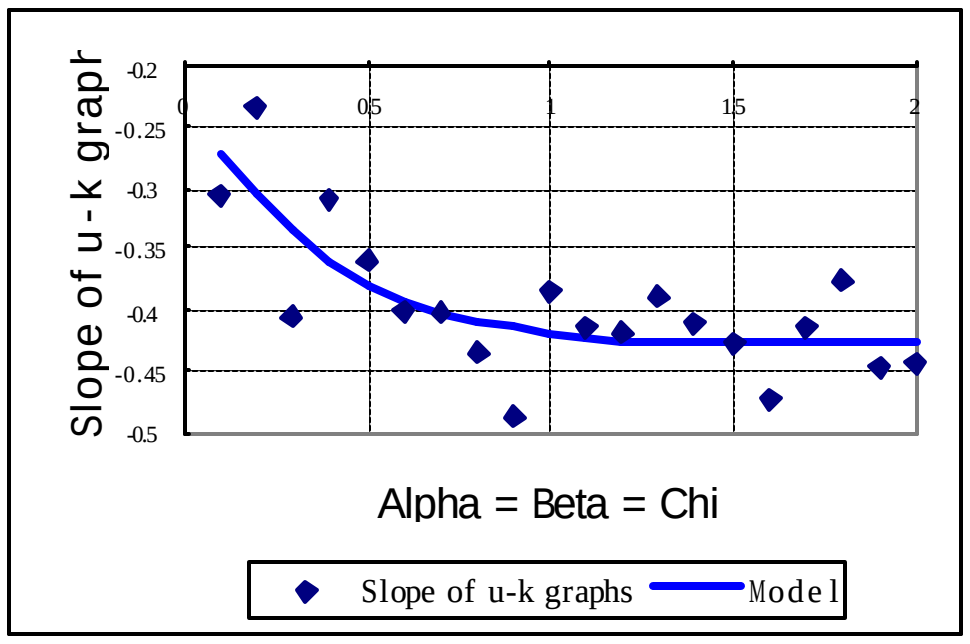

$$Y = \frac{-0.427}{1+0.8355.e^{-3.669X}},\ \text{SSE} = 0.0287$$

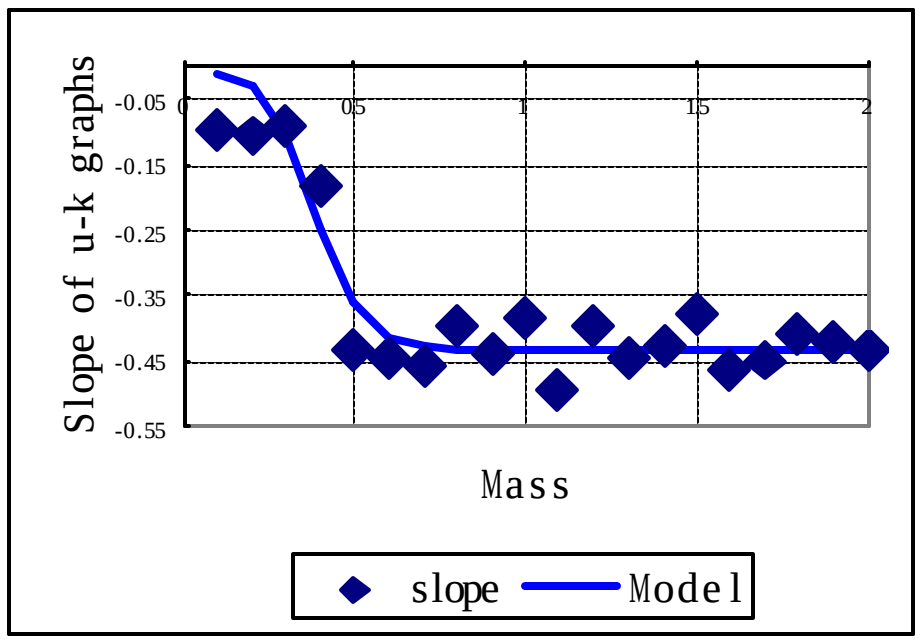

$$Y = \frac{-0.433}{1+231.4.e^{-14.268X}},\ \text{SSE} = 0.0377$$

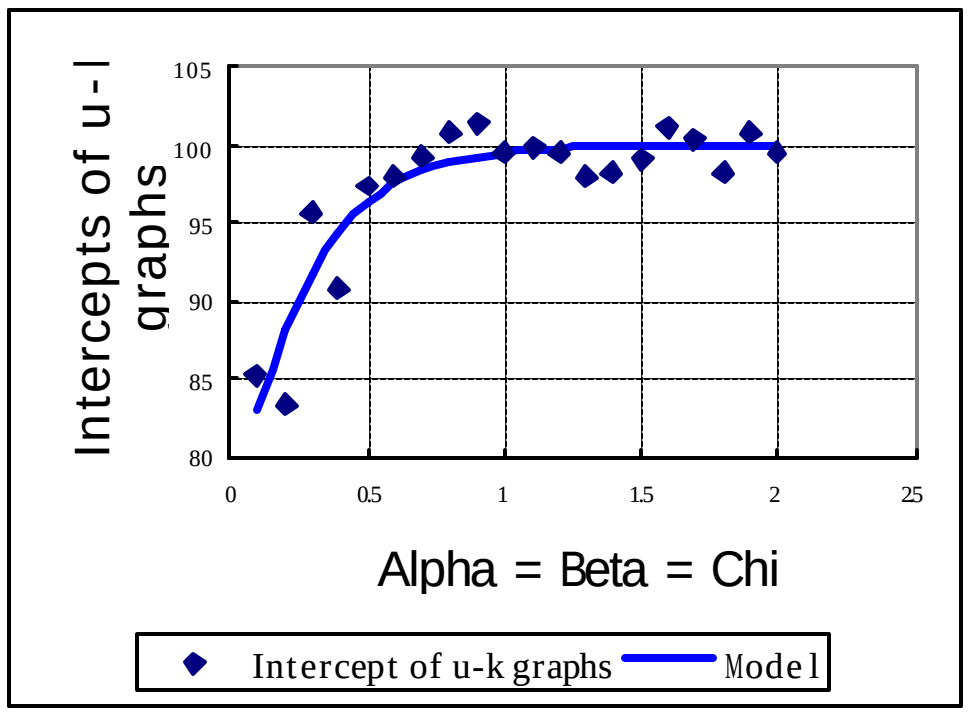

$$Y = \frac{99.979}{1+0.31.e^{-4.191X}},\ \text{SSE} = 77.71$$

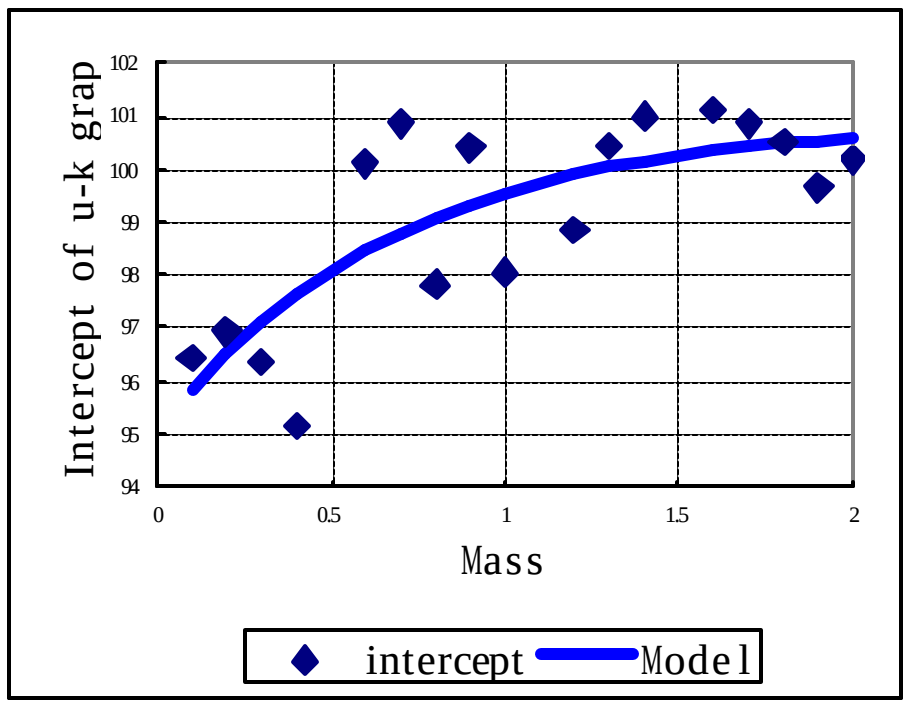

$$Y = \frac{100.87}{1+0.0617.e^{-1.53785X}},\ \text{SSE} = 23.10$$

**Figure 5-7 Sensitivity of Parameters Toward Average Speed and u-k Graphs**

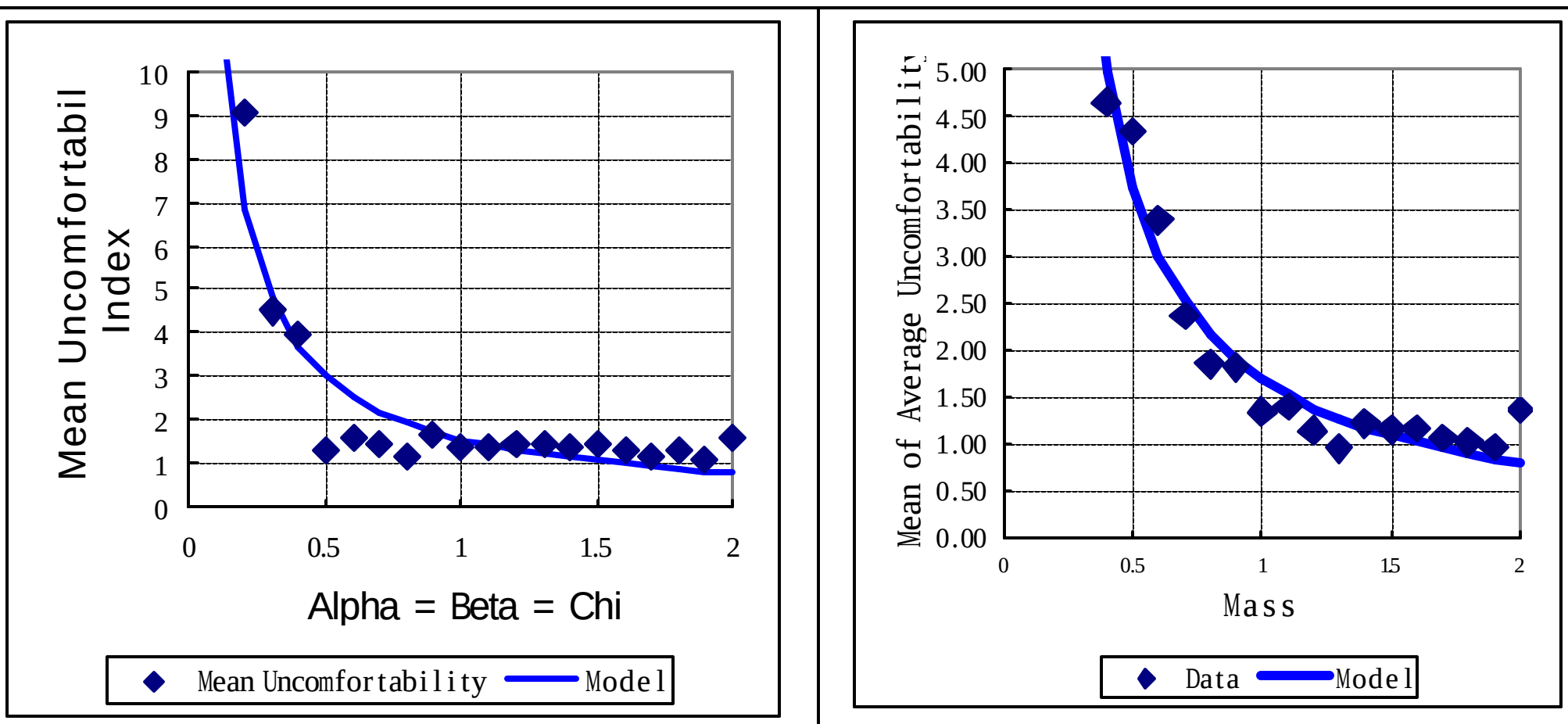

$$Y = \frac{0.093756}{1 - 0.998.e^{-0.060X}}, \text{ SSE} = 11.175$$

$$Y = \frac{-0.90148}{1 - 0.9991.e^{-0.0096X}}, \text{ SSE} = 1.381$$

**Figure 5-8 Sensitivity of Parameter toward Uncomfortability Index**

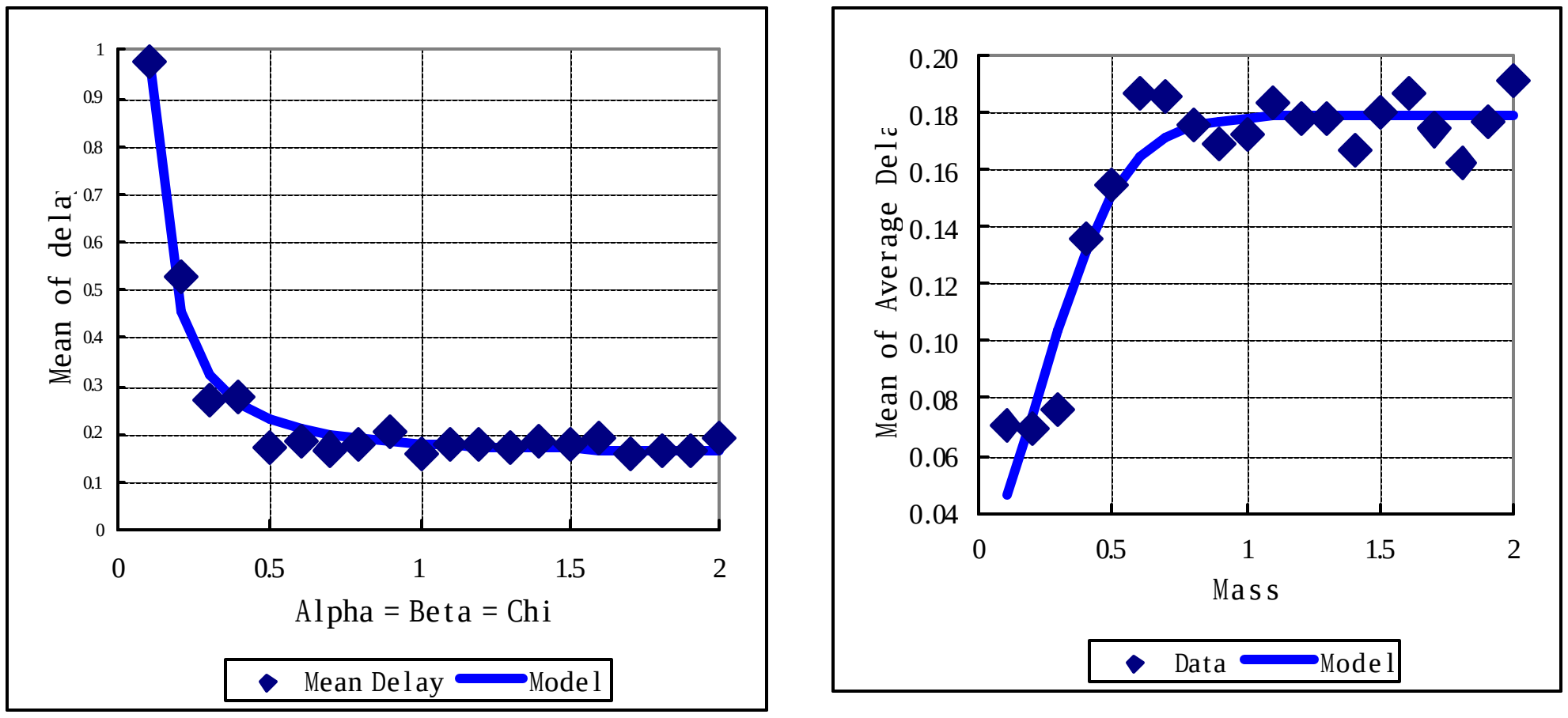

$$Y = \frac{0.166}{1 - 1.087.e^{-2.691X}}, \text{ SSE} = 0.01597$$

$$Y = \frac{0.1794}{1 - 5.7471.e^{-6.925X}}, \text{ SSE} = 0.00291$$

**Figure 5-9. Sensitivity of Parameters toward Average Delay**

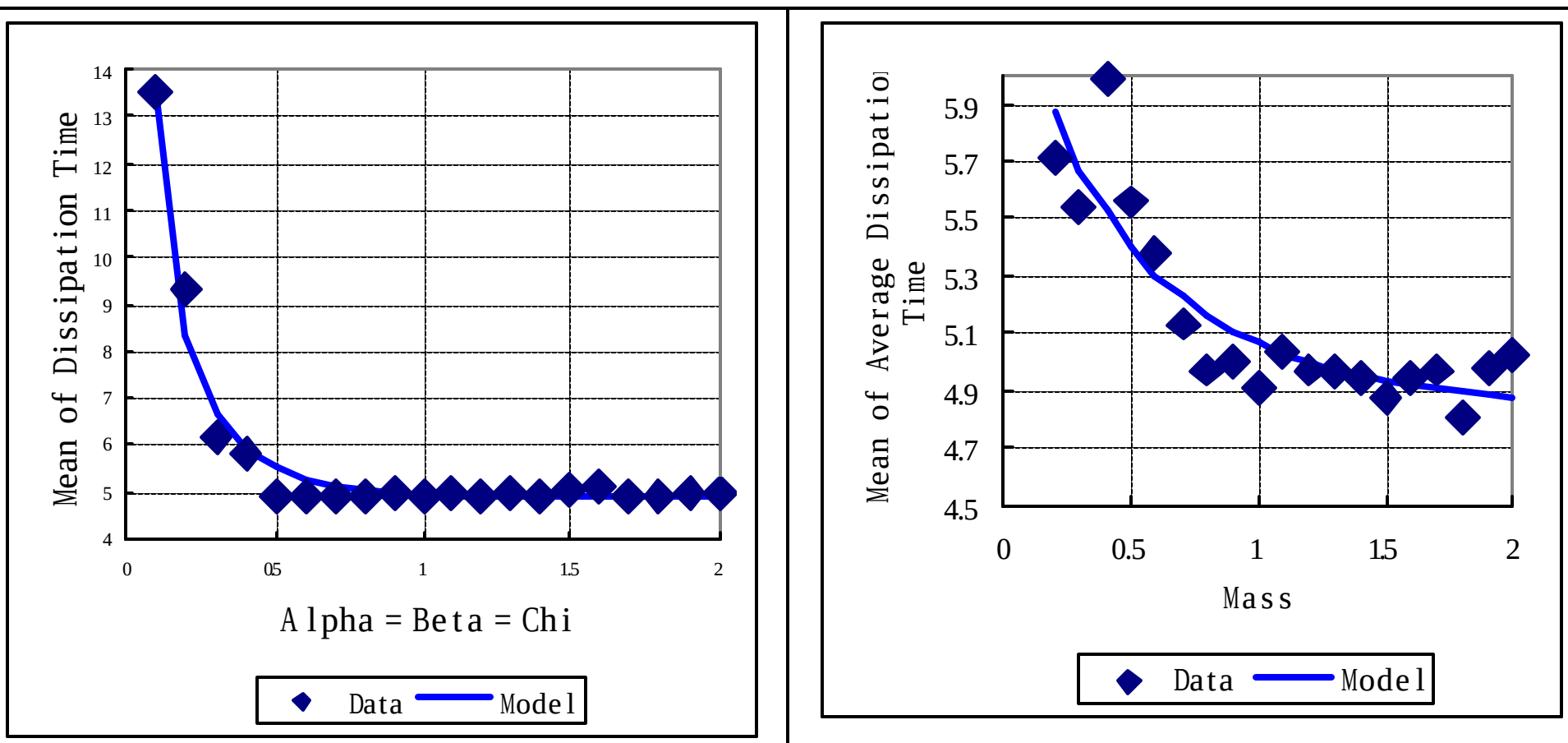

$$Y = \frac{4.8654}{1 - 0.9926.e^{-4.326X}}, \text{ SSE} = 1.808$$

$$Y = \frac{4.838}{1 - 0.246.e^{-2.691X}}, \text{ SSE} = 0.41$$

**Figure 5-10. Sensitivity of Parameters toward Dissipation Time**

### 5.3.4 Dissipation Time

Similar to the delay, the increase of parameter alpha, beta and chi together makes the dissipation decreases exponentially until the parameters reach one. The mass however, will reduce the dissipation time linearly. Figure 5-10 exhibits these behaviors.

### 5.3.5 Sensitivity of $\Delta t$

The change of value of the $\Delta t$ have no significant effect on the average instantaneous speed but smaller value of $\Delta t$ (bigger Hertz) tend to reduce the average instantaneous delay and increase the average instant uncomfortability index. Figure 5-11 shows the effect of $\Delta t$ toward speed, delay and uncomfortability.

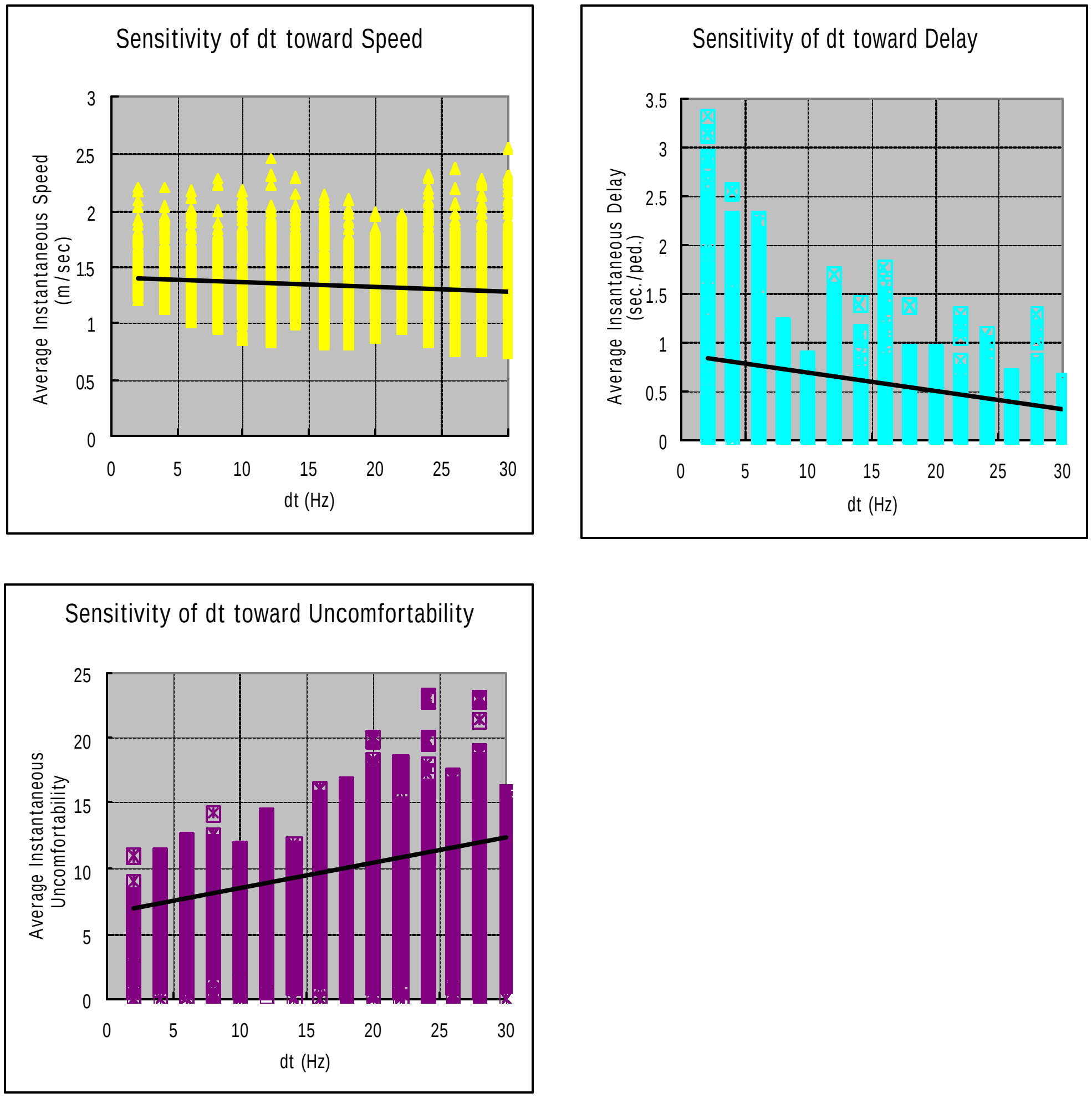

**Figure 5-11 Sensitivities $\Delta t$ toward Speed, Delay and Uncomfortability**

## 5.4 RELATIONSHIPS BETWEEN VARIABLES

Aside from the sensitivity of the parameters, the relationships between variables have interesting results that can be obtained from the simulation model. In this section, first the relationship of the density (total pedestrian) with other prominent variables is described. After that, the relationship of those variables with the average speed is also pointed out.

### 5.4.1 Relationship with Density

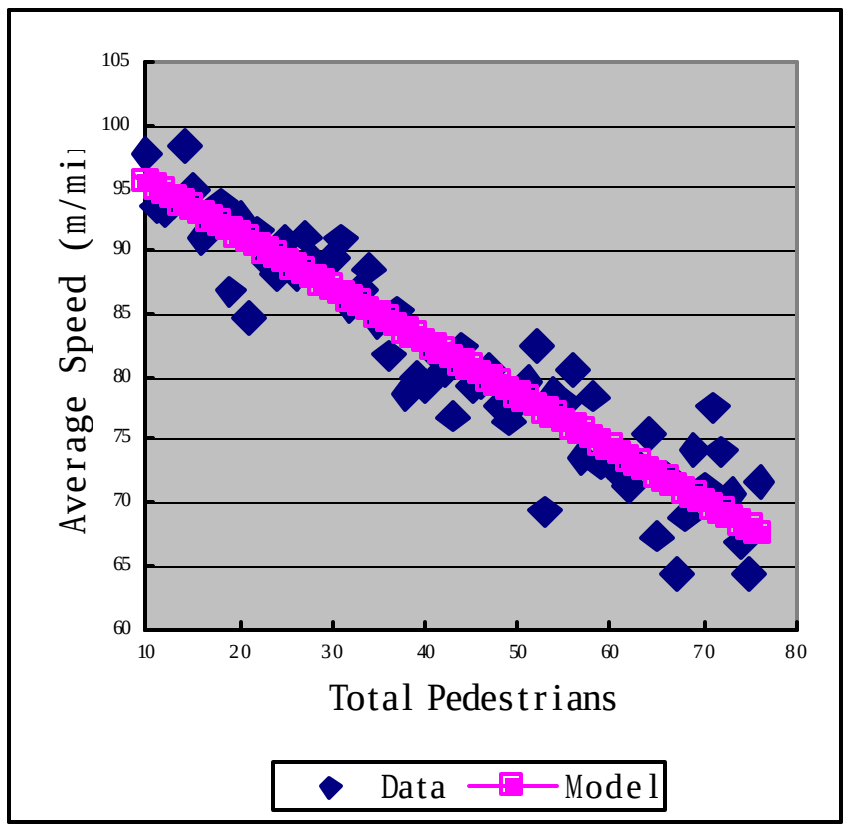

$$Y = 99.66 - 0.4236X \ , r^2 = 0.89$$

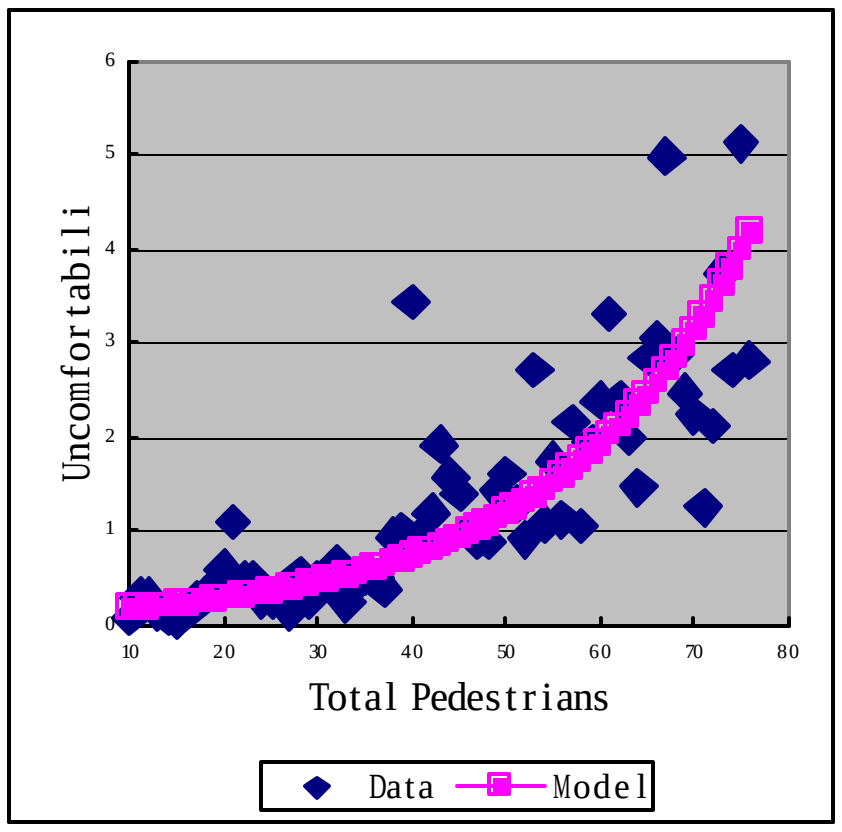

$$Y = 0.1057.e^{0.0484X} \ , \ r^2 = 0.76$$

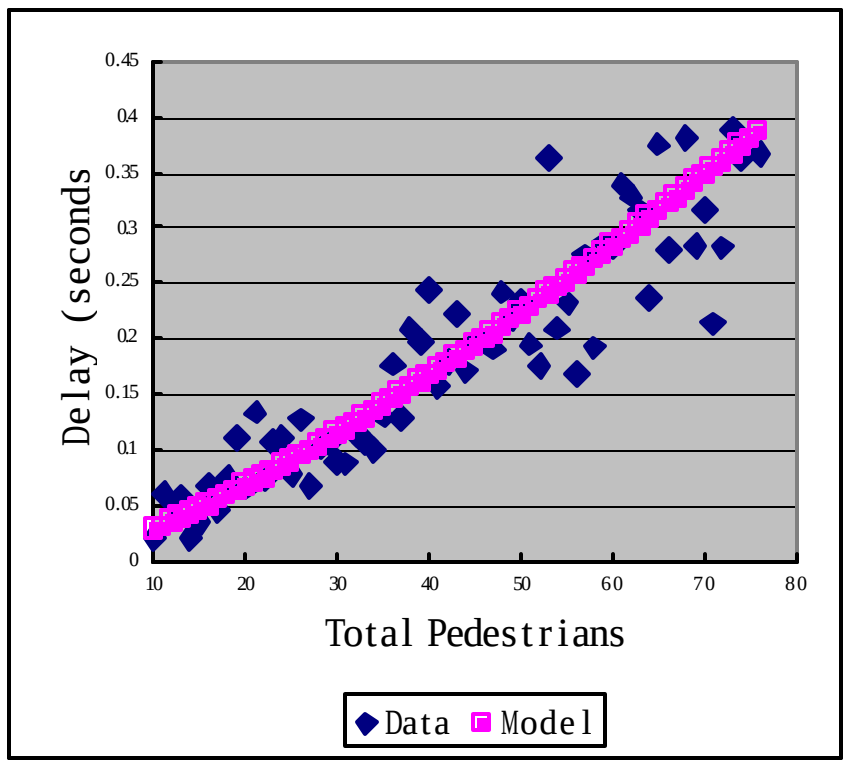

$$Y = 0.001276 \ X^{1.3194}, \ r^2 = 0.81$$

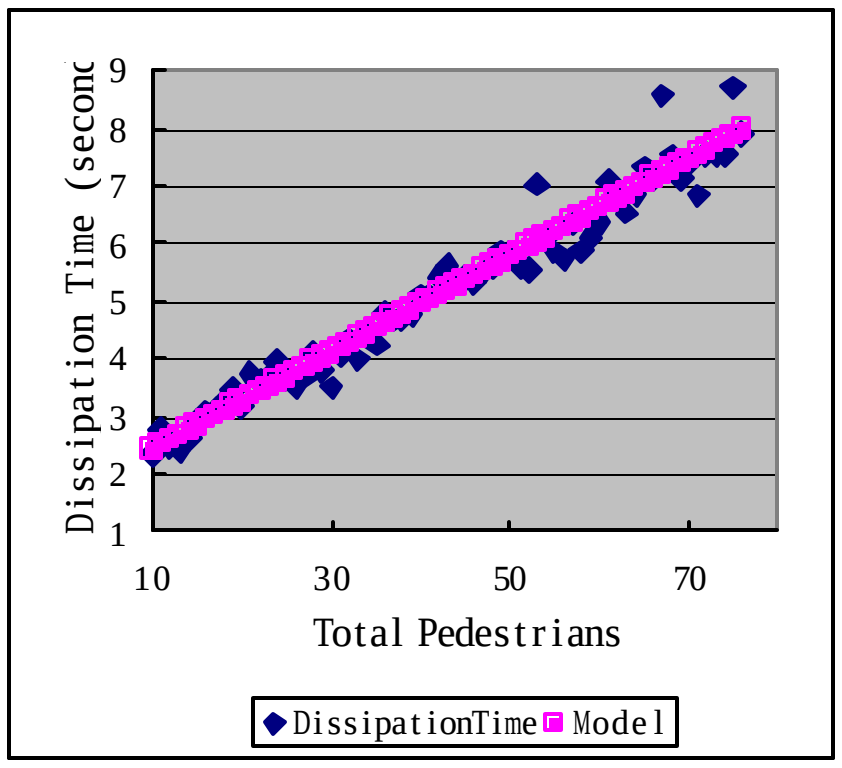

$$Y = 1.5764 + 0.084 \ X \ , \ r^2 = 0.96$$

**Figure 5-12 Relationships of Pedestrian Characteristics with Density**

Figure 5-12 shows the relationship of the prominent flow performances with the density that is represented as the total pedestrians. The common relationship of pedestrian speed and density is linear as illustrated in the top-left figure. Increase in the pedestrian density tends to increase the uncomfortability exponentially, as well as exponential increase in delay while dissipation time increases linearly. It should be noted that though the model of the delay is almost similar to the linear model, the best fit is revealed exponential. All the models have significant correlation values and the t-statistics are high.

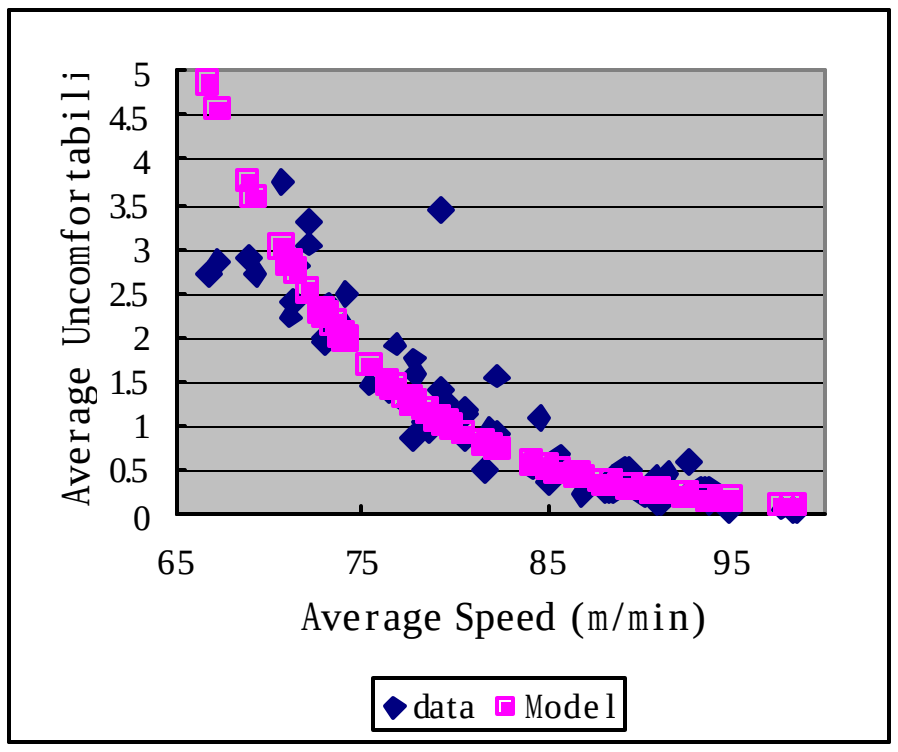

$$Y = 15694.37.e^{-0.12107X}, r^2 = 0.85$$

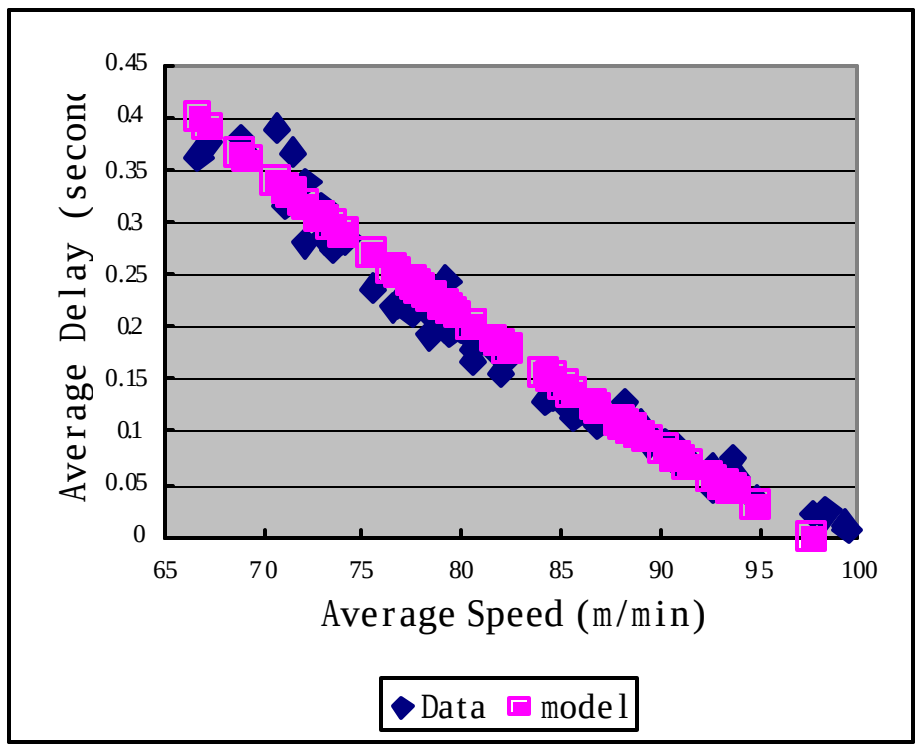

$$Y = \ln(121.01.X^{-1.0468}), r^2 = 0.94$$

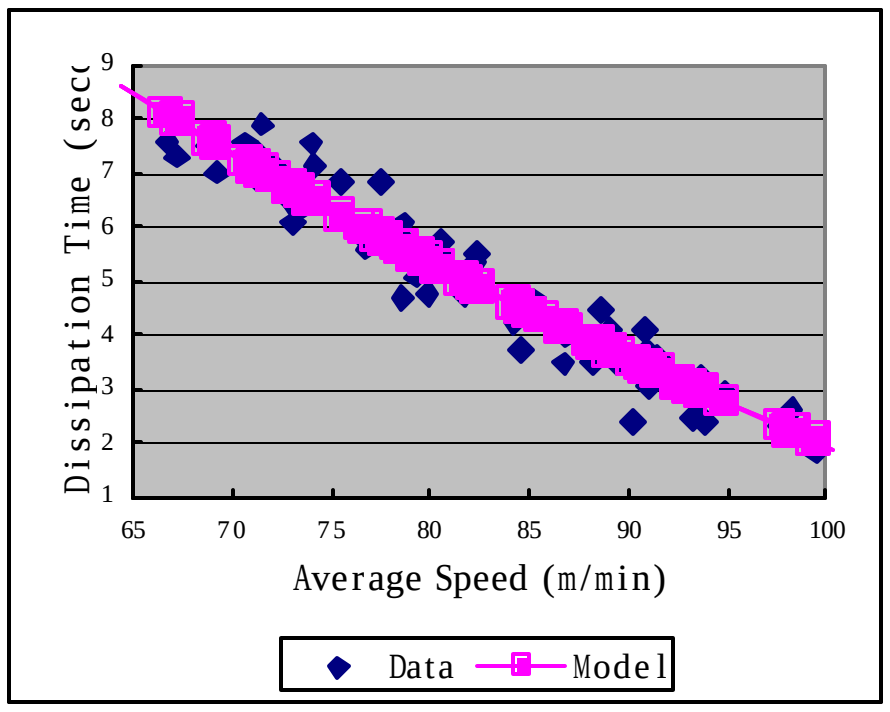

$$Y = \ln(1.416 * 10^{31}.X^{-15.1477}), r^2 = 0.94$$

**Figure 5-13 Relationships of Pedestrian Characteristics with Average Speed**

### 5.4.2 Relationship with Average Speed

Besides the relationship with the density, the relationship of average speed with the other three prominent pedestrian characteristics is interesting to be investigated. Figure 5-13 illustrates these relationships. As the average speed increases, the uncomfortability is likely to decrease exponentially while the delay and dissipation-time tend to decrease logarithmically.

## 5.5 TOWARD REAL WORLD DATA

This section describes the calibration and the validation of the simulation using the real world data. The calibration is concerned with the determination of the numerical value of the parameters and the results of the simulation. The validation is done to see whether there is an adequate agreement between the model and the system being modeled.

**Table 5-2 Parameters and Input Variables Default Values**

| | | |
|---|---|---|
| Parameters | ▪ Mass | 0.750 |
| | ▪ Alpha | 0.205 |
| | ▪ Beta | 0.001 |
| | ▪ Chi | 0.250 |
| Input Variables | ▪ Pedestrian diameter | 0.60 meter |
| | ▪ Influence diameter | 1.67 meter |
| | ▪ Time slice | 1/15 second |
| | ▪ Maximum Acceleration | 1.750 m/sec.$^2$ |
| | ▪ Maximum Speed (mean, std. deviation) | 1.775 m/s, 0.30 m/s |
| | ▪ Sight distance | 4.00 meter |
| | ▪ Pedestrian Trap size | 12 x 32 meter |
| | ▪ Number of way | 2 |
| | ▪ Generator Space (mean, std. deviation) | 50%, 10% |
| | ▪ Generator time | at once before simulation |
| | ▪ Distance to the nearest generator | 21 meter |
| | ▪ Total number of pedestrians generated | 300 |

### 5.5.1 Model Calibration

Before simulation is performed, the parameters and several input variables must be set. This section describes the determination of setting of these input and the effect of each input variable toward the performance of the simulation. Table 5-2 displays the list of parameters and input variables that should be determined before the simulation and their default value. The default value is set in the program and is used for the experiments throughout this dissertation unless stated explicitly.

As described in the Sensitivity section, the parameters Mass, Alpha, Beta and Chi have bigger influence, as the values are smaller than one. These phenomena happen because in the formulation of the intended velocity, the parameter is set in the denominator. As the parameter values are between zero to one, forces are getting stronger due to parameters value. If the value of any parameter is higher than one, the force influence by that parameter is smaller. The higher the value of parameters Alpha, Beta and Chi, in general will make a smaller delay; uncomfortability and dissipation time but will increase the average speed. While the higher the value of the parameter Mass tends to reduce the average speed, uncomfortability and dissipation time but increases the delay. In a more microscopic way, the parameter Alpha influences the force to move forward, Beta for collision avoidance and Chi for repulsing away.

To determine the default values of the parameters and the input variables, three considerations are utilized:

1. The pushing effect
2. The collision or overlapping effect
3. The real world value or range

When pedestrians from the two-way system meet each other, with certain setting of parameters and input variables, “stronger” pedestrians tend push back the “weaker” pedestrians. Pedestrians who have stronger positions are the ones who have greater maximum speed, or higher number of pedestrian as a group as they move together. For example, three or four pedestrians move at the same speed, may push back a single pedestrian from the other way that has a higher speed. The same number of pedestrians meeting each other, the ones with higher maximum speed may push back the pedestrian from the reverse direction that have smaller maximum speed.

On the other hand, a different set of parameters and input variables tend to make the pedestrians collide or overlap each other (which is unrealistic behavior). Setting a small value for the Influence diameter or maximum acceleration may produce this effect regardless of the value of the parameters.

Interestingly, setting a parameter of input variable in one extreme (i.e. smaller) will make the pushing effect higher while collision effect is smaller. Setting the parameter or the input variable in the other extreme (i.e. higher) tend to make the collision effect higher and reduce the pushing effect, as illustrated in Figure 5-14. For example, a bigger maximum acceleration makes greater pushing effect and smaller maximum acceleration will make pedestrians collide into each other. Unfortunately, both effects disappearing is a very rare case. Thus, a balance or minimizing both effect of collision and pushing is used to determine the value of the parameters or the input variables.

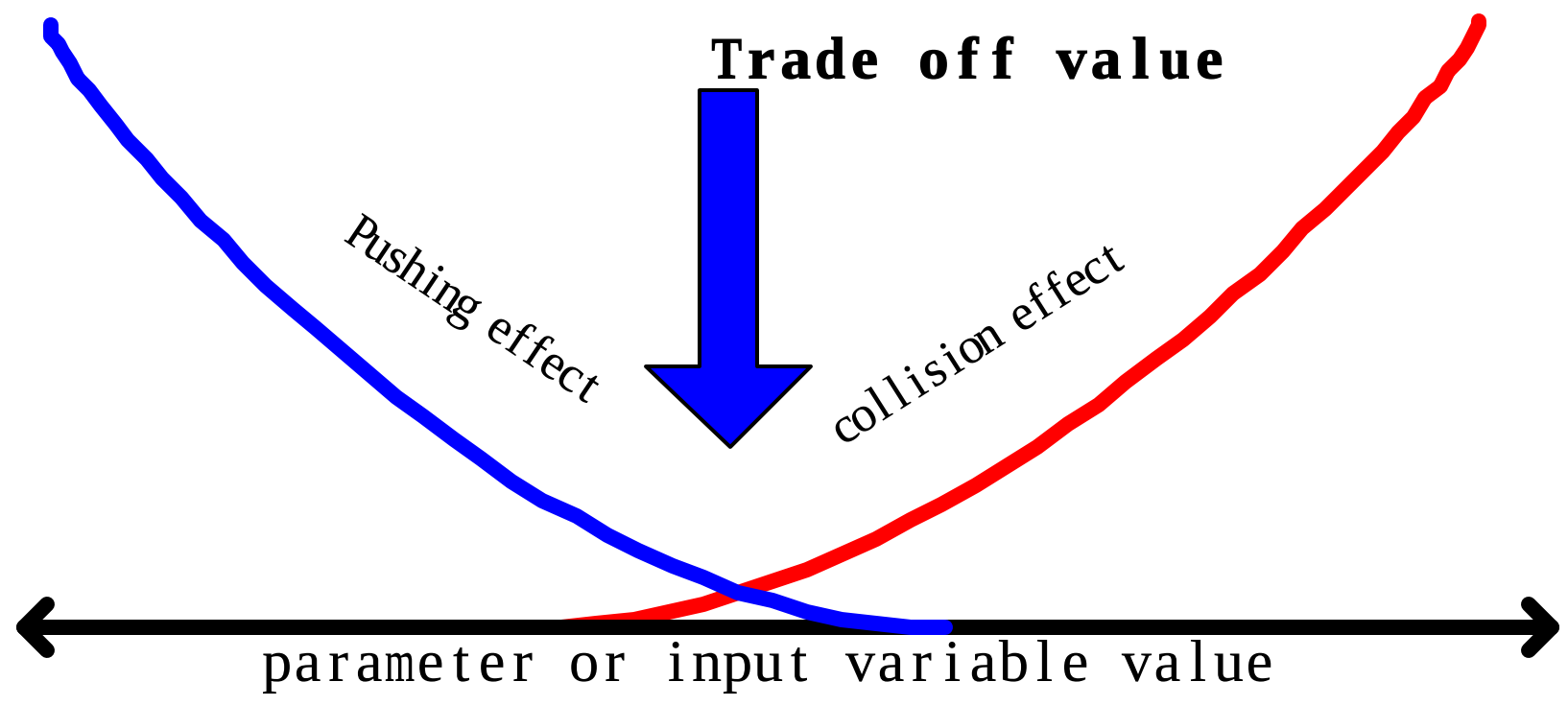

**Figure 5-14 Trade-off of Pushing Effect and Collision Effect to Determine the Parameter or Input Variable Values**

The real world value of the input variables, as described in last section of Chapter 3 provide important hints for the value or the range of the input variables of maximum speed, maximum acceleration, total number of pedestrians, number of ways and pedestrian trap size. The specific value of these input variables including the parameter setting is done to minimize the error between mean and standard deviation of the speed distributions. The next section in this chapter on Validation will demonstrate the comparison between the simulation and real world results.

The following list is the parameters and input variables that is determined by the trade off of collision and pushing effect after the other variables are set using the real world value: Mass, Alpha, Beta, Chi, pedestrian diameter, influence diameter, and sight distance. The trade-off values between pushing and collision effects give a range of values. Within these range, the values of the parameters and input variable are determined to minimize the error between speed distributions

The shoulder-to-shoulder length that has been measured by the author reveals that the average length (i.e. diameter) of a human body is about 60 cm. Based on this value the default of pedestrian diameter is established.

The value of time slices or $\Delta t$ is determined based on practical value. A smaller this value will produce a better result but take the simulation time much longer. Higher than 1/15 second, makes the pedestrians jump and the collision effect is very high. No pushing effect is produced by the smaller value of the time slice.

The pedestrians always have an erratic behavior for the first 5 second after it is generated due to random initial velocity value and positioning and searching the best position to go to the destination. Due to this fluctuation behavior, the distance to the nearest generator is set so that the erratic behavior will not be seen. Generator Space (mean, std. deviation) is set in the middle of the pedestrian trap and the variation is set to produce a most efficient simulation. A higher variation of the generator space will not make all pedestrians generated pass the pedestrian trap. Some of pedestrians may pass only outside of the trap and this simulation result will not be counted. The generator time generated before the simulation begins to resemble the pedestrian crossing data where all pedestrians are waiting for crossing during the red time and the start of the simulation is similar to the start of green time in real pedestrian crossing. The pedestrian trap is in the middle of the walkway for stability purposes of the simulation. When pedestrians are generated too near or overlapping each other, they need some time to be stable by moving themselves to a better position automatically. The pedestrian characteristics are gathered only when pedestrians are in the pedestrian trap.

### 5.5.2 Validation

The validation step ensures that the simulation model behaves as expected. One way to inspect this behavior is the decline of the average speed as the density increases. The previous sections on the sensitivity analysis and the relationship between variables have proven that this behavior is guaranteed. The regression results have strongly revealed not only the behavior of the declining average speed as the density increases, but it even showed that this relationship is linear with a very high correlation.

To set the default values of the parameters and the input variables, several methods were performed and failed. The closest method to validate the simulation using the real world data is to minimize the difference between the speed distributions. Since the distribution of the speed can be assumed normal, the mean and standard deviation are the basis for the comparison.

First, the balance between pushing and collision effect gives a range of values. Then, within these range, the values of the parameters and input variable are determined to minimize the error between mean and standard deviation of the speed distributions. This range dynamically changes as the other input variables also change. Trial and error method was performed to produce the range and then to make the difference of mean and standard deviation of speed between the simulation and the real world data as minimum as possible. After the parameters and the input variables produces the simulation results that have minimum pushing and collision effect, then the maximum speed (mean and standard deviation) and the maximum acceleration variables are adjusted to produce almost the same distribution. The best-fit distribution is given in Figure 5-15. Table 5-3 shows the result of the t-test between two samples of simulation and the real world data. It reveals that the two distributions are the same. The parameters and the input variables that produce this best fit distribution are determined as the default values of the simulation, as shown in Table 5-2.

**Table 5-3. Result of t-Test: Two-Sample Assuming Unequal Variances**

| Instantaneous Speed | Simulation | Real World |
|---|---|---|
| Mean | 1.385 | 1.363 |
| Variance | 0.061 | 0.050 |
| Observations | 1091 | 119 |
| Hypothesized Mean Difference | 0 | |
| Degree of freedom | 151 | |
| t Stat | 0.985 | |
| P(T<=t) one-tail | 0.163 | |
| t Critical one-tail | 1.655 | |
| P(T<=t) two-tail | 0.326 | |
| t Critical two-tail | 1.976 | |

**Table 5-4 Statistics of the Simulation Performance**

| | Mean | Median | Mode | std Dev | Range | Minimum | Maximum |
|---|---|---|---|---|---|---|---|
| Speed | 1.39 | 1.36 | 1.72 | 0.25 | 1.51 | 0.96 | 2.46 |
| Delay | 0.56 | 0.48 | 1.43 | 0.43 | 1.48 | 0.00 | 1.48 |
| Uncomfortability | 9.70 | 11.65 | 12.47 | 4.19 | 14.08 | 0.42 | 14.50 |
| Density | 0.09 | 0.07 | 0.02 | 0.07 | 0.22 | 0.00 | 0.22 |

Total number of pedestrians =1091.

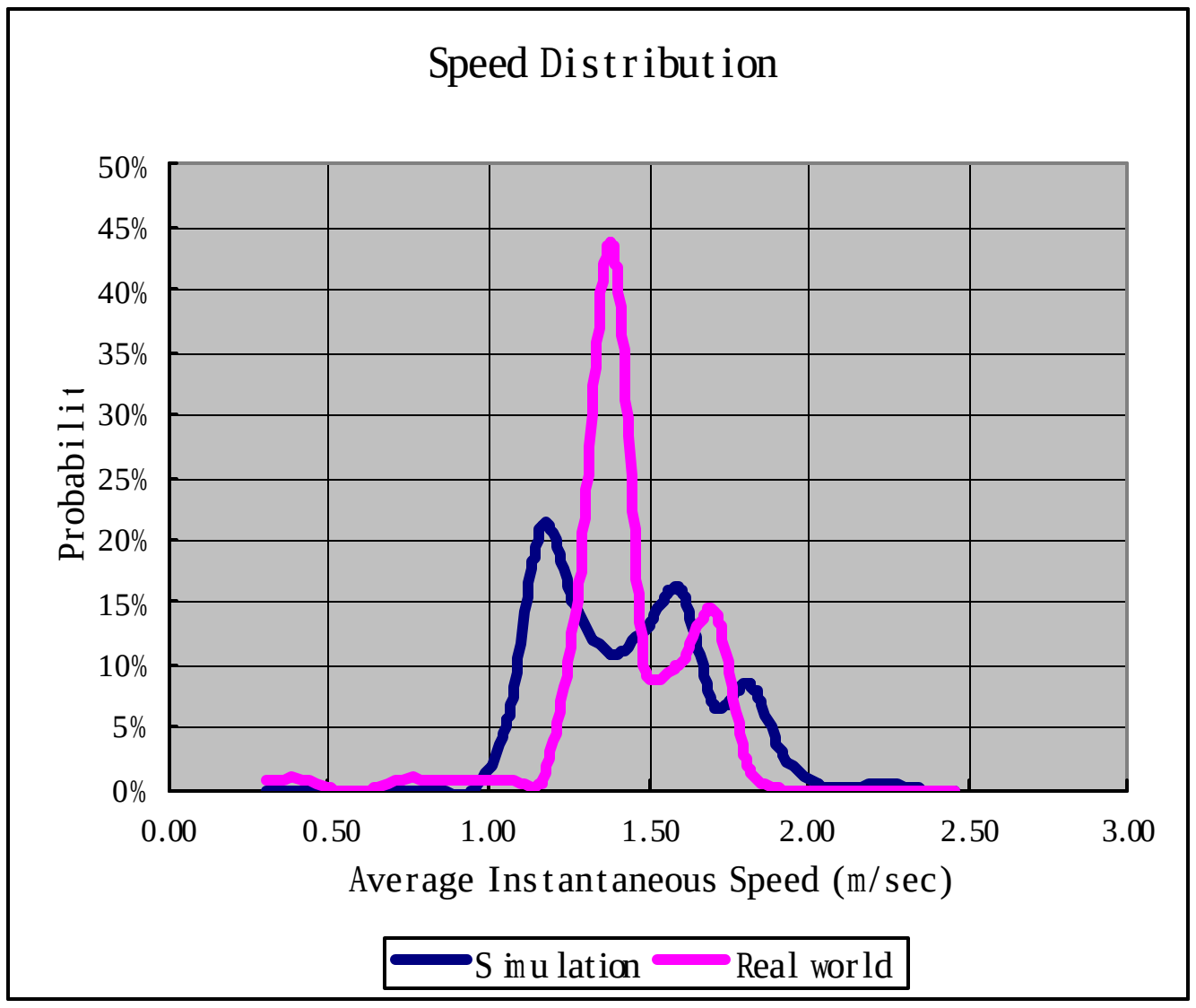

**Figure 5-15 Comparison of Average Speed Distribution**

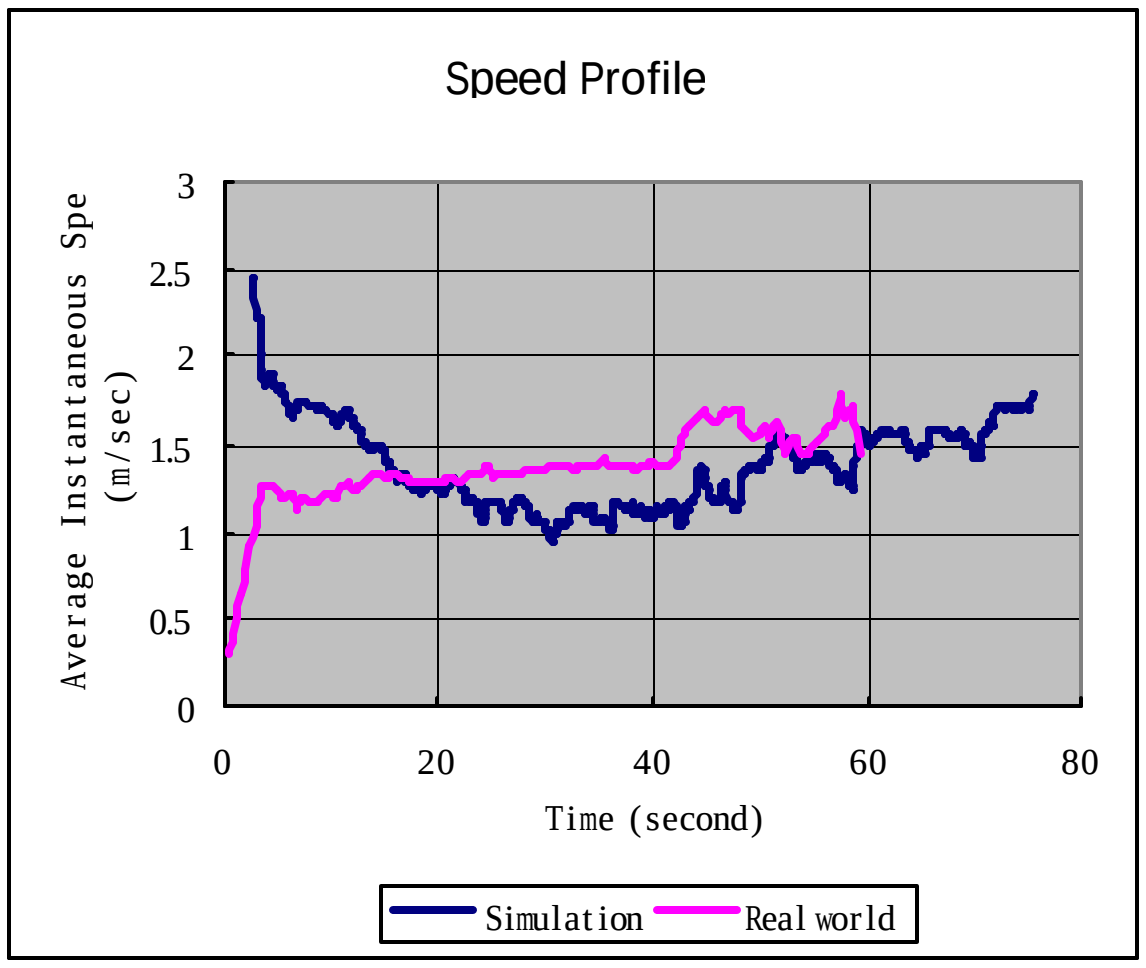

**Figure 5-16 Comparison of Average Speed Profile**

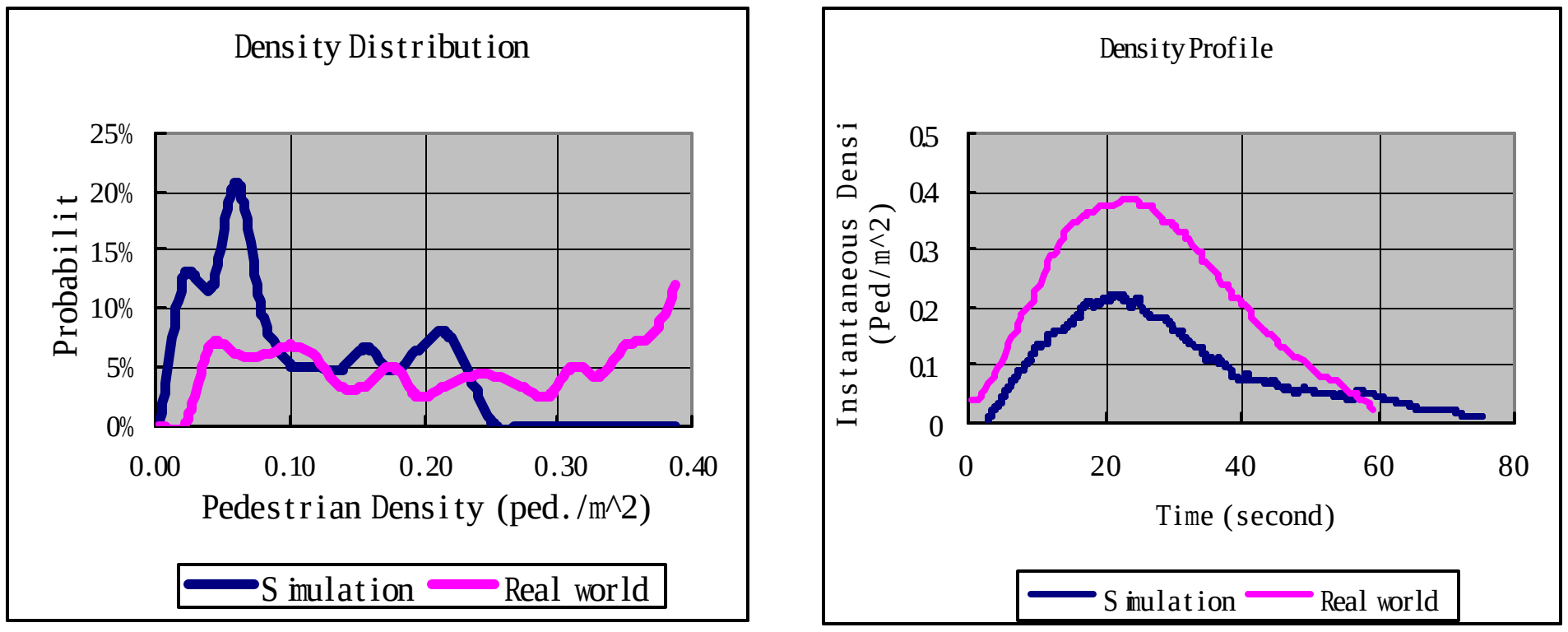

**Figure 5-17 Comparisons of Density Profile and Its Distribution**

Figure 5-17 shows the distribution and profile for density. Though the number of pedestrian generated by the simulation is quite high (300 pedestrians), not all of them will pass the pedestrian trap and produce lower density for the simulation compared to the real world data. Increasing the total number of pedestrian will make the speed distribution change. The shape of the density distribution and the density profile are almost the same. The simulation density contains more of the lower density compare to the real world density.

## 5.6 LANE FORMATION SELF ORGANIZATION

Both simulation and real world pedestrians who are crossing show that they automatically create a lane formation while they are walking. Figure 5-18 shows the velocity diagram of pedestrian from the real world data with a raster of the lane formation. Though the velocity of each pedestrian is not the same, pedestrians prefer to follow other pedestrians rather than make their own path. This microscopic behavior happens because the pedestrian tends to reduce their interaction effect, especially with a pedestrian from a different way.

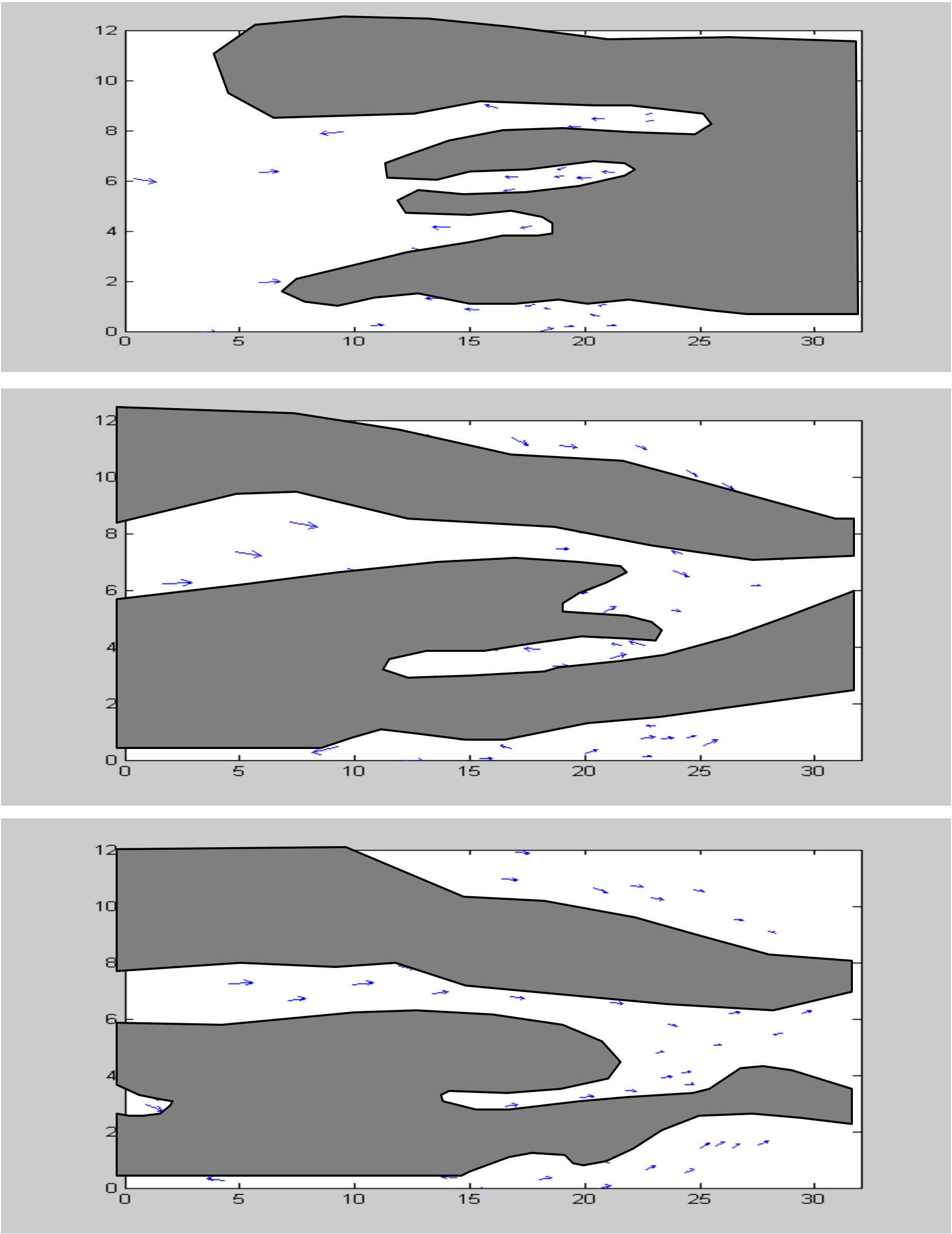

**Figure 5-18 Self Organization of Lane Formation (at 35, 40 and 43 seconds)**

Interestingly, with simple forces as explained in Chapter 4, the simulation also produce lane formation of self-organization similar to the real world phenomena. Further exploration was done to investigate what factors produce the self-organization of lane formation. It was found out that the most influence factors toward the lane formation are the value of total number of pedestrians, maximum acceleration and parameter chi.

The total number of pedestrians will influence the density and higher density is likely to make the lane formation. Small number of pedestrians may also form the lane formation but it is difficult to be manually distinguished. The effect of density is more to emphasize the lane formation rather than making a new type of formation.

The lane formation can be seen using the trace path of the velocity. If the absolute value of chi is less than one, then all pedestrians tend to move *together* and make a wider lane. If the absolute value of chi is bigger than one, a higher value of chi, the pedestrians tend to separate and tend to move as individual. If chi is positive, the pedestrians tend to move away to the left and if chi is negative, the pedestrians tend to move away to the right. Higher maximum acceleration tends to emphasize the effect of Chi, but also tends to increase the pushing effect. The other non-dominant factors to develop the lane formation are the maximum speed variation (i.e. standard deviation) and the variation of the spatial generators. The mean of maximum speed and the sight distance has no effect on the self-organization. Using uniform or small variation of the maximum speed, the lane formation is made by a group of pedestrians (two or three pedestrians move in almost parallel per lane). Thus, make the lane width is wider. In the other extreme, if the maximum speed is much dispersed (the variance of maximum speed is very high compared to the mean), the lane will be made by only single pedestrians who move in lines. This happens because the dispersed maximum speed of pedestrians will make them to walk as individuals rather than as a group.

The standard deviation of the spatial generator also has a little influence over the lane formation self organization: as the standard deviation of spatial generator increases, the width of the generator widens and of course the width of lanes increases.

# CHAPTER 6 APPLICATION OF THE MICROSCOPIC PEDESTRIAN STUDIES

The simulation model that has been calibrated and validated as explained in the previous chapters is now ready to be applied for some experiments on a hypothetical situation to gain more understanding of the pedestrian behavior or to evaluate some policies on some design. This chapter provides an insight regarding the possible application of the microscopic pedestrian studies that has been described. Three microscopic simulation projects were developed based on the model. The first project compares one way and two - way pedestrian traffic to gain more understanding about their characteristics. The second project is a theoretical work to know the behavior of the system if the number of elderly pedestrian increases. The third project is a more practical one. It proposes a policy of lane-like segregation towards pedestrian crossing and inspects the performance of the crossing.

## 6.1 BEHAVIOR OF ONE AND TWO WAY PEDESTRIAN FLOW

Microscopic pedestrian studies are more comprehensives compared to the macroscopic studies because microscopic pedestrian studies consider pedestrian interaction. Pedestrian interaction is the repulsive and attractive effect among pedestrians and between pedestrians with their environment. Since the movement quality of pedestrians can be improved by controlling the interaction between pedestrians, better pedestrian interaction is the objective of this approach.

This section demonstrates the effect of pedestrian interaction through a simple comparison of one way and two-way pedestrian traffic. Though it seems obvious that one way pedestrian is better than two way due to their interaction, however, how if the pedestrian traffic density or the total pedestrians in the system are kept constant for both system, will the interaction affect their characteristics? If it so, how can the characteristics of the two way pedestrian traffic flow compare to the one way system? Will the speed – density relationship of both systems be just similar or will it change due to the interaction?

An experiment is designed to answer those questions. Figure 6-1 shows the schematic diagrams of the experiment. The total number of pedestrian or the pedestrian density (the area of the trap is constant) is kept the same for both scenarios. The maximum speeds and the total pedestrians vary as the experimental variables.

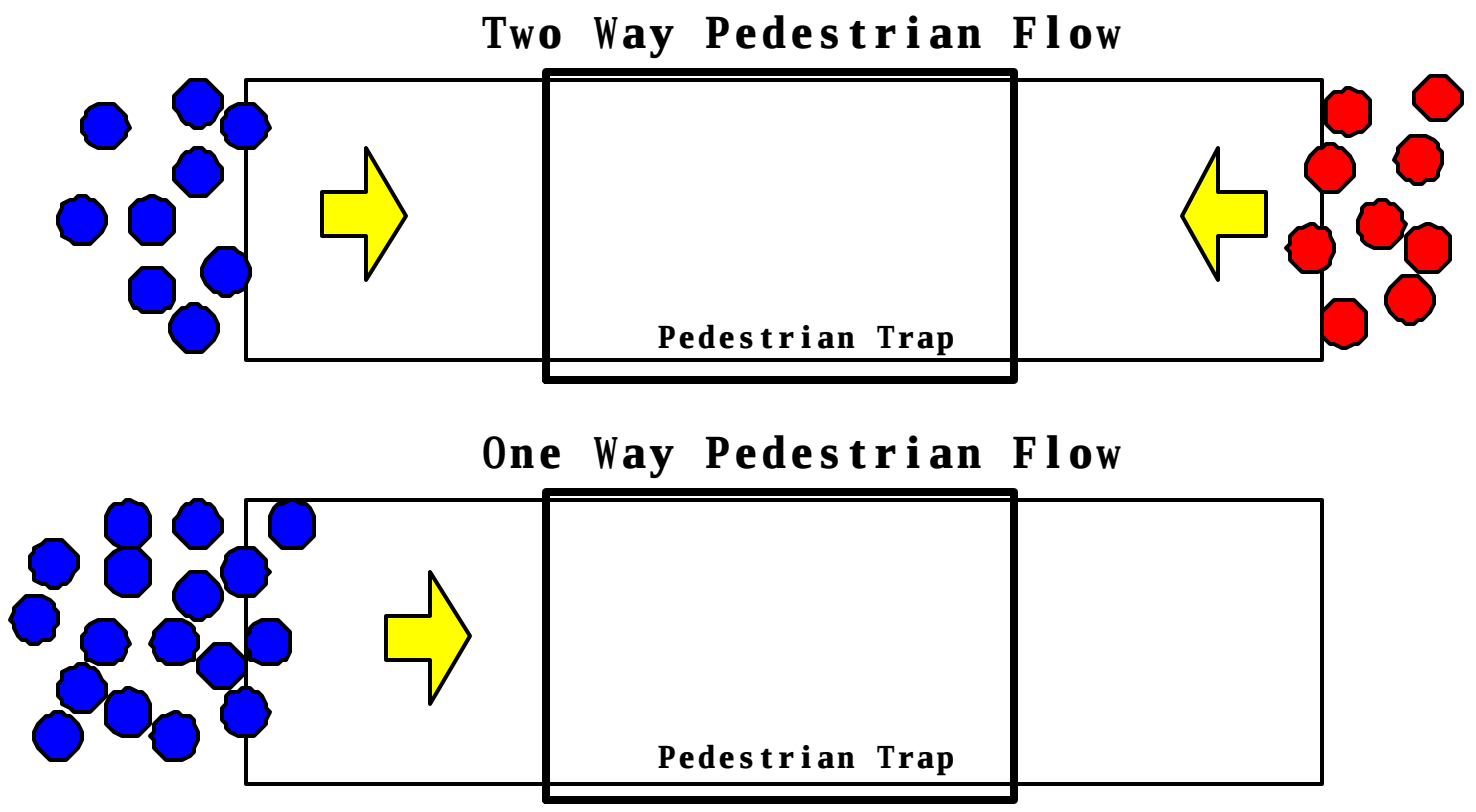

**Figure 6-1. Experiments to Obtain Behavior of One and Two Way Pedestrian Traffic**

Figure 6-2 presents the relationship between the total pedestrian or density with the average speed (u-k graphs). The model and the r-square value are also given in the graphs. The speed density relationship for one-way pedestrian traffic is linear as have been explained in Chapter 5, but the speed-density relationship for a two way pedestrian traffic is found to be logarithmic. The gradient of u-k graphs for a two way pedestrian traffic declines as the number of the pedestrian in the system increases. When the density is high, the gradients of both scenarios are almost the same. This finding is rather interesting since most of pedestrian studies (refer to chapter 2, table 2-2) assume that the speed density relationship is linear. The average speed of the two-way pedestrian traffic dropped up to one third

compared to one-way traffic.

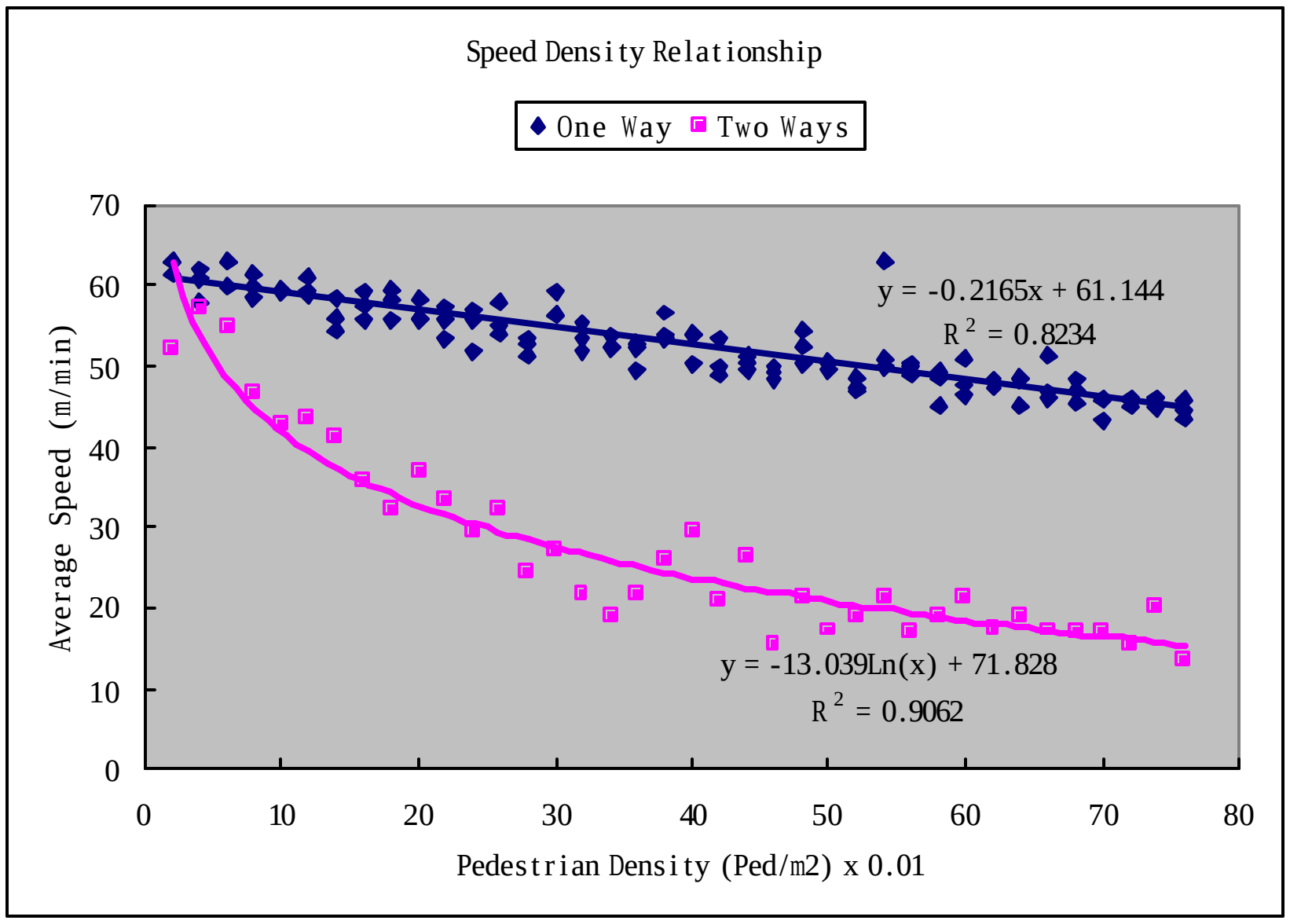

**Figure 6-2. Speed Density Relationship of One and Two Way Pedestrian Traffic**

Since the speed-density relationship of two-way pedestrian traffic is logarithmic (compared to the linear relationship of one way traffic), it is interesting to further see the effect of the maximum speed.

Figure 6-3 shows these relationships. Similar to one-way traffic, the u-k graphs of a two way pedestrian traffic with higher maximum speed is on the top of the lower one. It is interesting to note that the lines may intersect each other when the pedestrian density is very high. The gradient and intercepts of the graph change as the maximum speed changes. The scattered data is added to show the variation of the data toward the model. The eight categories (as shown in the legend) of maximum speed, which intersect the vertical axis, represent the free flow speed.

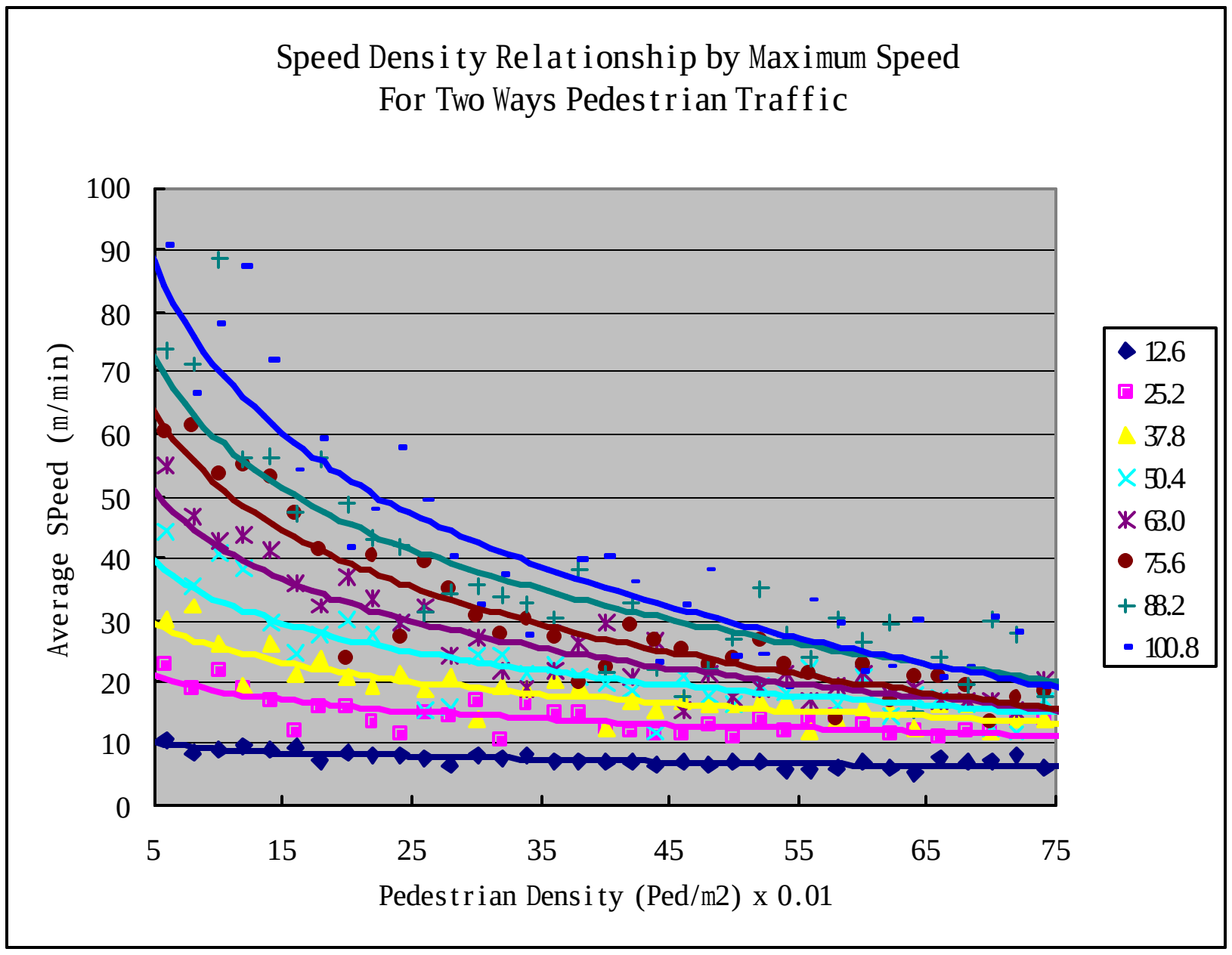

**Figure 6-3 Speed Density Relationship of Two-Ways Pedestrian Traffic for Varies of Maximum Speed**

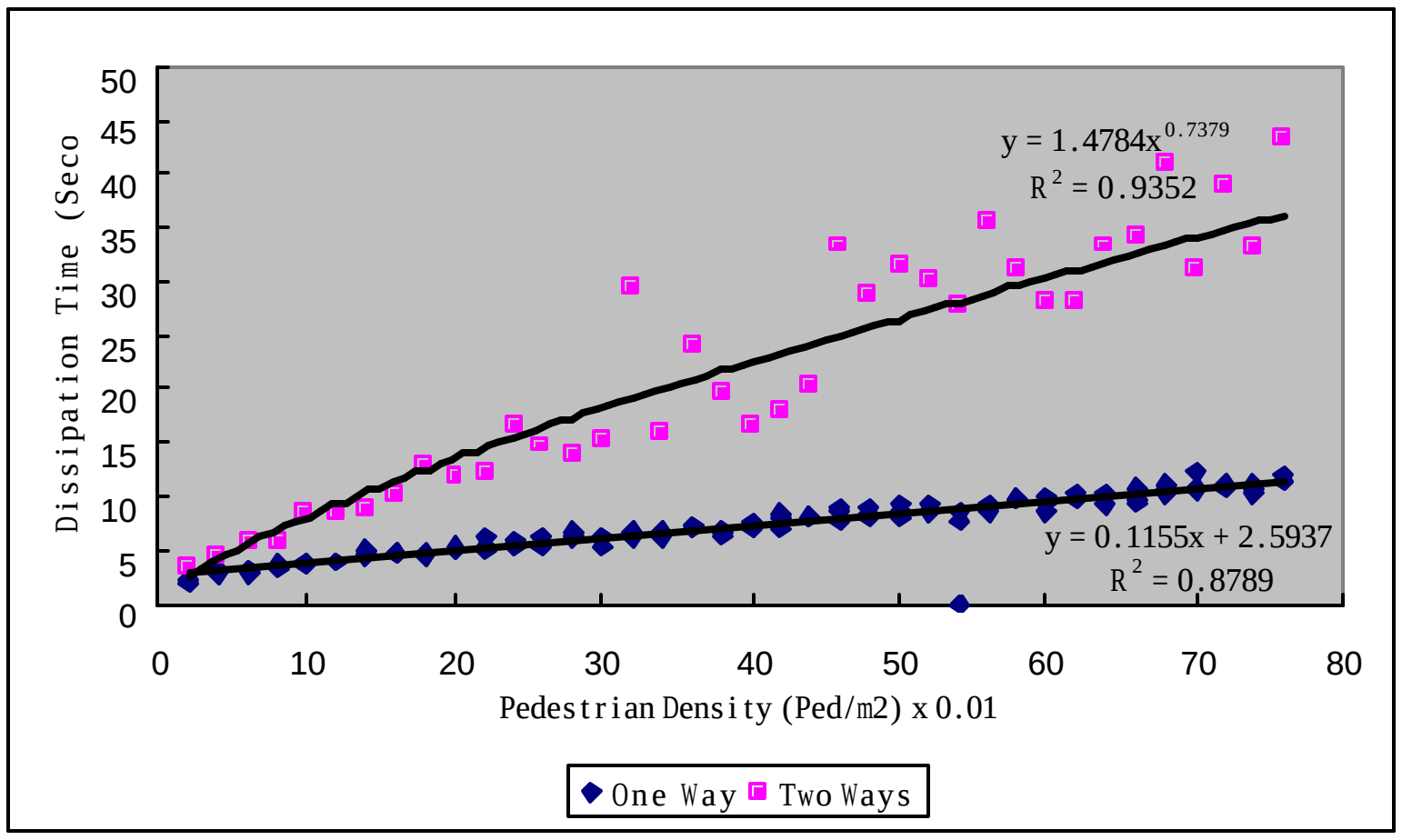

**Figure 6-4 Dissipation Time of One and Two-Ways Systems**

The dissipation time of one-way system is linear while for the two-ways is 1.5 times with the power of 0.8 as displayed in Figure 6-4.

All these results show that one way pedestrian traffic is much better than the two-way system and as the number of way increases by one, the speed decreases by three, dissipation time increase exponentially. This result explains the phenomena reported by [1] that two doors with separate ways are much better than two doors with mix ways. Separate ways for pedestrian traffic produce much more efficient flow performance than two-way. The pedestrian interactions strongly influence the pedestrian flow characteristics.

## 6.2 EXPERIMENT ON ELDERLY PEDESTRIANS

In this section, a theoretical experiment on a hypothetical situation was performed. We want to know how the percentage of elderly pedestrians may affect the system performance due to their speed difference.

The total pedestrian is set to be 75 people walking in one direction. The maximum speed of an elderly pedestrian is 50.4 m/min, while the maximum speed for normal people is 88.2 m/min. The approximation speed difference is based on the suggestion of [2].

It is interesting to see that the average speed of the system is decreasing logarithmically as the percentage of the elderly pedestrian increases. A few slow pedestrians will greatly affect average speed of the system, but additional slow pedestrians will have little influence to the system. Figure 6-5 displays the results of the numerical experiments. Only the variation of the average speed can be explained by the variation of the percentage of elderly pedestrians as revealed by the high r-squared value. The other pedestrian characteristics cannot be explained strongly by the variation of elderly pedestrians. The decrease of uncomfortability has very small r-squared, but it may happen because the elderly pedestrians are quite happy with their own slow speed.

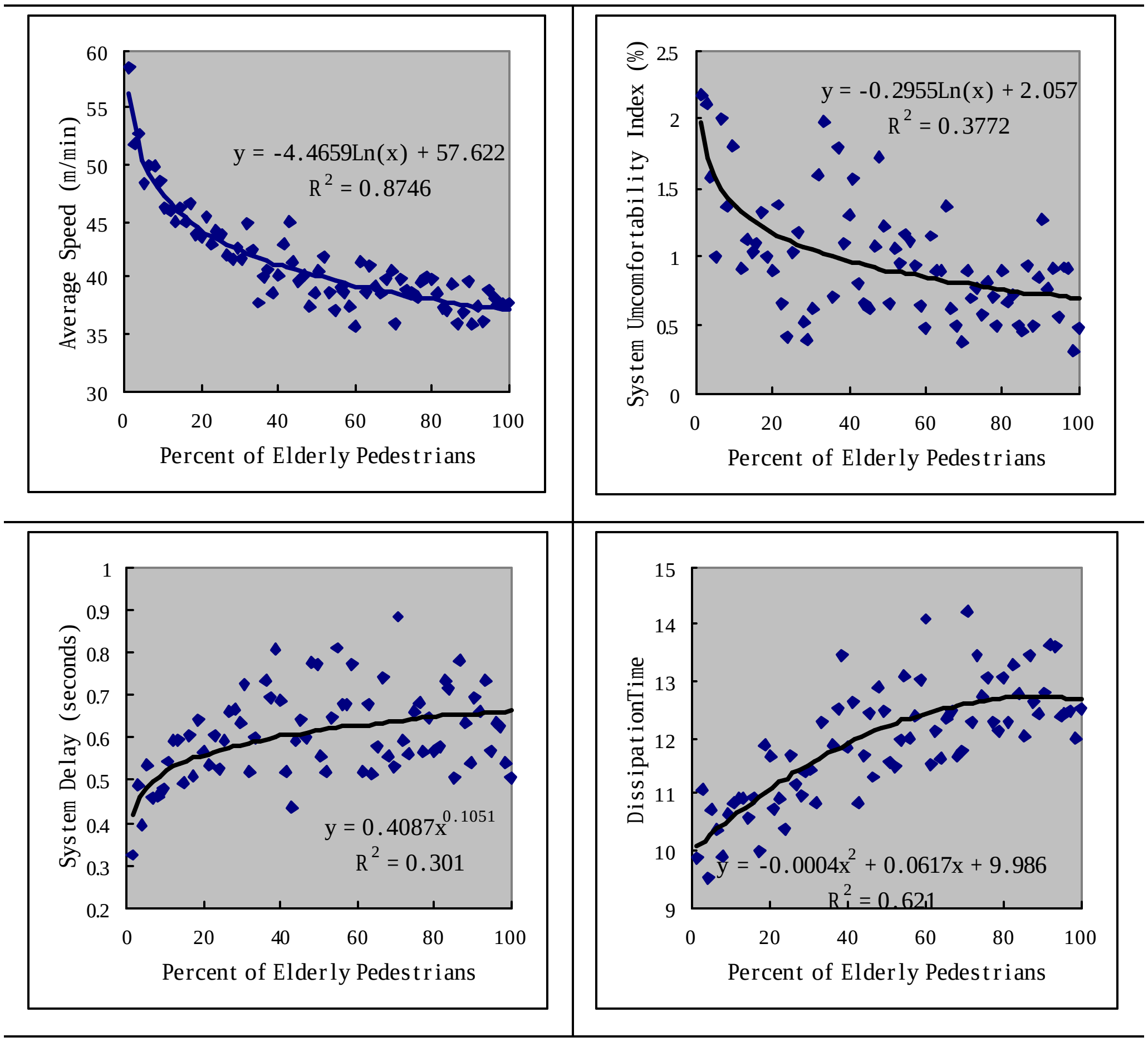

**Figure 6-5 Effect of Elderly Pedestrians**

## 6.3 POLICY ANALYSIS ON PEDESTRIAN CROSSING

One of the objectives of the pedestrian analysis is to evaluate the effects of a proposed policy on the pedestrian facilities before its implementation. The implementation of a policy without pedestrian analysis might lead to a very costly trial and error due to the implementation cost (i.e. user cost, construction time and cost, etc.). On the other hand, using good analysis tools, the trial and error of policy could be done in the analysis level. Once the analysis could prove a good performance, the implementation of the policy is straightforward. The problem is how to evaluate the impact of the policy quantitatively toward the behavior of pedestrians before its implementation. Since the interaction of pedestrians cannot be well addressed using a macroscopic level of analysis, a microscopic level of analysis is the choice.

In this section, a practical case study for policy analysis on a pedestrian crossing was done. A typical pedestrian crossing is a "mix-lane" where pedestrians from both directions meet in the middle of the crossing and interact to avoid each other. As the results of those interactions, it may slow down the walking speed of the pedestrians and increase the delay and dissipation time to cross the road for the same number of pedestrians.

A simple policy such as to keep left (or right), or called "lane-like segregation" might be proposed to reduce the interaction. The implementation of this policy is straightforward. It can be done by marking an arrow on the left side of the starting point of the zebra cross. Using those markings, the pedestrians might be guided to keep left during the crossing. The reduction of the interaction due to lane-like segregation policy may increase the average walking-speed; reduce the delay and the dissipation time.

Using the numerical simulation model as explained in Chapter 4, an experiment on pedestrian crossing was done in two scenarios. The existing condition is called "mix-lane" where pedestrians' initial and target locations are randomly generated at both ends of the crossing. The keep right policy or the "lane-like segregation" was implemented by generating the pedestrians in the lower half (for west to east) and the above half (for east to west). Figure 6-6 shows the condition of both scenarios before they meet in the middle of the crossing.

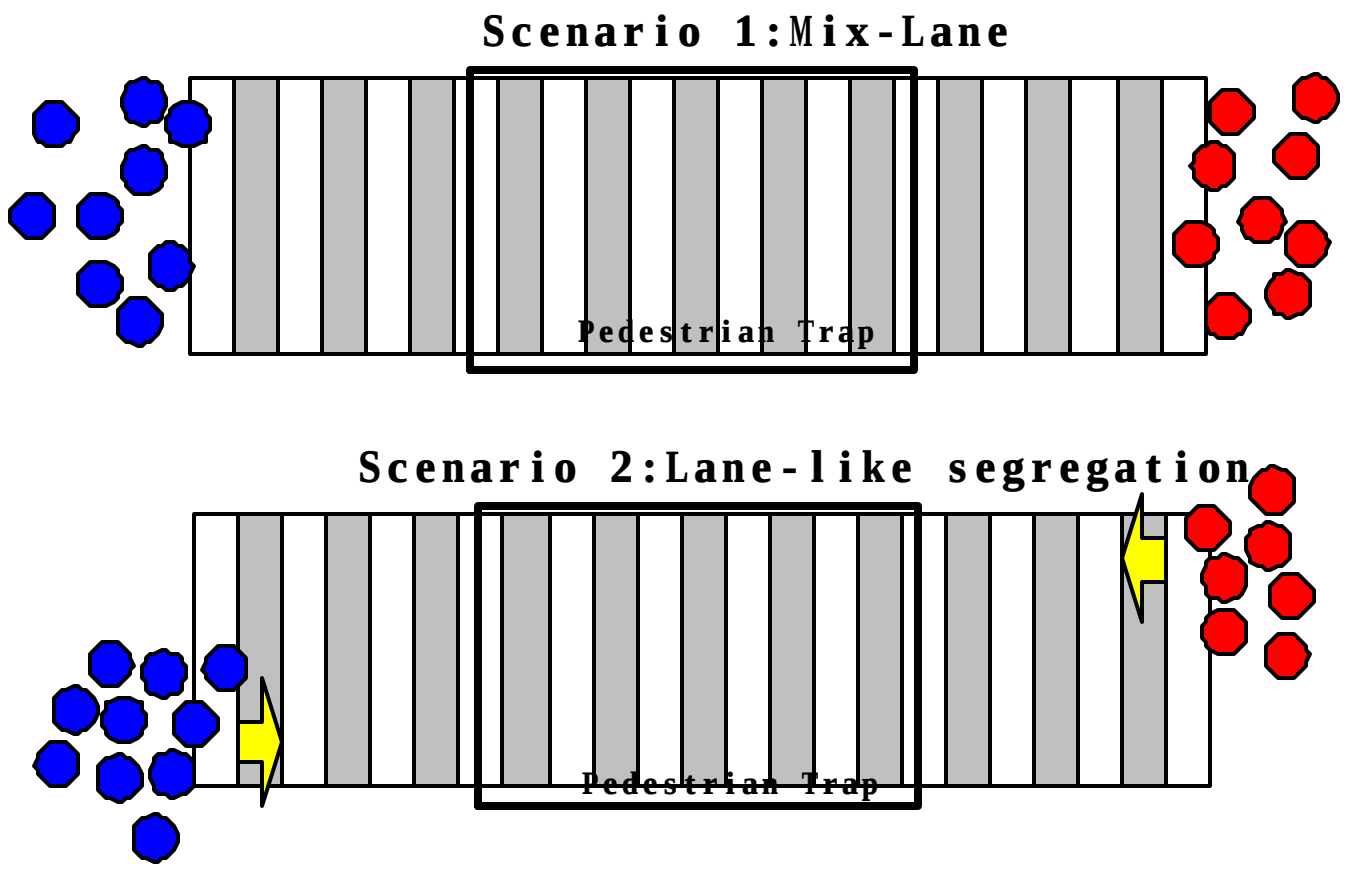

**Figure 6-6 Two scenarios of the experiments of pedestrian crossing**

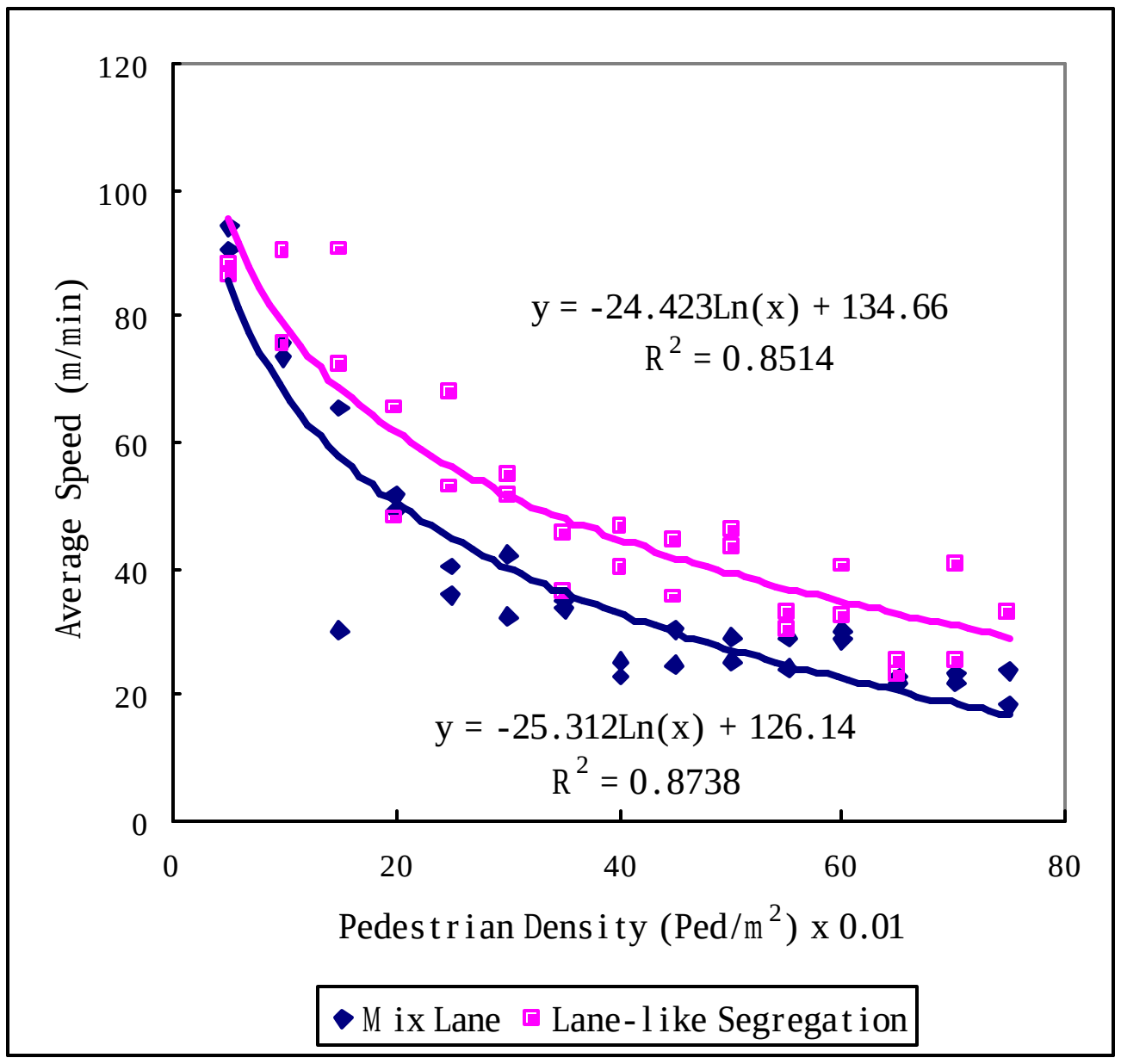

**Figure 6-7. Speed Density Relationship of Mix Lane and Lane-Like Segregation**

The experimental results in Figure 6-7 show that the increasing number of pedestrians will reduce the average speed almost logarithmically. The average speed of Lane-like segregation is higher than about 1.5 times than the mix lane. A higher number of pedestrians tend to increase the average speed difference between the two policies. Increasing the number of pedestrians in a lane-like segregation policy has a tendency to slower the drop of the average speed.

The uncomfortability of both systems is increasing logarithmically as the pedestrian density increases as exemplified in Figure 6-8. The mix lane is much more uncomfortable than the keep right policy. Figure 6-9 illustrates that the time delay is increasing exponentially as the number of pedestrians increase. The lane-like segregation policy has a much lower delay than the mix-lane policy. The change of delay due to the change of the number of pedestrians is also very low for the lane-like segregation policy.

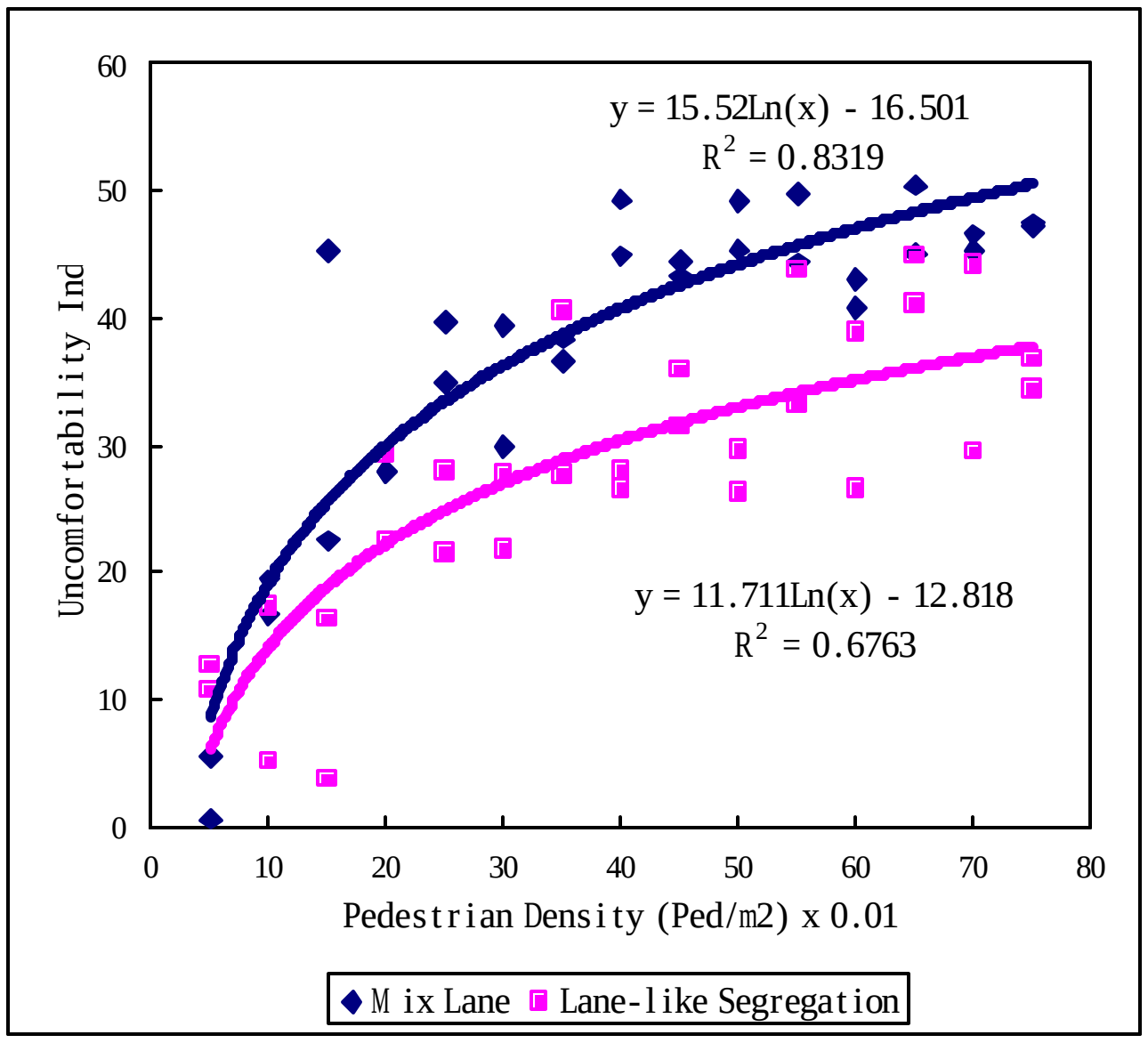

**Figure 6-8 Uncomfortabilities of Mix Lane and Lane-Like Segregation**

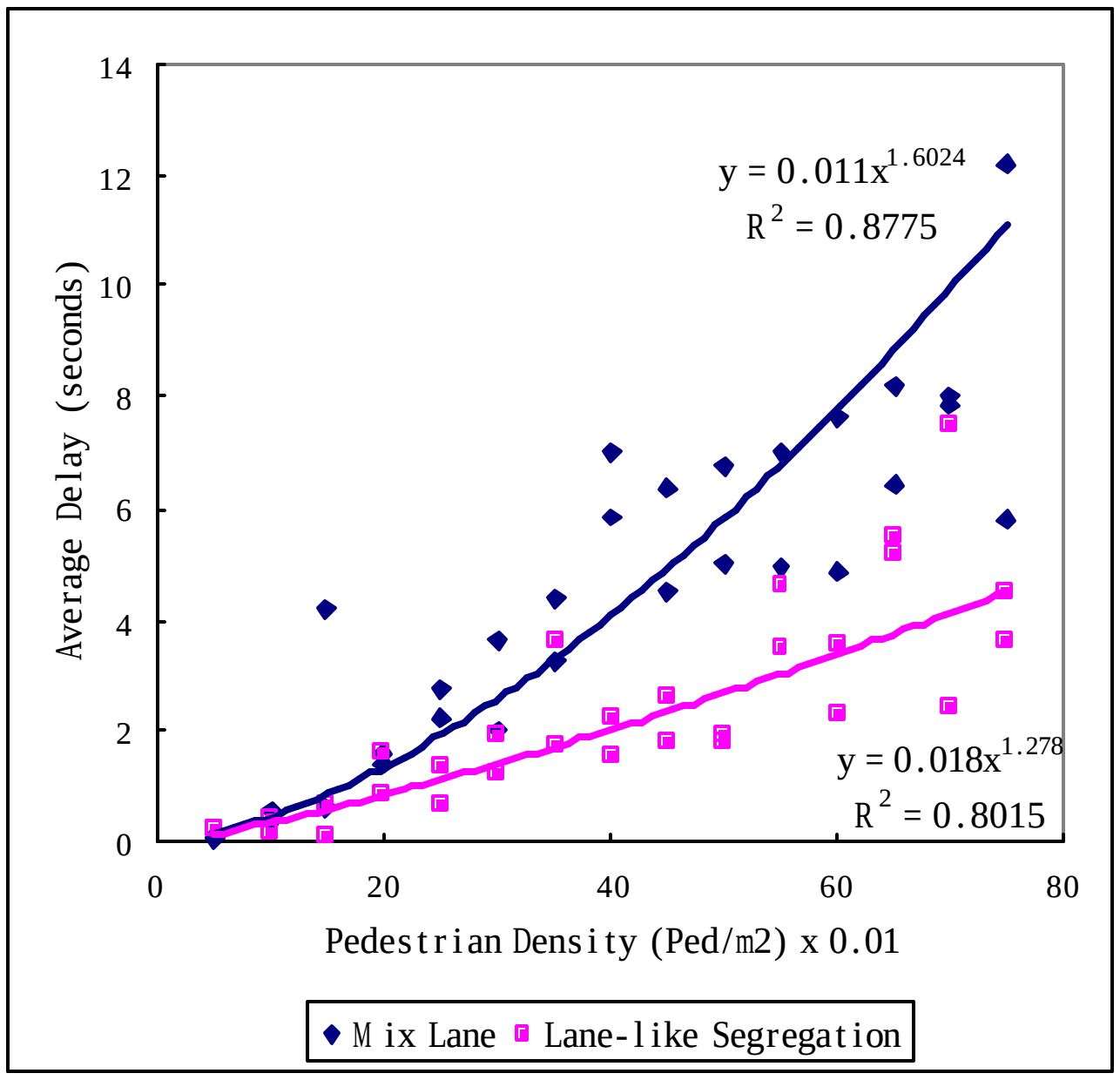

**Figure 6-9. System Delay of Mix Lane and Lane-Like Segregation**

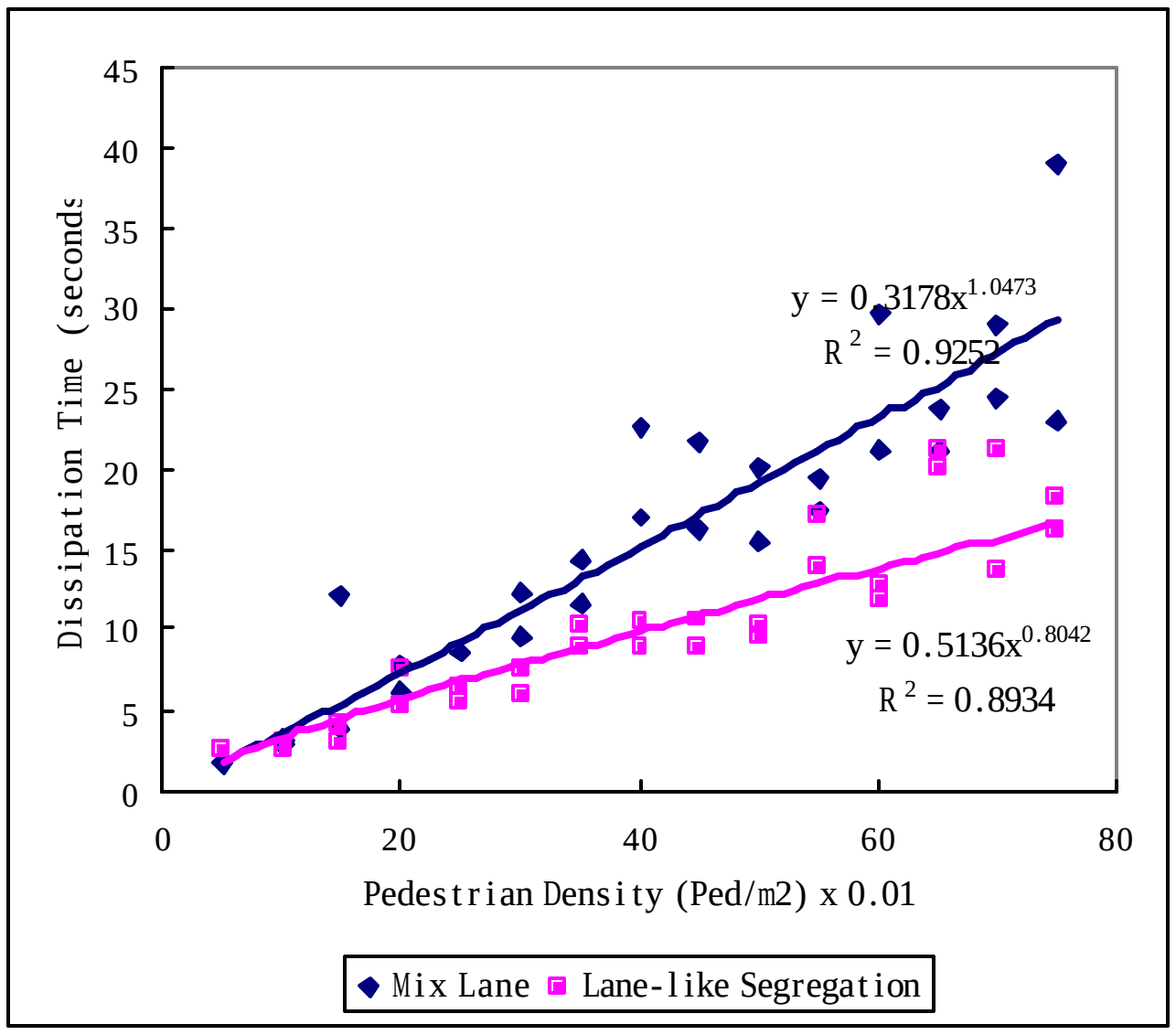

**Figure 6-10 Dissipation-Time of Mix Lane and Lane-Like Segregation**

Figure 6-10 explains that the dissipation time is increasing as the number of pedestrians increase. Similar to the delay, the dissipation time of lane-like segregation is smaller than the mix-lane policy. It explains that for the same numbers of pedestrians that will cross the road, the lane-like segregation policy tend to reduce the time to cross due to less interaction.

Thus, the simulation results showed that the keep -right policy o r the lane-like segregation policy is inclined to be superior to do minimum or mix-lane policy in terms of average speed, uncomfortability, and average delay and dissipation time.

# CHAPTER 7 CONCLUSIONS AND RECOMMENDATIONS

## 7.1 CONCLUSIONS

The study is aimed to improve the quality of pedestrian movement behavior analysis through microscopic pedestrian studies. To achieve this goal, this research has developed a microscopic data collection system and analysis through the simulation model. This development leading to two important messages about pedestrian studies in general:

1. *Digitalization paradigm*. In traditional video data collection, a video is taken from the real world and manual work is done to count the number of pedestrians in the laboratory. When another data (e.g. speed) is needed, another manual work is performed repeatedly toward the same video. Those manual works are not only expensive and exhaustive but also reduce accuracy of the data due manual search. Different results may be obtained by the same person using the same data. Using the microscopic pedestrian video data collection, as explained in this dissertation, the locations of each pedestrian at each time are recorded into a database called *a*TXY database that can be used as a bridge between microscopic pedestrian simulation models, video data collection and microscopic pedestrian characteristics. Such huge information of pedestrian movement now can be reduced into information that can be readily understood and interpreted. The remaining process is executed by computers, which reduce the processing cost and time, and improve accuracy. The same result is always gathered for the same data. This result leads to a new emphasis on the pedestrian research that instead of collecting and reproducing data from the real world, we can use the digitized data by modifying them. Using the simulation model, we are not only modifying the data, but also even creating the database. Pedestrian movement behavior as the quantification of the individual movement and its aggregation into several valuation rates, can be measured through distances, and angles of moving direction. It may be valued over time as a change of distances and angles (e.g. speed) or as a rate of change (e.g. acceleration), or as rate of change of the change (e.g. jerk). They may be quantified for individuals at each time or toward other pedestrians nearby or against the boundary of the surroundings. Variation of ratios, indices and comparison toward some references may also produce different kinds of pedestrian flow performances. The prominent microscopic pedestrian characteristics that have been found through this research are the pedestrian density, average speed, uncomfortability, average delay and dissipation time.

2. *Pedestrian control paradigm*. In the traditional approach of macroscopic pedestrian studies, given a number of pedestrians and a level of service, the model may give the space allocation. To maintain the level of service, the increase of flow always demands wider space. By considering pedestrian interaction in the microscopic level, however, the design of pedestrian facilities is not merely a space allocation but it can also utilize other forms of flow controls in space, time and direction. In the microscopic level, given the same number of pedestrian and the same space, with a better set of rules and detailed design, a better flow performance may be produced. Space of the pedestrian facilities is only one type of pedestrian flow control in the microscopic level. Pedestrian flow that is more efficient can even be reached with less space. The results of the simulation in this research have rejected the direct relationship assumption of space and flow in the macroscopic level. More comprehensive pedestrian-flow control happens because microscopic pedestrian studies consider pedestrian interaction, which is the repulsive and attractive effect among pedestrians and between pedestrians with their environment. Since the movement quality of pedestrians can be improved by controlling the interaction between pedestrians, better pedestrian interaction is the objective of this approach.

Aside from the above paradigms, the following outcomes can be drawn from the results of

this research:

- A data collection system for microscopic pedestrian studies was developed based on the image processing of video data. Semi-automatic data collection was developed successfully while the automatic one was developed for very low pedestrian density. The movement trajectory including the speed profiles can be gathered through the system. It was found that the microscopic speed resemble a normal distribution with mean of 1.38 m/second and standard deviation of 0.37 m/second. The acceleration distribution also bears a resemblance to the normal distribution with the average of 0.68 m/ square second.

- A new microscopic pedestrian simulation model was developed to improve the existing models based on a physical based model. The model was calibrated and validated based on the real world crossing data by minimizing the error between speed distributions.

- The relationship between pedestrian speed and density for one-way traffic was linear while for a two-way pedestrian traffic it was logarithmic.

- For one-way pedestrian traffic, the increase of the pedestrian density tends to increase the uncomfortability and delay exponentially, while the dissipation time increases linearly. For the two-way system, the uncomfortability index increases logarithmically as the total pedestrians in the system increases, while the delay and dissipation time increase exponentially.

- The average speed of the system decreases logarithmically as the percentage of the elderly pedestrian increases. A few slow pedestrians will greatly affect average speed of the system, but additional slow pedestrians will have little influence to the system.

- The simulation results showed that the keep-right policy or the lane-like segregation policy is inclined to be superior to do minimum or mix-lane policy in terms of average speed, uncomfortability, and average delay and dissipation time.

It was revealed that the microscopic pedestrian studies have been successfully applied to give more understanding to the behavior of pedestrians, predict the theoretical and practical situation and evaluate some design policies before its implementation.

## 7.2 FURTHER RESEARCH RECOMMENDATIONS

This research is a beginning of a new understanding for the microscopic pedestrian studies. Limitations of resources and time have been the major constraints. The recommendations below may further improve this study.

- Improvement of the automatic video data collection toward the occlusion problem is highly recommended to enhance the performance of the system for higher pedestrian traffic density.
- Additional model for obstructions and wall avoidance of the microscopic pedestrian simulation model is suggested to perform a better capacity analysis.
- Further search of microscopic pedestrian characteristics may reveal better understanding on the microscopic behavior of pedestrian interaction. Comparison of the microscopic pedestrian characteristics (after obstruction and wall model) with highway capacity manual may suggest a better design standard than the existing level of service standard.
- Further exploration of many applications of the microscopic pedestrian simulation model is recommended.